%% file: main.tex
\documentclass{elsarticle}

\input{preamble.pap.tex}

\input{preamble.macros.tex}

\input{preamble.tikz.tex}

\usepackage{setspace}
\usepackage{enumitem}
\usepackage{ulem}

\newbool{fastcompile}
\setbool{fastcompile}{false}

\begin{document}
\title{Generative Learning of the Solution of Parametric Partial Differential Equations Using Guided Diffusion Models and Virtual Observations }

\author[rvt1]{Han Gao}
\ead{hgao1@seas.harvard.edu}

\author[rvt1]{Sebastian Kaltenbach}
\ead{skaltenbach@seas.harvard.edu}

\author[rvt1]{Petros Koumoutsakos\corref{cor1}}
\ead{petros@seas.harvard.edu}

\address[rvt1]{School of Engineering and Applied Sciences, Harvard University, 29 Oxford Street, Cambridge, MA 02138, US}
\cortext[cor1]{Corresponding author}


\begin{abstract}

We introduce a generative learning framework to model high-dimensional parametric systems using gradient guidance and virtual observations. We consider systems described by Partial Differential Equations (PDEs) discretized with structured or unstructured grids. The framework integrates multi-level information to generate high fidelity time sequences of the system dynamics. We demonstrate the effectiveness and versatility of our framework with two case studies in incompressible, two dimensional,  low Reynolds  cylinder flow on an unstructured mesh and incompressible turbulent channel flow on a structured mesh, both parameterized by the Reynolds number. Our results illustrate the framework's robustness and ability to generate accurate flow sequences across various parameter settings, significantly reducing computational costs allowing for efficient forecasting and reconstruction of flow dynamics.
\end{abstract}
    
\maketitle
\section{Introduction}
High-dimensional Partial Differential Equations (PDEs)  provide  the foundation for simulating complex phenomena such as climate \cite{climatenas}, turbulence \cite{wilcox1988multiscale,brunton2019machine,moser2023numerical}, material behavior \cite{holzapfel2002nonlinear} and epidemics \cite{keeling2005networks}. These simulations often depend on a multitude of input parameters, including variations in boundary conditions and intrinsic parameters of the PDE itself. This parametric dependence, coupled with the substantial computational cost of high-fidelity simulations, poses significant challenges for many-query analyses in the context of optimization, uncertainty quantification, and ``what-if'' scenarios. \cite{koutsourelakis2009accurate,Palmer2015}.\\

Machine learning approaches using neural networks for approximating the unknown variables \cite{raissi2019physics,karniadakis2021physics}, have demonstrated promise in solving PDEs formulated as inverse problems. However,  they fall  short in scenarios involving parameter variations and are less computationally efficient when compared with classical discretisations cast as inverse problems \cite{karnakov2024solving,karnakov2022optimizing}. Neural Operators, such as Fourier Neural Operator (FNO) \cite{li2020fourier} and DeepONet \cite{lu2021learning,wang2021learning}, offer a potent framework for incorporating parametric dependence but struggle with high-dimensional and complex system dynamics. Approaches based on Autoencoders \cite{kingma2013auto} to learn the effective dynamics of high-dimensional PDE systems \cite{kaltenbach2021physics,vlachas2022multiscale,geneva2022transformers,han2022predicting,menier2023interpretable} provide the potential to incorporate parametric dependencies within a suitably constructed latent space. However, the  application of these methods to complex dynamical systems remains limited and has not been extensively explored.\\
Recent diffusion-based models for generative learning \cite{jacobsen2023cocogen, shu2023physics,  li2023multi, gao2024bayesian,du2024confild,wan2024debias} have shown potential for generating high-fidelity solutions of high-dimensional PDE systems. Christian et al. \cite{jacobsen2023cocogen} extended generative models to physical domains by demonstrating their versatility in surrogate modeling, field reconstruction, and inversion from sparse data; Shu et al. \cite{shu2023physics} proposed a diffusion model that uses only high-fidelity data for training to reconstruct high-fidelity CFD data from various inputs, including low-fidelity samples and random measurements; Li et al. \cite{li2023multi} addressed data augmentation in rotating turbulence, finding diffusion models promising and enabling probabilistic reconstructions and uncertainty quantification; Gao et al. \cite{gao2024bayesian} introduced a generative framework using probabilistic diffusion models for versatile turbulence generation, unifying unconditional and conditional sampling within a Bayesian framework, and demonstrated its capabilities through experiments, including LES synthesis, wall-bounded turbulence generation, and super-resolution of turbulent flows; Wan et al. \cite{wan2024debias} introduced a two-stage probabilistic framework for statistical downscaling using unpaired data, involving debiasing via optimal transport and upsampling with a probabilistic diffusion model, demonstrating its effectiveness on fluid flow problems and matching physical statistics accurately from low-resolution inputs; Recently, Du et al. \cite{du2024confild} introduced the Conditional Neural Field Latent Diffusion (CoNFiLD) model, a generative learning framework for rapid simulation of spatiotemporal dynamics in chaotic and turbulent systems within irregular domains. In this paper the authors  integrate  conditional neural field encoding with latent diffusion processes for efficient and probabilistic turbulence generation, adaptable to various scenarios without retraining, and demonstrate its effectiveness through numerical experiments. 

Despite these promising advancements, there remains a significant gap in the exploration of diffusion models for effectively capturing parametric dependencies in complex high-dimensional systems while  preserving  the inherent versatility and efficiency of these models. \\

This paper contributes to addressing these challenges  through a novel generative learning framework that leverages gradient guidance and virtual observations. We deploy  state-of-the-art diffusion models that have achieved remarkable results in Computer Vision \cite{saharia2022photorealistic} as well as both forward predictions \cite{jadhav2023stressd,lienen2023generative,price2023gencast,liu2024uncertainty} and inverse problems \cite{jacobsen2023cocogen, shu2023physics}. Moreover, we enforce desired properties or constraints during the predictions phase by employing the concepts of Virtual Observables \cite{kaltenbach2020incorporating} and gradient guidance \cite{song2020score,dhariwal2021diffusion}. The proposed framework is designed to handle high-dimensional systems and varying parameter inputs without the need for retraining. This capability enables rapid generation of diverse solutions, significantly reducing computational costs and facilitating robust experimentation. We do not attempt to directly predict the full system response due to the resulting large computational costs during both training and predictions of the diffusion model. Instead, we first find a suitable lower-dimensional representation of the system of interest that is able to represent the full high-dimensional solution. In case the high-dimensional PDE solution is given on a structured grid, we deploy a  Convolutional Neural Network (CNN)  whereas in case of data provided on an unstructured grid we resort to a Graph Neural Network (GNN) formulation. The  latter is particularly  relevant as many PDE systems are solved on complex geometries using unstructured grids which makes the application of a CNN not possible.\\ 

The contributions of this work include: 
\begin{enumerate}
\item  a diffusion model augmented with gradient guidance to accurately capture the dynamics of a latent representation of a high-dimensional system. Trained CNNs and GNNs map this latent representation to the full-high-dimensional representation of the system, adapting to diverse structured  and unstructured spatial discretizations. 
\item the integration of multi-level information, ensuring that the generated solutions maintain high fidelity to the underlying physical phenomena across different parameter settings.
\item demonstrations of the robustness and versatility of the framework in  two case studies: an incompressible flow past a 2D  cylinder at low Reynolds numbers on an unstructured mesh and an incompressible 3D turbulent channel flow on a structured mesh, both parameterized by the Reynolds number. 
\end{enumerate}

The remainder of this paper is organized as follows: Section \ref{sec:method} introduces the components of the  proposed framework, including the CNN/GNN for structured/unstructured meshes for its latent space, the diffusion model, gradient guidance, and virtual observation. Section \ref{sec:results} presents the  case studies and results, demonstrating the framework's versatility, while \ref{sec:cflow} highlights how the proposed generative model can handle very complex parametric systems. Finally, Section \ref{sec:conclu} concludes the paper with a discussion of our findings and potential future directions.

\section{Methodology}
\label{sec:method}
 The focus of this paper  is on modeling the dynamics of a parameterized, large-scale system of  equations that  result from the discretization of PDEs. Our approach leverages both micro and macro-level models, structured and unstructured discretizations, and incorporates guidance and virtual observations to enhance the accuracy and efficiency of the model.
\subsection{Micro system}
We consider a parameterized, large-scale system of nonlinear equations, which we refer to as the micro-level model. Given a system parameter vector $\mu\in\mathbb{R}^{N_\mu}$, the goal is to approximately model the implicit distribution 
of the underlying dynamics using generative learning techniques. In more detail, we target the sequence $\mathbf{U}_{n_\mathrm{len}}(\mu)$ consists of $n_\mathrm{len}$ snapshots $\mathbf{u}\in\mathbb{R}^{N_\mathbf{u}}$:
\begin{equation}
    \mathbf{U}_{n_\mathrm{len}}(\mu) = [\mathbf{u}_0(\mu), \mathbf{u}_1(\mu), \dots, \mathbf{u}_{n_\mathrm{len}-1}(\mu)],
\end{equation}
where each snapshot $\mathbf{u}_i$ is implicitly parameterized by $\mu$. This sequence is governed by the large-scale discrete dynamical system:
\begin{equation}
    \mathbf{u}_1(\mu) = F(\mathbf{u}_0(\mu); \mu),
\end{equation}
where $\mu\in\mathbb{R}^{N_\mu}$ also influences the dynamics of the sequence, and $F:\mathbb{R}^{N_\mathbf{u}} \times \mathbb{R}^{N_\mu} \rightarrow \mathbb{R}^{N_\mathbf{u}}$ is a function describing the evolution of the state. The function $F$ determines how the initial state $\mathbf{u}_0(\mu)$ progresses to the subsequent state $\mathbf{u}_1(\mu)$, encapsulating the dynamical behavior of the system under the influence of the parameter $\mu$. The spatial domain ($\Omega$) on which the dynamical system of interest is solved is split into $N_\mathrm{c}$ cells ($\Omega_c$) such that:
\begin{equation}
\Omega = \bigcup_{c=1}^{N_\mathrm{c}} \Omega_c,    
\end{equation}
where each cell $\Omega_c$ is associated with a set of degrees of freedom (DoFs) denoted as $\mathbf{v}_c \in \mathbb{R}^{N_\mathbf{v}}$.\\
We note that the micro-level model can be computationally expensive due to its nonlinearity, the large amount of spatial discretization points, and the necessity of small time increments. While the spatial discretization of the solution can be regularly structured for most geometries and PDEs, in general a varying or unstructured discretization is preferable as some areas require a finer discretization than others.

\subsubsection{Structured Discretization}
In a structured discretization, the state $\mathbf{u}$ is represented on a structured grid:
\begin{equation}
    \mathbf{u} = \begin{bmatrix}
            \mathbf{v}_1 & \mathbf{v}_2 & \dots & \mathbf{v}_{N_\mathrm{w}-1} & \mathbf{v}_{N_\mathrm{w}} \\
            \mathbf{v}_{N_\mathrm{w}+1} & \mathbf{v}_{N_\mathrm{w}+2} & \dots & \mathbf{v}_{2N_\mathrm{w}-1} & \mathbf{v}_{2N_\mathrm{w}} \\
            \vdots & \vdots & \ddots & \vdots & \vdots \\
            \mathbf{v}_{N_\mathrm{w}(N_\mathrm{h}-1)+1} & \mathbf{v}_{N_\mathrm{w}(N_\mathrm{h}-1)+2} & \dots & \mathbf{v}_{N_\mathrm{w}N_\mathrm{h}-1} & \mathbf{v}_{N_\mathrm{w}N_\mathrm{h}}
    \end{bmatrix} \in \mathbb{R}^{N_\mathrm{h} \times N_\mathrm{w} \times N_\mathbf{v}},
\end{equation}
where $N_\mathbf{u}$ can be factorized as $N_\mathbf{u} = N_\mathrm{h} \times N_\mathrm{w} \times N_\mathbf{v}$. Here, $N_\mathrm{h}$ and $N_\mathrm{w}$ represent the number of elements along the height and width of the grid, respectively, while $N_\mathbf{v}$ indicates the degrees of freedom per element. This structured grid allows for an well ordered representation of the state, facilitating straightforward convolutional operations and data manipulation.

\subsubsection{Unstructured Discretization}
Alternatively, in an unstructured discretization, the state $\mathbf{u}$ is represented in an unstructured format:
\begin{equation}
    \mathbf{u} = \begin{bmatrix}
           \mathbf{v}_1 \\ 
           \mathbf{v}_2 \\ 
           \vdots \\ 
           \mathbf{v}_{N_\mathrm{c}-1} \\
           \mathbf{v}_{N_\mathrm{c}}
    \end{bmatrix} \in \mathbb{R}^{N_\mathrm{c} \times N_\mathbf{v}},
\end{equation}
where $N_\mathbf{u}$ can be factorized as $N_\mathbf{u} = N_\mathrm{c} \times N_\mathbf{v}$. Here, $N_\mathrm{c}$ denotes the number of unstructured cells, each with $N_\mathbf{v}$ degrees of freedom. This approach is more flexible than structured discretization, allowing for complex geometries and varying element sizes, which can be advantageous in capturing intricate features of the solution domain.

\subsection{Macro system}
Directly learning the micro-level system can be computationally prohibitive due to the high dimensionality, nonlinearity, and the need for small time increments.  To address these challenges, we convert the micro-level system to a macro-level representation. This conversion facilitates more efficient learning and computation by reducing the complexity of the system \cite{kingma2013auto,vlachas2018}. Different discretization methods, whether structured or unstructured, can be employed in this conversion process to ensure that the macro-level model accurately captures the essential dynamics of the original micro-level system while significantly reducing computational costs.
\subsubsection{Macro states of structured data}
We assume the micro state $\ubm$ from a structured discretization lies in a low-dimensional subspace  
\begin{equation}
    \Vcal_{\thetabold_\mathrm{CNN}} = \bigl\{ D_{\thetabold_\mathrm{CNN}}(\zbm)\;\big|\; \zbm\in\Rbb^{N_\mathrm{h_r} \times N_\mathrm{w_r}\times N_{\vbm_\mathrm{r}}  }  \bigr\}\ \subset\Rbb^{N_\mathrm{h}\times N_\mathrm{w} \times N_\vbm},
\end{equation}
where $D_{\thetabold_\mathrm{CNN}}:\Rbb^{N_\mathrm{h_r} \times N_\mathrm{w_r}\times N_{\vbm_\mathrm{r}}  }\rightarrow\Rbb^{N_\mathrm{h}\times N_\mathrm{w} \times N_\vbm}$ is a convolutional decoder that maps the macro state $\zbm$ to its micro state $D_{\thetabold_\mathrm{CNN}}(\zbm)$ with trainable parameter $\thetabold_\mathrm{CNN}$, where $N_\mathrm{h_r}\ll N_\mathrm{h}, N_\mathrm{w_r}\ll N_\mathrm{w}, \frac{N_{\vbm_\mathrm{r}}}{N_{\vbm}}\sim O(1)$. The parameter $\thetabold_\mathrm{CNN}$ represents the weights and bias parameters of the convolutional filter. 

A convolutional layer applies a set of filters $\Wbm\in\Rbb^{F_\mathrm{h}\times F_\mathrm{w} \times F_\mathrm{f}\times F_\mathrm{k}}$ and a bias vector $\bbm\in\Rbb^{F_\mathrm{k}}$, where $F_\mathrm{h}, F_\mathrm{w}, F_\mathrm{f}, F_\mathrm{k}$ are the filter height, filter width, the number of features, and the number of filters in the layer, respectively. The output of the convolutional layer given an input $\Xbm\in\Rbb^{\mathrm{H}\times\mathrm{W}\times F_\mathrm{f}}$, is given by:
\begin{equation}
\Xbm\mapsto\Ybm\in\Rbb^{\mathrm{H}\times\mathrm{W}\times F_\mathrm{k}}:Y_{i,j,k}=\sum_{c=1}^{F_\mathrm{f}}\sum_{m=1}^{F_\mathrm{h}}\sum_{n=1}^{F_\mathrm{w}} X_{i+m-1,j+n-1,c}\cdot W_{m,n,c,k}+b_k.
\end{equation}
In the decoder, the small height ($N_\mathrm{h_r}$) and width ($N_\mathrm{w_r}$) can be gradually increased to the original height ($N_\mathrm{h}$) and width ($N_\mathrm{w}$) by applying up-sampling before applying a standard convolution and non-linear activation functions \cite{geneva2022transformers}. 

To train the model, we let $\Pcal_\mathrm{train} = \{\mu_1,\mu_2,\dots\}\subset\Rbb^{N_\mu}$ be a collection of decoder training parameters and $\Ucal_\mathrm{train} = \{\Ubm_{n_\mathrm{len}}(\mu_1), \Ubm_{n_\mathrm{len}}(\mu_2),\dots\}$ be a collection of decoder training sequences. The optimal parameters $\thetabold_\mathrm{CNN}^*$ are obtained by solving the following optimization problem:
\begin{equation}
    \thetabold_\mathrm{CNN}^* = \argoptunc{\thetabold_\mathrm{CNN}}{\sum_{\Ubm\in\Ucal_\mathrm{train}} \sum_{\ubm\in\Ubm} || \ubm - D_{\thetabold_\mathrm{CNN}}(E(\ubm)) ||},
\end{equation}
where $E:\Rbb^{N_\mathrm{h}\times N_\mathrm{w} \times N_\vbm} \rightarrow  \Rbb^{N_\mathrm{h_r} \times N_\mathrm{w_r}\times N_{\vbm_\mathrm{r}} }$ is a parameter-free down-sampler acting as the encoder.

\subsubsection{Macro states of unstructured data}
Next, we consider the micro state $\ubm$ from an unstructured discretization that lies in a low-dimensional subspace
\begin{equation}
    \Vcal_{\thetabold_\mathrm{GNN}} = \bigl\{ 
    D_{\thetabold_\mathrm{GNN}}(\zbm) \;\big|\;
    \zbm\in\Rbb^{N_\mathrm{h_r} \times N_\mathrm{w_r}\times N_{\vbm_\mathrm{r}}} 
    \bigr\}  \subset  \Rbb^{N_\mathrm{c} \times N_\vbm},
\end{equation}
where $D_{\thetabold_\mathrm{GNN}}:\Rbb^{N_\mathrm{h_r} \times N_\mathrm{w_r}\times N_{\vbm_\mathrm{r}}}\rightarrow \Rbb^{N_\mathrm{c} \times N_\vbm}$ is a decoder based on a graph neural network with trainable parameter $\thetabold_\mathrm{GNN}$. 

The general convolutional operation over an unstructured mesh can be expressed as a message-passing process,
\begin{equation}
    \Xbm=\begin{bmatrix}
           \xbm_1 \\ 
           \xbm_2 \\ 
           \dots  \\ 
           \xbm_{N_\mathrm{c}-1} \\
           \xbm_{N_\mathrm{c}} 
    \end{bmatrix}\mapsto 
    \Ybm=\begin{bmatrix}
           \ybm_1 \\ 
           \ybm_2 \\ 
           \dots  \\ 
           \ybm_{N_\mathrm{c}-1} \\
           \ybm_{N_\mathrm{c}} 
        \end{bmatrix}:
\ybm_i = \gamma_{\thetabold_\mathrm{GNN}}\bigl(
\xbm_i,\oplus_{j\in\Ncal(i)} \phi_{\thetabold_\mathrm{GNN}}(\xbm_i,\xbm_j,\ebm_{j,i})
\bigr),
\label{eqn:gnn_encdoer}
\end{equation}
where $\oplus$ denotes a differentiable, permutation-invariant function, such as sum, mean, or max, $\ebm\in\Rbb^{N_\ebm}$ are edge features from cell $j$ to cell $i$, and $\phi_{\thetabold_\mathrm{GNN}}:\Rbb^{N_\xbm}\times\Rbb^{N_\xbm}\times\Rbb^{N_\ebm}\rightarrow\Rbb^{N_\phi}$ is a differentiable function that processes the node pair and their edge interaction. The function $\gamma_{\thetabold_\mathrm{GNN}}:\Rbb^{N_\xbm}\times\Rbb^{N_\phi}\rightarrow\Rbb^{N_\ybm}$ updates the node feature.

To optimize the parameters of the GNN, we define $\thetabold_\mathrm{GNN}^*$ as the solution to the following optimization problem:
\begin{equation}
    \thetabold_\mathrm{GNN}^* = \argoptunc{\thetabold_\mathrm{GNN}}{\sum_{\Ubm\in\Ucal_\mathrm{train}} \sum_{\ubm\in\Ubm} || \ubm - D_{\thetabold_\mathrm{GNN}}(E_{\thetabold_\mathrm{GNN}}(\ubm)) ||},
\end{equation}
where the construction of the encoder and decoder is based on the methodology described in \cite{han2022predicting}.

\subsection{Generative Learning with Diffusion Modeling}

Given the encoder and decoder introduced in the previous sections, we can map the micro state sequence $\Ubm_{n_\mathrm{len}}(\mu)$ to a low-dimensional latent space representation $\Zbm_{n_\mathrm{len}}(\mu)$ as follows:
\begin{equation}
    \Zbm_{n_\mathrm{len}}(\mu) = 
    [
    \zbm_0(\mu), \zbm_1(\mu), \dots, \zbm_{n_\mathrm{len}-1}(\mu)
    ]\in\Rbb^{N_\mathrm{h_r} \times N_\mathrm{w_r}\times N_{\vbm_\mathrm{r}}  \times n_\mathrm{len}} ,
    \label{eqn:macro_seq}
\end{equation}
where each $\zbm_i(\mu) = E_{\thetabold^*}(\ubm(\mu))$ is obtained through the encoder $E_{\thetabold^*}$. By performing diffusion learning on the macro level, we achieve computational efficiency. The forward process, which models the addition of noise on the macro level, is expressed as:
\begin{equation}
    \Tilde{\Zbm}^{(i)}_{n_\mathrm{len}}(\mu) \sim q^{[\mu]}_i\Bigl(\Tilde{\Zbm}^{(i)}_{n_\mathrm{len}}(\mu) \Big| \Zbm_{n_\mathrm{len}}(\mu)\Bigr) \coloneqq \Ncal\Bigl(\Zbm_{n_\mathrm{len}}(\mu),\sigma^2_i\Ibm\Bigr),\quad i = 1,\dots,N_\mathrm{noise},
\end{equation}
where $\Ibm\in\Rbb^{N_\mathrm{h_r}  N_\mathrm{w_r} N_{\vbm_\mathrm{r}} n_\mathrm{len} \times N_\mathrm{h_r}  N_\mathrm{w_r} N_{\vbm_\mathrm{r}} n_\mathrm{len}}$ is the identity matrix. This step-by-step addition of Gaussian noise ensures that the data distribution becomes more tractable, simplifying the training of the generative model. 
Conversely, the reverse process involves denoising, leveraging a neural network to iteratively refine the noisy latent representations back to their original form. This bidirectional process not only aids in learning the underlying data distribution but also enhances the robustness of the model, enabling it to generate high-quality samples. 
The reverse process, aimed at denoising and sampling, is defined as:
\begin{equation}
    p^{[\mu]}\Bigl(\Tilde{\Zbm}^{(i)}_{n_\mathrm{len}}(\mu)\Big|\Zbm_{n_\mathrm{len}}(\mu),  \Tilde{\Zbm}^{(i+1)}_{n_\mathrm{len}}(\mu) \Bigr)=
    \Ncal\Biggl(   
    \frac{\sigma^2_{i+1}-\sigma^2_i}{\sigma^2_{i+1}}
    \Zbm_{n_\mathrm{len}}(\mu)+
    \frac{\sigma^2_i}{\sigma^2_{i+1}} \Tilde{\Zbm}^{(i+1)}_{n_\mathrm{len}}(\mu),
    \frac{(\sigma^2_{i+1}-\sigma^2_i)\sigma^2_i}{\sigma^2_{i+1}}
    \Biggr),
\end{equation}
To approximate the reverse process, we use a neural network parameterized by $\thetabold_{\mathrm{diff}}$:
\begin{equation}
p^{[\mu]}_{\thetabold_{\mathrm{diff}}}\Bigl(\Tilde{\Zbm}^{(i)}_{n_\mathrm{len}}(\mu)\Big|\Zbm_{n_\mathrm{len}}(\mu),  \Tilde{\Zbm}^{(i+1)}_{n_\mathrm{len}}(\mu),i \Bigr)=
\Ncal\Biggl(   
    \frac{\sigma^2_{i+1}-\sigma^2_i}{\sigma^2_{i+1}}
    \hat{\Zbm}_{n_\mathrm{len},\thetabold_{\mathrm{diff}}}\bigl(\mu,\Tilde{\Zbm}^{(i+1)}_{n_\mathrm{len}}(\mu),i\bigr)+
    \frac{\sigma^2_i}{\sigma^2_{i+1}} \Tilde{\Zbm}^{(i+1)}_{n_\mathrm{len}}(\mu),
    \frac{(\sigma^2_{i+1}-\sigma^2_i)\sigma^2_i}{\sigma^2_{i+1}}
    \Biggr),
\end{equation}
where $\hat{\Zbm}_{n_\mathrm{len},\thetabold_{\mathrm{diff}}}\bigl(\mu,\Tilde{\Zbm}^{(i+1)}_{n_\mathrm{len}}(\mu),i\bigr):\Rbb^{N_\mu}\times \Rbb^{N_\mathrm{h_r} \times N_\mathrm{w_r}\times N_{\vbm_\mathrm{r}}  \times n_\mathrm{len}}\times \Nbb\rightarrow \Rbb^{N_\mathrm{h_r} \times N_\mathrm{w_r}\times N_{\vbm_\mathrm{r}}  \times n_\mathrm{len}}$ is a neural network parametrized by $\thetabold_\mathrm{diff}$. The optimal parameters $\thetabold_\mathrm{diff}^*$ are obtained by solving the following optimization problem \cite{saharia2022photorealistic}:
\begin{equation}
\thetabold_\mathrm{diff}^* =\argoptunc{\thetabold_\mathrm{diff}}{\sum_{\mu\in\Pcal_\mathrm{train}}\Ebb_{i\sim \mathrm{Uniform}(1,N_\mathrm{noise})}\Bigl[\Bigl|\Bigl|\hat{\Zbm}_{n_\mathrm{len},\thetabold_{\mathrm{diff}}}\bigl(\mu,\Tilde{\Zbm}^{(i+1)}_{n_\mathrm{len}}(\mu),i\bigr)-\Zbm_{n_\mathrm{len}}(\mu)\Bigr|\Bigr|^2_2\Bigr]}.
\end{equation}
After training, the term $\hat{\Zbm}_{n_\mathrm{len},\thetabold^*_{\mathrm{diff}}}(\mu, \Tilde{\Zbm}^{(i)}_{n_\mathrm{len}}, i)$ represents the estimated latent representation given $\Tilde{\Zbm}^{(i)}_{n_\mathrm{len}}$.

\subsection{Multi-resolution guidance and virtual guidance}
During the reverse process, the score function is modified to incorporate guidance, ensuring that the generated samples adhere to desired properties or constraints.
\subsubsection{Gradient guidance for diffusion models}
The gradient of the log probability density (score) function is approximated as follows:
\begin{equation}
    \nabla_{\Tilde{\Zbm}} \log p^{[\mu]}\Bigl(  
    \Tilde{\Zbm}^{(i)}_{n_\mathrm{len}}(\mu)
    \Bigr) \approx \frac{\hat{\Zbm}_{n_\mathrm{len},\thetabold^*_{\mathrm{diff}}}\bigl(\mu,\Tilde{\Zbm}^{(i)}_{n_\mathrm{len}}(\mu),i\bigr)-\Tilde{\Zbm}^{(i)}_{n_\mathrm{len}}(\mu) }{\sigma^2_i}.
\end{equation}
This equation represents the originally-learned score function, where the guidance term will modify the gradient to reflect the desired properties of the generated samples. To guide the generation process, physical information can be incorporated. The residual term which represents the mismatch between the sample and observation is defined as,
\begin{equation}
\hat{\Rbm}\Bigl(\hat{\Zbm}_{n_\mathrm{len},\thetabold^*_{\mathrm{diff}}}\bigl(\mu,\Tilde{\Zbm}^{(i)}_{n_\mathrm{len}}(\mu),i\bigr), \mu \Bigr)= H\Bigl(\hat{\Zbm}_{n_\mathrm{len},\thetabold^*_{\mathrm{diff}}}\bigl(\mu,\Tilde{\Zbm}^{(i)}_{n_\mathrm{len}}(\mu),i\bigr)\Bigr)-P_\mathrm{obs}(\mu),
\end{equation}
where $H:\Rbb^{N_\mathrm{h_r} \times N_\mathrm{w_r}\times N_{\vbm_\mathrm{r}}  \times n_\mathrm{len}}\rightarrow\Rbb^{N_\mathrm{obs}}$ maps the macro state to a vector that is implicitly governed by the parameter $\mu$ via $P_\mathrm{obs}:\Rbb^{N_\mu}\rightarrow \Rbb^{N_\mathrm{obs}}$. To guide the latent representations to satisfy the physical constraints, we then define the virtual likelihood \cite{kaltenbach2020incorporating} as follows:
\begin{equation}
   p^{[\mu]}\Bigl(
\hat{\Rbm}\Bigl(\hat{\Zbm}_{n_\mathrm{len},\thetabold^*_{\mathrm{diff}}}\bigl(\mu,\Tilde{\Zbm}^{(i)}_{n_\mathrm{len}}(\mu),i\bigr), \mu \Bigr)= 0
\Big|
\Tilde{\Zbm}^{(i)}_{n_\mathrm{len}}(\mu)
   \Bigr)
   =\Ncal\Bigl(
\hat{\Rbm}\Bigl(\hat{\Zbm}_{n_\mathrm{len},\thetabold^*_{\mathrm{diff}}}\bigl(\mu,\Tilde{\Zbm}^{(i)}_{n_\mathrm{len}}(\mu),i\bigr), \mu \Bigr),\sigma_R^2\Ibm
   \Bigr)\Big|_{0}.
\end{equation}
Combining the virtual likelihood and the Bayes law,  we obtain the original score function with the guidance term for the guided reverse process:
\begin{equation}
\begin{aligned}
&p^{[\mu]} \Bigl(
\Tilde{\Zbm}^{(i)}_{n_\mathrm{len}}(\mu) \Big|
\hat{\Rbm}\Bigl(\hat{\Zbm}_{n_\mathrm{len},\thetabold^*_{\mathrm{diff}}}\bigl(\mu,\Tilde{\Zbm}^{(i)}_{n_\mathrm{len}}(\mu),i\bigr), \mu \Bigr)=0
\Bigr) \propto \\
&p^{[\mu]}\Bigl(\Tilde{\Zbm}^{(i)}_{n_\mathrm{len}}(\mu)\Bigr)\cdot
p^{[\mu]} \Bigl(
\hat{\Rbm}\Bigl(\hat{\Zbm}_{n_\mathrm{len},\thetabold^*_{\mathrm{diff}}}\bigl(\mu,\Tilde{\Zbm}^{(i)}_{n_\mathrm{len}}(\mu),i\bigr), \mu \Bigr)=0
\Big|
\Tilde{\Zbm}^{(i)}_{n_\mathrm{len}}(\mu)
\Bigr).
\end{aligned}
\end{equation}
Then, the combined score function with guidance can be expressed as:
\begin{equation}
\begin{aligned}
& \nabla_{\Tilde{\Zbm}} \log p^{[\mu]} \Bigl(
\Tilde{\Zbm}^{(i)}_{n_\mathrm{len}}(\mu) \Big|
\hat{\Rbm}\Bigl(\hat{\Zbm}_{n_\mathrm{len},\thetabold^*_{\mathrm{diff}}}\bigl(\mu,\Tilde{\Zbm}^{(i)}_{n_\mathrm{len}}(\mu),i\bigr), \mu \Bigr)=0
\Bigr) = \\
& \nabla_{\Tilde{\Zbm}} \log p^{[\mu]}\Bigl(\Tilde{\Zbm}^{(i)}_{n_\mathrm{len}}(\mu)\Bigr) + 
 \nabla_{\Tilde{\Zbm}} \log p^{[\mu]} \Bigl(
\hat{\Rbm}\Bigl(\hat{\Zbm}_{n_\mathrm{len},\thetabold^*_{\mathrm{diff}}}\bigl(\mu,\Tilde{\Zbm}^{(i)}_{n_\mathrm{len}}(\mu),i\bigr), \mu \Bigr)=0
\Big|
\Tilde{\Zbm}^{(i)}_{n_\mathrm{len}}(\mu)
\Bigr).
\end{aligned}
\end{equation}
We calculate the gradient of the log probability of the guidance term using automatic differentiation,
\begin{equation}
\begin{aligned}
    &\nabla_{\Tilde{\Zbm}} \log p^{[\mu]} \Bigl(
\hat{\Rbm}\Bigl(\hat{\Zbm}_{n_\mathrm{len},\thetabold^*_{\mathrm{diff}}}\bigl(\mu,\Tilde{\Zbm}^{(i)}_{n_\mathrm{len}}(\mu),i\bigr), \mu \Bigr)=0
\Big|
\Tilde{\Zbm}^{(i)}_{n_\mathrm{len}}(\mu)
\Bigr)  =\\
&-\frac{1}{\sigma_R^2}\hat{\Rbm}\Bigl(\hat{\Zbm}_{n_\mathrm{len},\thetabold^*_{\mathrm{diff}}}\bigl(\mu,\Tilde{\Zbm}^{(i)}_{n_\mathrm{len}}(\mu),i\bigr), \mu \Bigr) \pder{H\Bigl(\hat{\Zbm}_{n_\mathrm{len},\thetabold^*_{\mathrm{diff}}}\bigl(\mu,\Tilde{\Zbm}^{(i)}_{n_\mathrm{len}}(\mu),i\bigr)\Bigr)}{\Tilde{\Zbm}^{(i)}_{n_\mathrm{len}}(\mu)}.
\end{aligned}
\end{equation}

\subsubsection{Multi-resolution information}

In the multi-resolution guidance framework, we provide guidance at both micro and macro levels to ensure that the generated samples adhere to the desired properties across different scales. The mapping functions for both levels of guidance are defined as follows:
\begin{equation}
    H: \begin{cases}
        \Zbm \mapsto   G^{\mathrm{micro}}_\mathrm{readout}(D_{\thetabold^*}(\Zbm))\quad&\text{micro-level guidance}, \\
        \Zbm\mapsto G^{\mathrm{macro}}_{\mathrm{readout}}(\Zbm)&\text{macro-level guidance}.
    \end{cases}
\end{equation}
Here, $H$ represents the mapping function that transforms the latent space representation $\Zbm$ into either a micro-level or macro-level guidance signal. The micro-level guidance is obtained through the decoder $D_{\thetabold^*}$, which maps $\Zbm$ back to the micro-level state before applying the readout function $G^{\mathrm{micro}}_\mathrm{readout}$. On the other hand, the macro-level guidance directly applies the readout function $G^{\mathrm{macro}}_{\mathrm{readout}}$ to the latent representation $\Zbm$.

The observational data $P_\mathrm{obs}$ is also mapped for both micro and macro levels to provide consistent guidance:
\begin{equation}
    P_\mathrm{obs}:
    \begin{cases}
        \mu \mapsto G^\mathrm{micro}_\mathrm{readout}(\Ubm_{n_\mathrm{len}}(\mu)) &\text{micro-level guidance},\\
        \mu \mapsto G^\mathrm{macro}_\mathrm{readout}(E_{\thetabold^*}(\Ubm_{n_\mathrm{len}} (\mu)))\quad&\text{macro-level guidance}.\\
    \end{cases}
\end{equation}
In this context, $P_\mathrm{obs}$ maps the system parameter $\mu$ to the observed data, providing the reference for guidance at both levels. The micro-level observation $G^\mathrm{micro}_\mathrm{readout}(\Ubm_{n_\mathrm{len}}(\mu))$ is derived directly from the micro state sequence $\Ubm_{n_\mathrm{len}}(\mu)$. Conversely, the macro-level observation $G^\mathrm{macro}_\mathrm{readout}(E_{\thetabold^*}(\Ubm_{n_\mathrm{len}} (\mu)))$ is obtained by first encoding the micro state sequence into the latent space and then applying the macro-level readout function.

This multi-resolution approach ensures that the model captures the essential dynamics and constraints at both detailed (micro) and coarse (macro) scales, thus improving the fidelity and robustness of the generative process.

\subsubsection{Virtual observation}

The concept of virtual observation is employed to ensure that the generated samples closely follow the observed data distribution. In many scenarios, direct observations may be sparse or incomplete, making it challenging to train the generative model effectively. To address this, we leverage other models, such as multilayer perceptrons (MLP) or Gaussian processes, to predict these properties as virtual observations.
The virtual observation is used to approximate the true observation as
\begin{equation}
    \hat{P}_{\mathrm{obs},\thetabold^*}(\mu)\approx  G^\mathrm{micro}_\mathrm{readout}(\Ubm_{n_\mathrm{len}}(\mu)).
\end{equation}
Here, $\hat{P}_{\mathrm{obs},\thetabold^*}(\mu)$ represents the estimated observation given the parameter $\mu$, which is trained by applying the micro-level readout function $G^\mathrm{micro}_\mathrm{readout}$ to the training micro state sequence $\Ubm_{n_\mathrm{len}}(\mu)$. In cases where direct observational data for a unseen parameter is not available, these predictions can be generated using pre-trained models or Gaussian processes that capture the relevant system properties. By integrating such predictive models, our framework is well-equipped to leverage existing knowledge and models to generate high-fidelity virtual observations. This not only enriches the training data for the generative model but also enhances the robustness and accuracy of the generated samples. The flexibility of using different predictive models allows us to incorporate a wide range of prior knowledge, making the generative process more reliable and efficient.

\section{Results}
\label{sec:results}
In this section, we demonstrate the effectiveness of the proposed generative framework through two case studies: incompressible laminar cylinder flow on an unstructured mesh (Section~\ref{sec:fpc}) and incompressible turbulent channel flow on a structured mesh (Section~\ref{sec:cflow}), both parameterized by the Reynolds number. In Section~\ref{sec:fpc}, we focus on evaluating the robustness and versatility of the framework under various settings to illustrate its capability to directly generate, forecast, and reconstruct flow sequences across different Reynolds-numbers. Additionally, in Section~\ref{sec:cflow}, we emphasize how the framework leverages multi-level information from virtual observations to guide the generative process for complex dynamical systems.

\subsection{Laminar cylinder flow}
\label{sec:fpc}
\begin{figure}[htp]
	\centering
	\begin{tikzpicture}
		\begin{groupplot}[
			group style={
				group size=1  by 1,
				horizontal sep=1cm
			},
			width=1\textwidth,
			axis equal image,
			xlabel={$x$},
			ylabel={$y$},
			xtick = {-4, 0, 30},
			ytick = {-5, 0, 5},
			xmin=-4, xmax=30,
			ymin=-5, ymax=5
			]
			\nextgroupplot[]
			\addplot graphics [xmin=-4, xmax=30, ymin=-5, ymax=5] {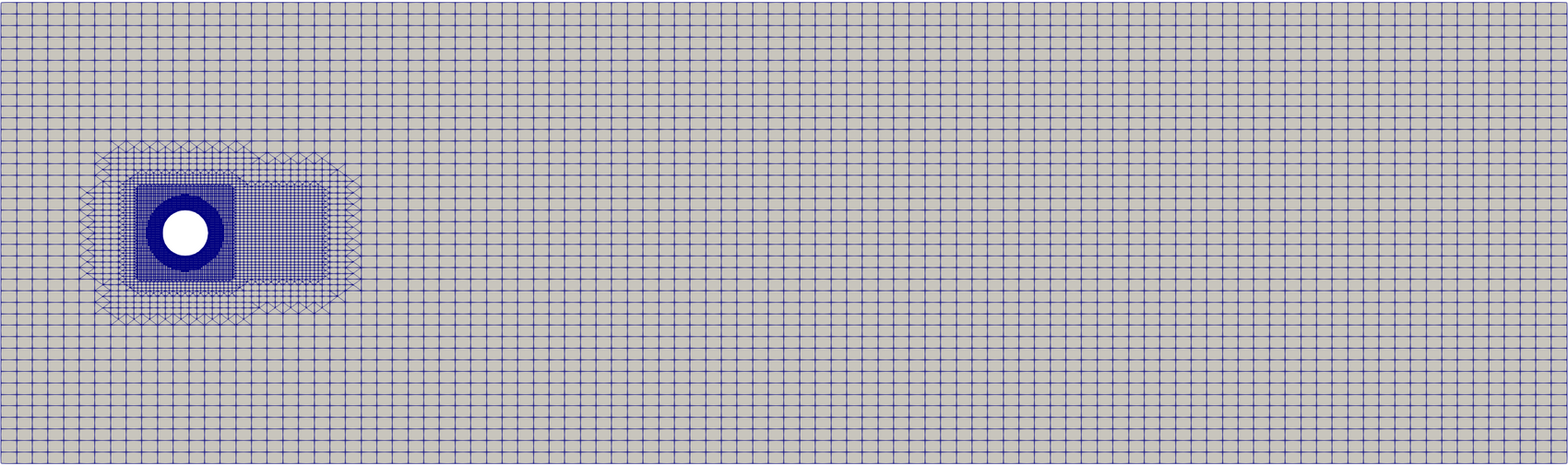};			
		\end{groupplot}
	\end{tikzpicture}
	\caption{A general overview of the unstructured mesh (11644 cells) for cylinder flow.}
	\label{fig:fpc_mesh}
\end{figure}

In this test case, we aim to generate laminar flow over a cylinder at Reynolds number between $Re\in[100,388]$. The physical problem involves simulating an incompressible flow around an 2-D cylinder. The rectangular simulation domain ($[-4,30]\times[-5,5]$) is discretized with an unstructured mesh consisting of 11644 simplex cells, with a cylinder of 0.5 unit radius positioned at the origin and a uniform inlet velocity of 1 unit that remains constant throughout the entire simulation (Figure~\ref{fig:fpc_mesh}). The training and testing datasets are generated using the DNS simulator \cite{weller1998tensorial}, employing the numerical methods described in \cite{stabile2018finite}. The training dataset comprises 25 fluid flow simulations, while the testing dataset consists of 24 simulations, each with 150 macro time-steps and a physical time-step size of $\Delta t = 1$.

\subsubsection{Macro state Identification}
\begin{figure}[htp]
    \centering
    \input{figures/pivotal_nodes.tikz}
    \caption{The reduced graph (1024 nodes) from the original mesh in Figure~\ref{fig:fpc_mesh} using the GNN auto-encoder.
}
    \label{fig:pivotal_nodes}
\end{figure}
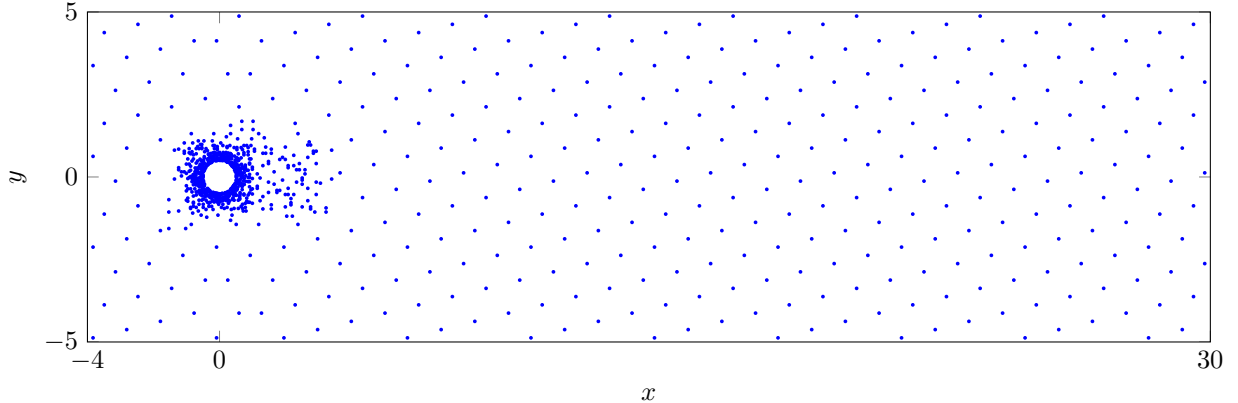

Applying CNNs to unstructured domains results in inefficient data representation and inflates the required computational resources. CNNs are optimized for regular grids, and their application to irregular structures leads to computational overhead and suboptimal performance \cite{geneva2022transformers, vlachas2022multiscale}. Therefore, we employ the GNN auto-encoder as defined in \eqref{eqn:gnn_encdoer}, where $\gamma_{\thetabold_\mathrm{GNN}^*}$ and $\phi_{\thetabold_\mathrm{GNN}^*}$ in each GNN layer are implemented as simple MLPs with 128 neurons and residual connections. The encoder and decoder each consist of three GNN layers. The nodes at the macro level are non-uniformly distributed to effectively capture and resolve the complex flow patterns around the cylinder. Figure~\ref{fig:pivotal_nodes} illustrates the encoded graph for the macro states, where the 1024 nodes with 4-dimensional hidden node feature are arranged as $N_\mathrm{h_r} \times N_\mathrm{w_r} \times N_{\vbm_\mathrm{r}}= 32 \times 32 \times 4$ in \eqref{eqn:macro_seq}. This arrangement allows us to directly utilize the backbone of the diffusion model based on CNN.

\begin{figure}[htp]
    \centering
    \includegraphics[width=0.49\textwidth]{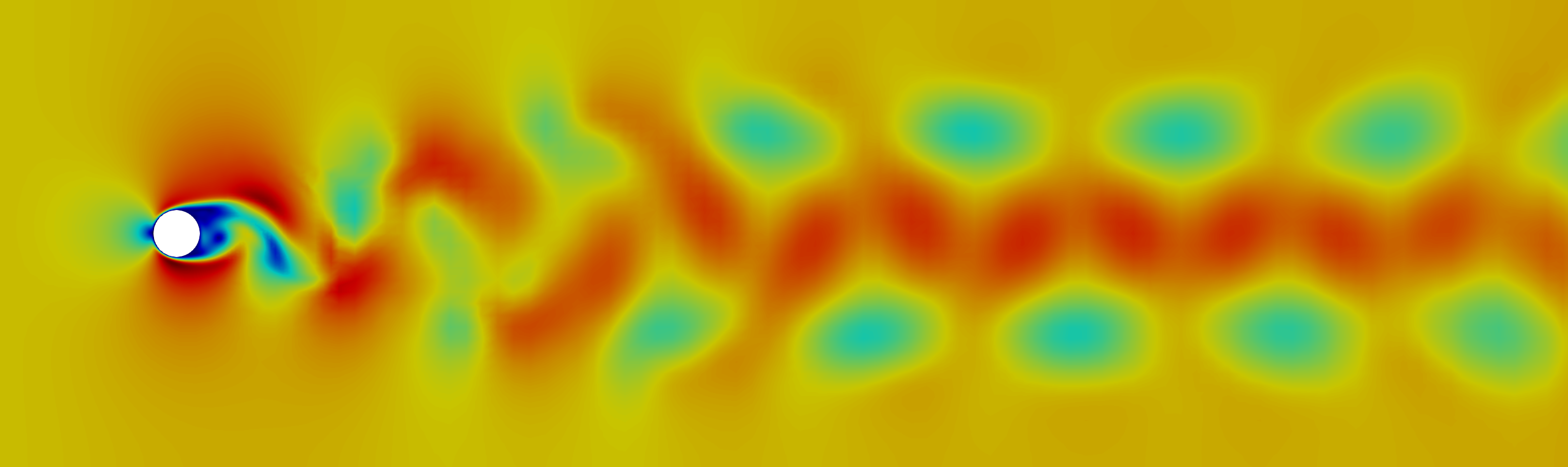}
    \includegraphics[width=0.49\textwidth]{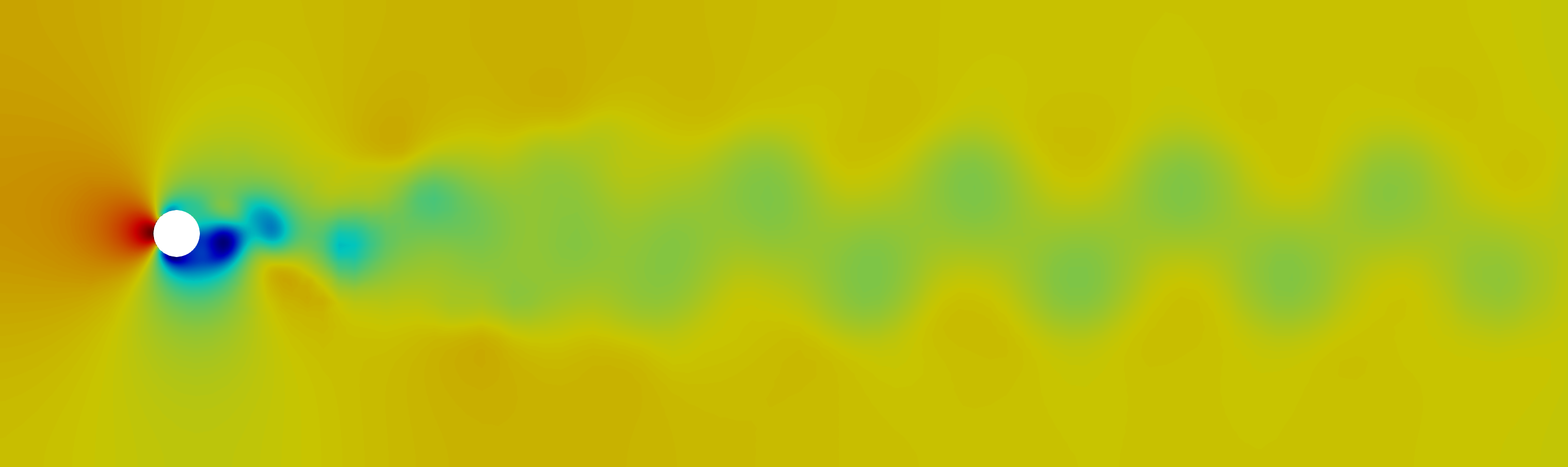}
    \includegraphics[width=0.95\textwidth]{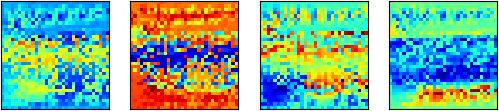}
    \includegraphics[width=0.49\textwidth]{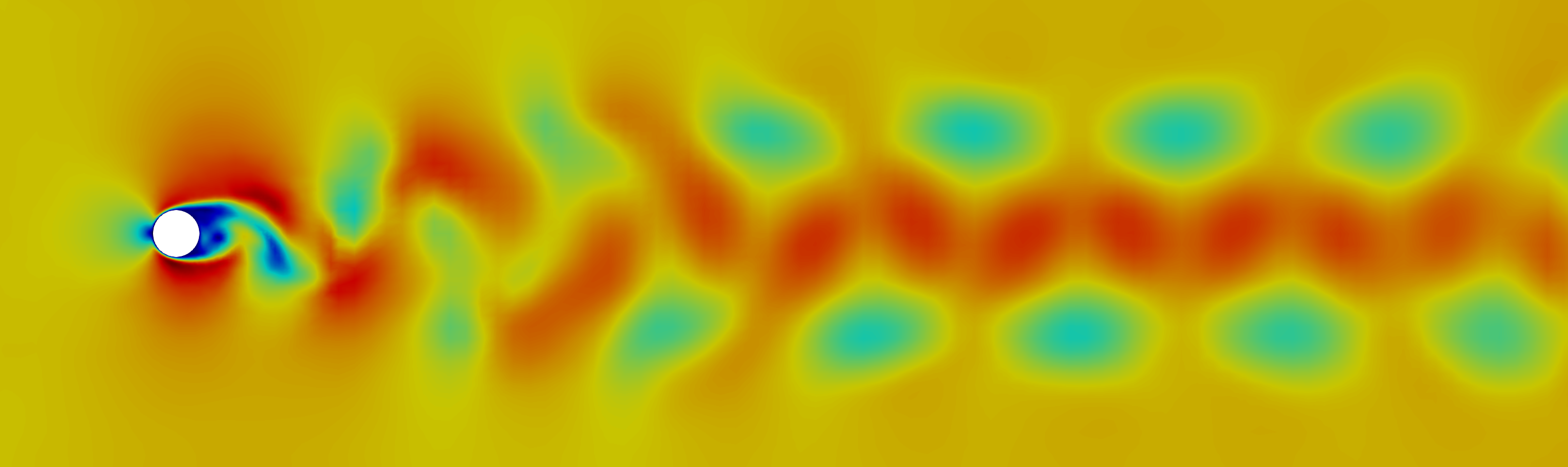}
    \includegraphics[width=0.49\textwidth]{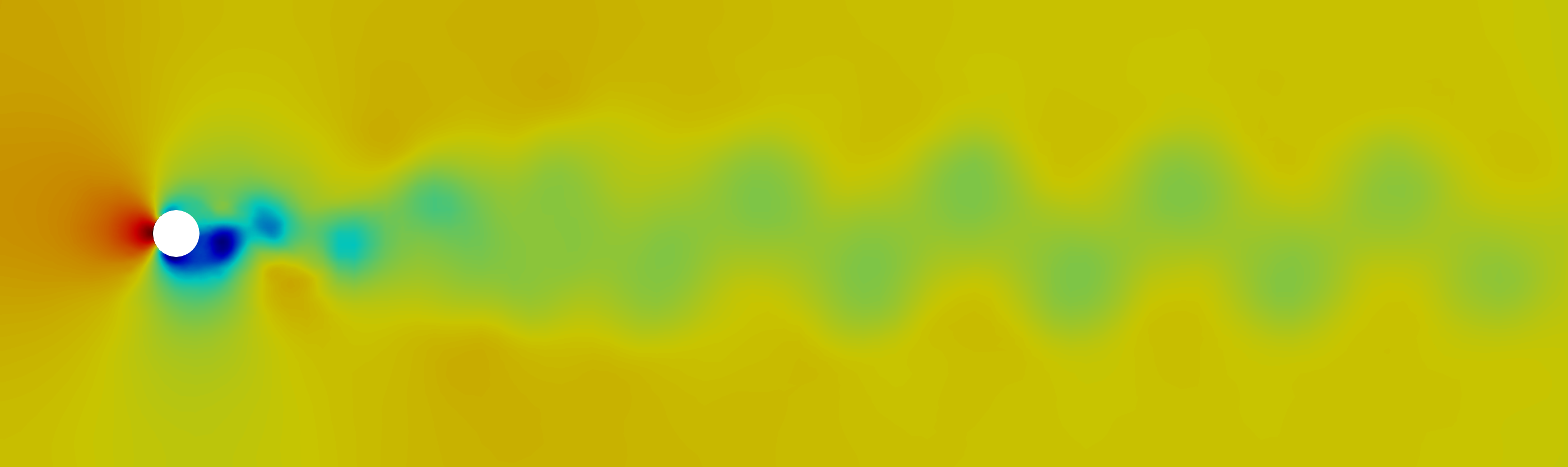}
    \caption{An overview of a micro-level state ($\ubm$), its corresponding macro-level state ($\zbm$) and decoded back to its micro-level ($D_{\thetabold^*_\mathrm{CNN}}(\zbm)$): velocity magnitude (\textit{top row, left}), pressure (\textit{top row, right}),  macro-level state components 1 to 4 (\textit{middle row, from left to right}), and  decoded velocity magnitude (\textit{bottom row, left}), decoded pressure (\textit{bottom row, right}). For visualization purposes, the colorbar is omitted.}
    \label{fig:overviewEdcoder}
\end{figure}

Although the macro-level state lacks a direct physical interpretation, the diffusion model is trained and generates outputs at this level. To provide insight into the appearance of the macro-state, Figure~\ref{fig:overviewEdcoder} presents an example of the micro-level state, its corresponding macro-level state, and the output of the decoder which maps the macro-level state back to the micro-level.

\subsubsection{Generation of Flow Sequences at Different Reynolds-numbers}
\begin{figure}[htp]
	\centering
	\begin{tikzpicture}
		\begin{groupplot}[
			group style={
				group size=2 by 3,
				horizontal sep=0.5cm
			},
			width=0.5\textwidth,
			axis equal image,
			xlabel={$x$},
			ylabel={$y$},
			xtick = {-4, 0, 30},
			ytick = {-5, 0.0, 5},
			xmin=-4, xmax=30,
			ymin=-5, ymax=5
			]
			\nextgroupplot[title={Diffusion model at $Re=106$},xlabel={},xtick=\empty]
			\addplot graphics [xmin=-4, xmax=30, ymin=-5, ymax=5] {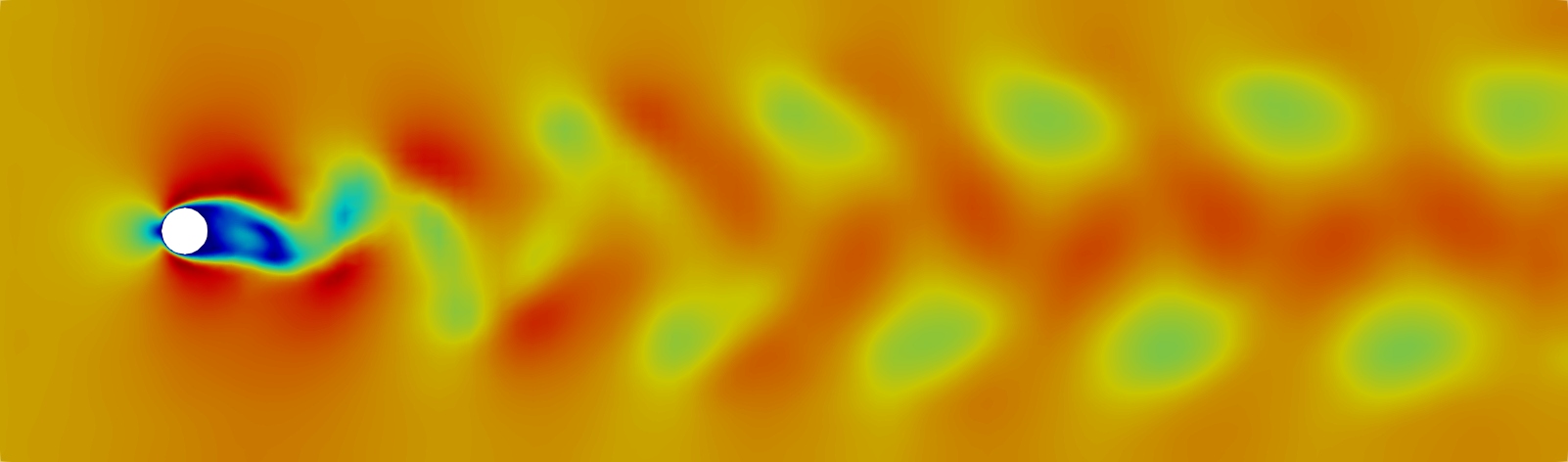};
                \nextgroupplot[title={DNS at $Re=106$},xlabel={},ylabel={},xtick=\empty,ytick=\empty]
			\addplot graphics [xmin=-4, xmax=30, ymin=-5, ymax=5] {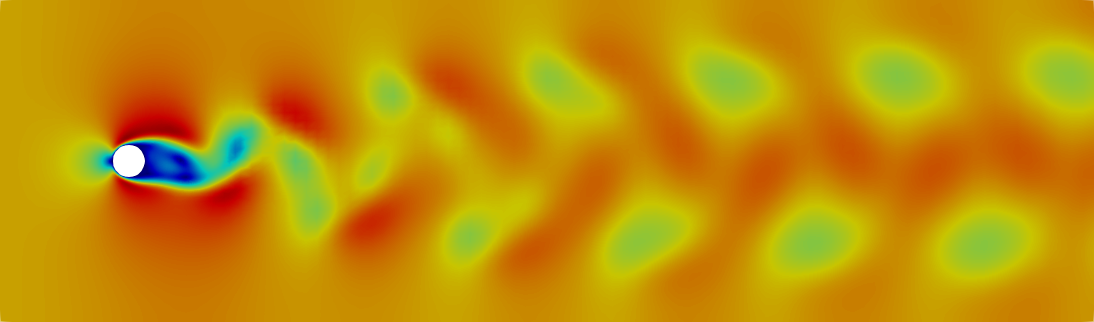};

                \nextgroupplot[title={Diffusion model at $Re=238$},xlabel={},xtick=\empty]
			\addplot graphics [xmin=-4, xmax=30, ymin=-5, ymax=5] {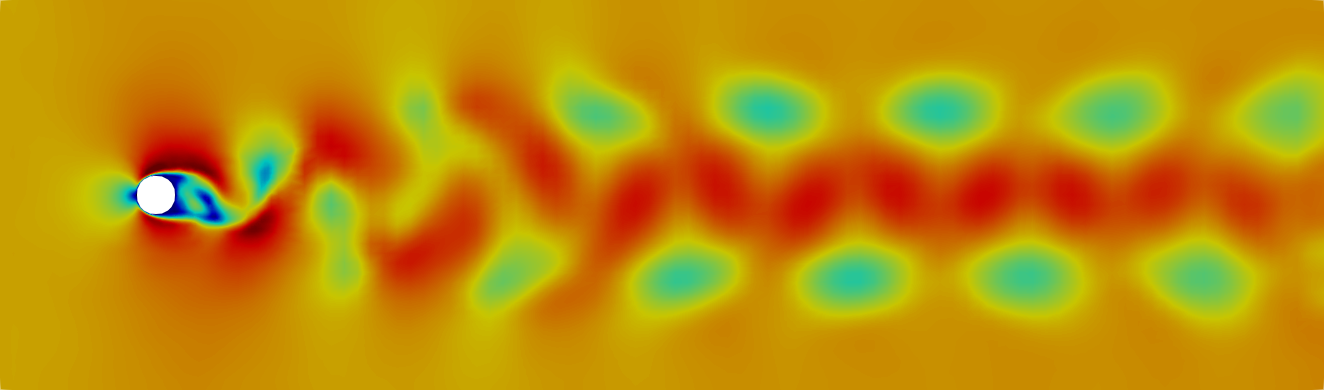};
                \nextgroupplot[title={DNS at $Re=238$},xlabel={},ylabel={},xtick=\empty,ytick=\empty]
			\addplot graphics [xmin=-4, xmax=30, ymin=-5, ymax=5] {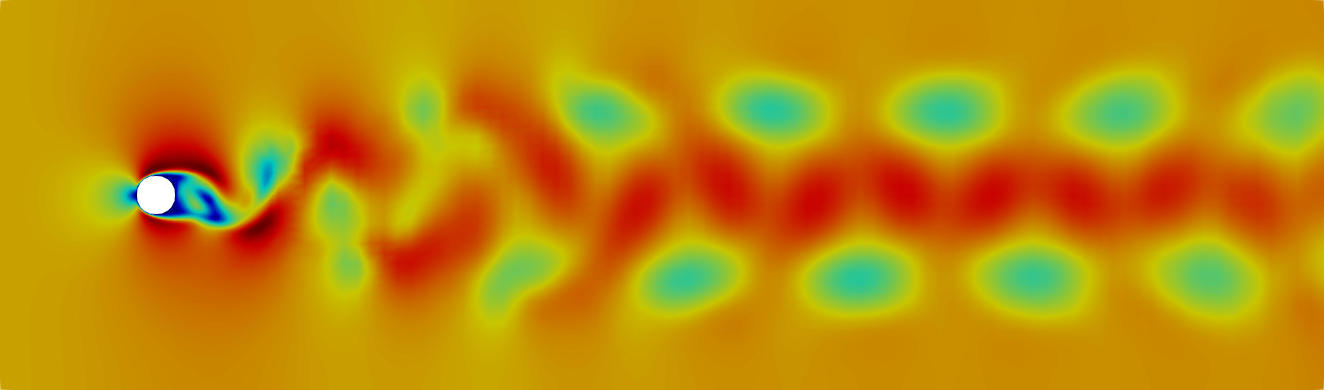};

            \nextgroupplot[title={Diffusion model at $Re=382$}]
			\addplot graphics [xmin=-4, xmax=30, ymin=-5, ymax=5] {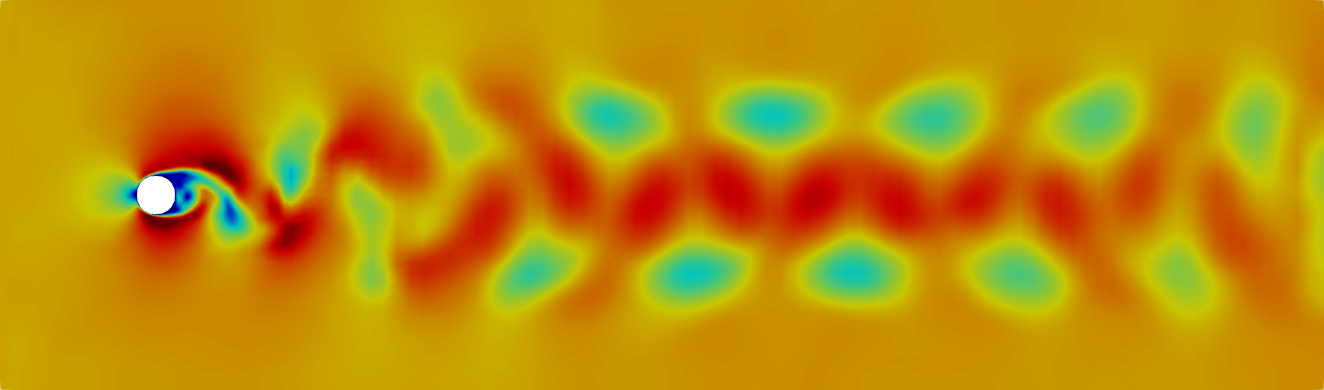};
                \nextgroupplot[title={DNS at $Re=382$},ylabel={},ytick=\empty]
			\addplot graphics [xmin=-4, xmax=30, ymin=-5, ymax=5] {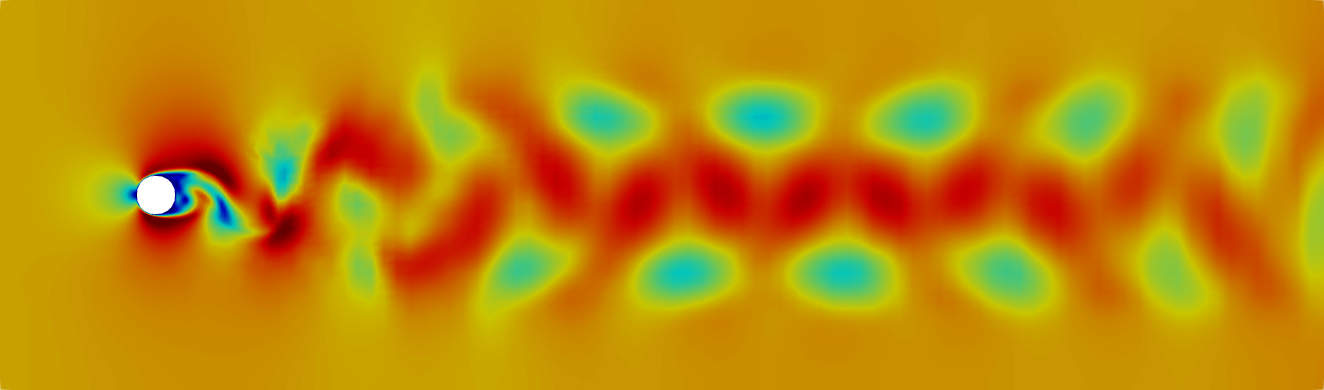};
			
		\end{groupplot}
	\end{tikzpicture}
\colorbarMatlabJet{0}{0.4}{0.8}{1.1}{1.5}
	\caption{Velocity magnitude of the viscous flow at three test Reynolds numbers, as generated by the diffusion model and compared with results from DNS.}
	\label{fig:ns_test_vmag}
\end{figure}

We evaluate the performance of the generative framework across a range of Reynolds numbers not seen during training.  At very low Reynolds numbers, the wake behind the cylinder exhibits longer and more stable tails, indicative of a more laminar flow regime. In contrast, at higher Reynolds numbers, the wake shortens and becomes more complex, with smaller-scale vortices emerging in the flow (Figure~\ref{fig:ns_test_vmag}, \ref{fig:ns_test_pressure}). A speed-up factor of over 500 compared to the simulation cost is achieved. These results highlight the model's ability to accurately reproduce the characteristic flow patterns associated with different Reynolds numbers, thereby validating its robustness and effectiveness in generating fluid flow sequences parameterized by the Reynolds-number.

\subsubsection{Forecasting}
\begin{figure}[htp]
	\centering
	\begin{tikzpicture}
		\begin{groupplot}[
			group style={
				group size=2 by 3,
				horizontal sep=0.5cm
			},
			width=0.5\textwidth,
			axis equal image,
			xlabel={$x$},
			ylabel={$y$},
			xtick = {-4, 0, 30},
			ytick = {-5, 0.0, 5},
			xmin=-4, xmax=30,
			ymin=-5, ymax=5
			]
			\nextgroupplot[title={Guided initial state for dynamics at $Re=106$},xlabel={},xtick=\empty]
			\addplot graphics [xmin=-4, xmax=30, ymin=-5, ymax=5] {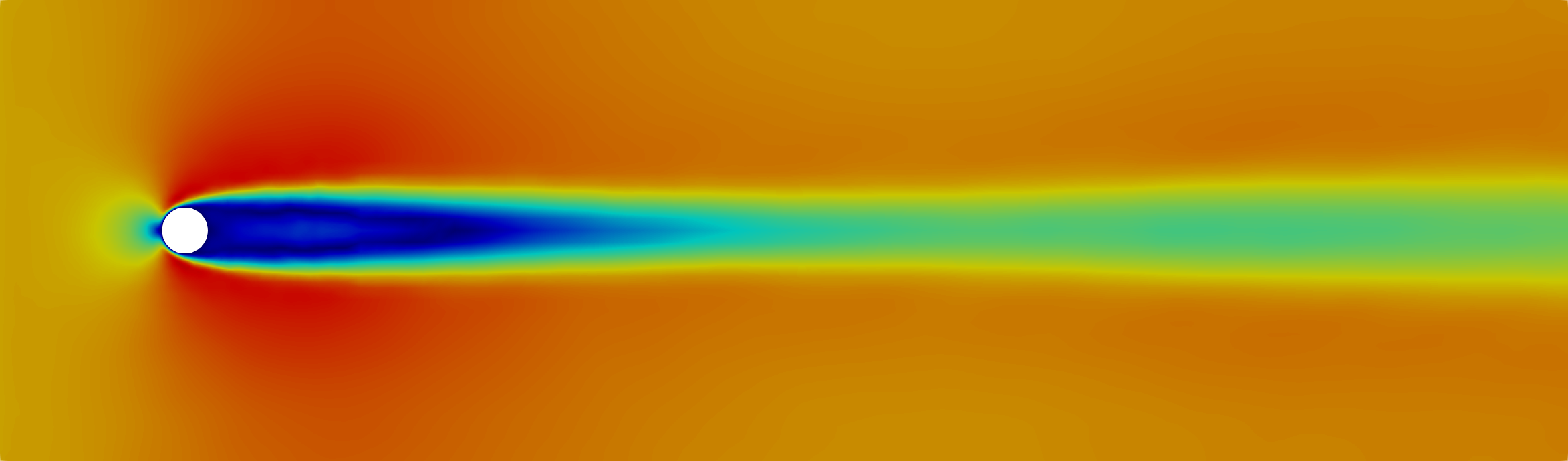};
                \nextgroupplot[title={Initial state for dynamics at $Re=106$},xlabel={},ylabel={},xtick=\empty,ytick=\empty]
			\addplot graphics [xmin=-4, xmax=30, ymin=-5, ymax=5] {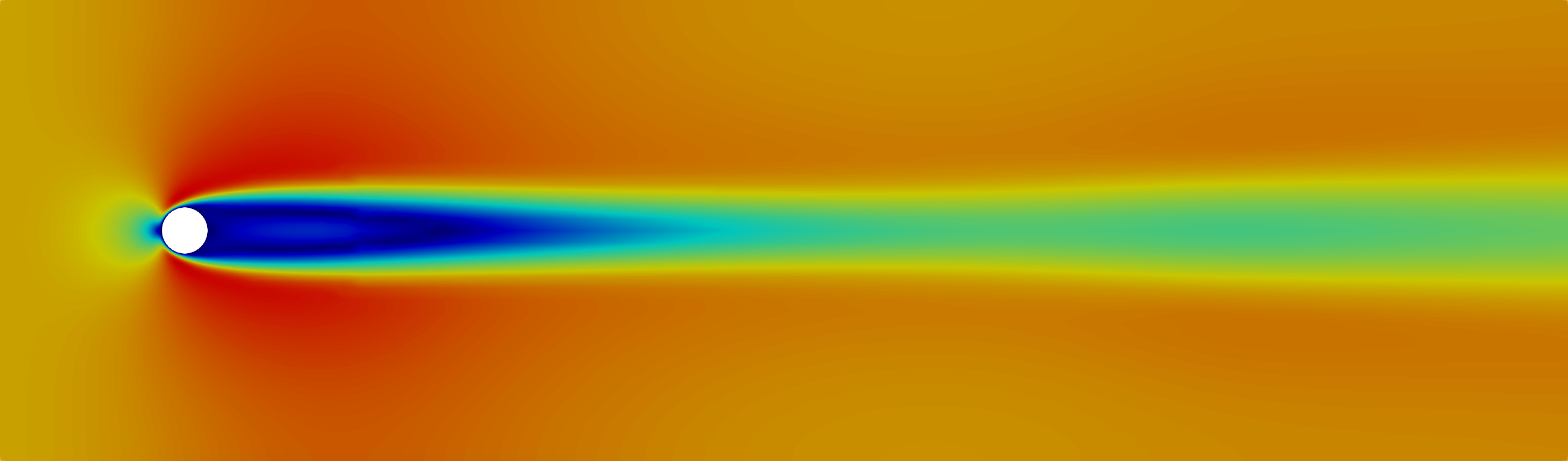};

                \nextgroupplot[title={Guided initial state for dynamics at $Re=238$},xlabel={},xtick=\empty]
			\addplot graphics [xmin=-4, xmax=30, ymin=-5, ymax=5] {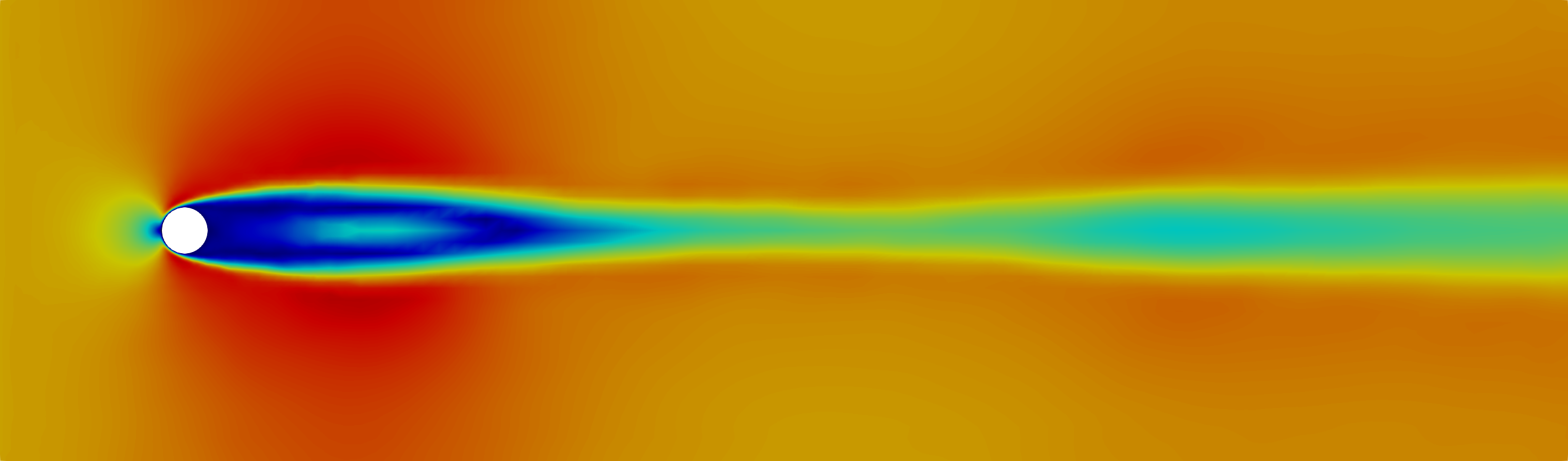};
                \nextgroupplot[title={Initial state for dynamics at $Re=238$},xlabel={},ylabel={},xtick=\empty,ytick=\empty]
			\addplot graphics [xmin=-4, xmax=30, ymin=-5, ymax=5] {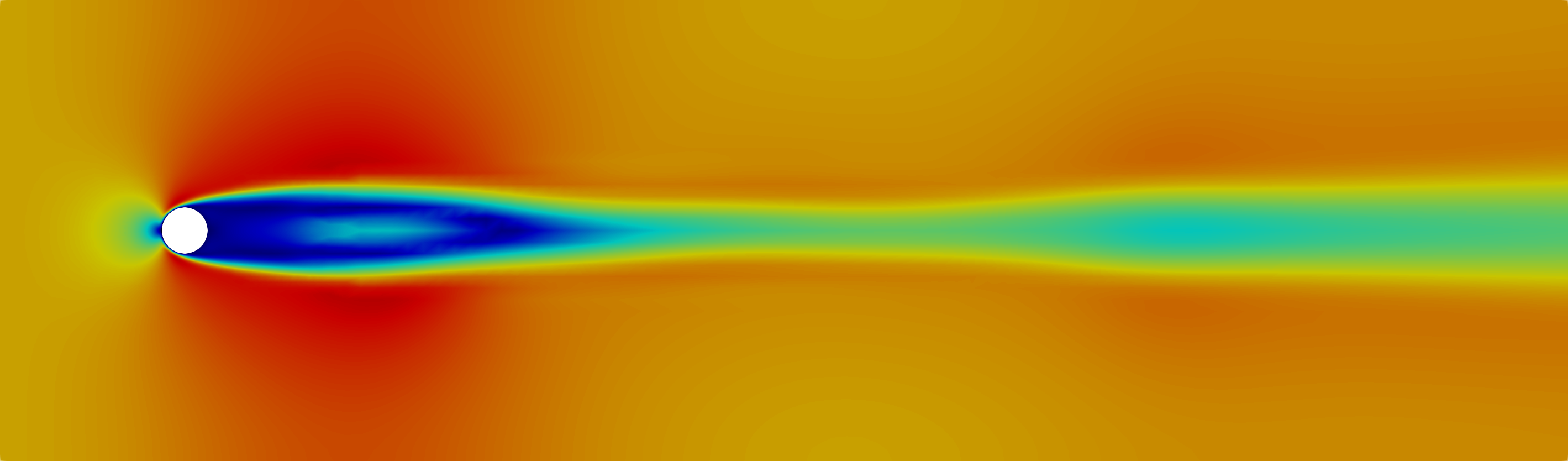};

            \nextgroupplot[title={Guided initial state for dynamics at $Re=382$}]
			\addplot graphics [xmin=-4, xmax=30, ymin=-5, ymax=5] {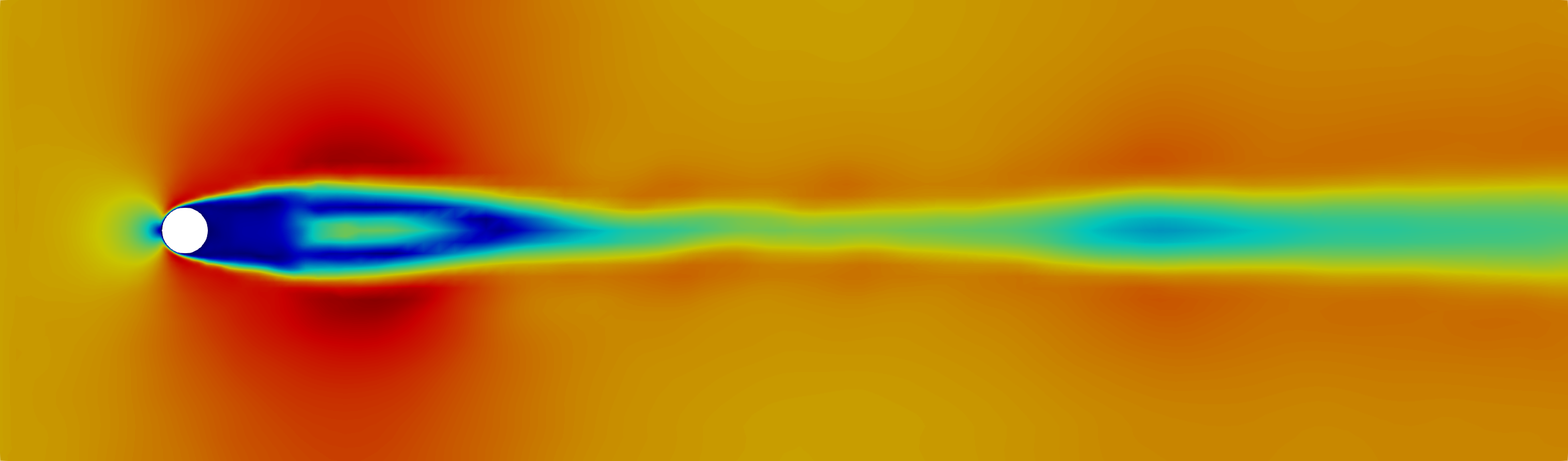};
                \nextgroupplot[title={Initial state for dynamics at $Re=382$},ylabel={},ytick=\empty]
			\addplot graphics [xmin=-4, xmax=30, ymin=-5, ymax=5] {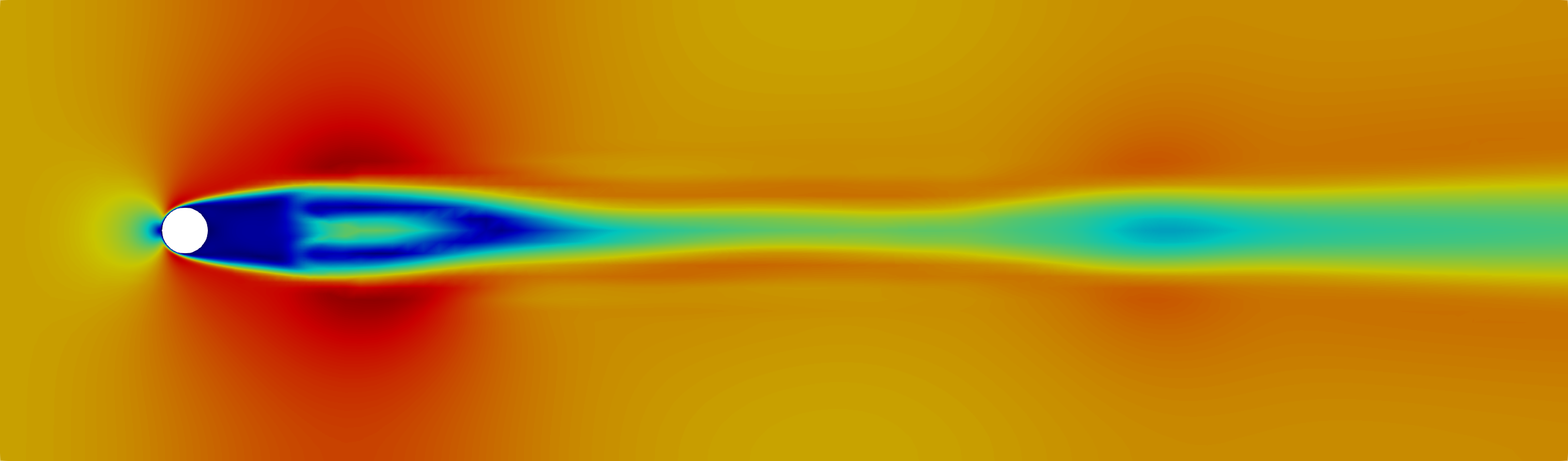};
			
		\end{groupplot}
	\end{tikzpicture}
\colorbarMatlabJet{0}{0.4}{0.8}{1.1}{1.5}
	\caption{Initial velocity magnitude of the viscous flow at three test Reynolds numbers, as generated by the diffusion model and compared with results from DNS.}
	\label{fig:ns_ini_vmag}
\end{figure}

The proposed framework is also capable of making short-term predictions. Given a set of initial states (in this study, we utilize 10 states as the initial conditions) and the corresponding parameters, the generative model demonstrates its robustness by eliminating the need for retraining. This is accomplished by evaluating the encoder and defining the macro-level readout function as simply selecting the first several macro states from the generative sequence. The process of enforcing the initial state leverages guided diffusion to ensure that the generated initial velocity and pressure fields align accurately with the provided initial conditions. Figures~\ref{fig:ns_ini_vmag} and \ref{fig:ns_ini_pressure} illustrate that the guided initial velocity and pressure fields at all tested Reynolds numbers exhibit a high degree of accuracy, closely matching the expected physical states. Following this initialization, the framework proceeds to forecast the evolution of the flow, including critical quantities such as force coefficients. As depicted in Figure~\ref{fig:cylinder_forca_clcd}, the model reasonably predicts the force coefficients across various Reynolds numbers, effectively capturing the transition from the laminar developing state to the vortex shedding state. This includes predicting the symmetry-breaking phenomena inherent in these fluid flow regimes. Thus, the framework's ability to maintain fidelity to the physical system while providing substantial computational efficiency is promising. 

\subsubsection{Sparse reconstruction}
Next, we  perform sparse reconstruction for different test Reynolds numbers. Sparse reconstruction involves using a very limited number of observation points—fewer than the pivotal points used for the encoder (Figure~\ref{fig:cylinder_sparse_obspoints}). To tackle this challenge, we employ the same generative model as in the previous scenario, without any additional retraining, and extend the sequence length to 150 steps to capture the flow dynamics over a longer period. The guidance framework remains consistent with the gradient method, but we redefine the micro readout function to select the sparse observation points from the generated micro states. This involves leveraging the decoder to map these points back to the macro level. By doing so, we aim to ensure that the framework can still effectively reconstruct the flow fields, despite the limited observational data.

\begin{figure}[htp]
	\centering
	\begin{tikzpicture}
		\begin{groupplot}[
			group style={
				group size=2 by 3,
				horizontal sep=0.5cm
			},
			width=0.5\textwidth,
			axis equal image,
			xlabel={$x$},
			ylabel={$y$},
			xtick = {-4, 0, 30},
			ytick = {-5, 0.0, 5},
			xmin=-4, xmax=30,
			ymin=-5, ymax=5
			]
			\nextgroupplot[title={Reconstruction at $t=10$ for $Re=106$},xlabel={},xtick=\empty]
			\addplot graphics [xmin=-4, xmax=30, ymin=-5, ymax=5] {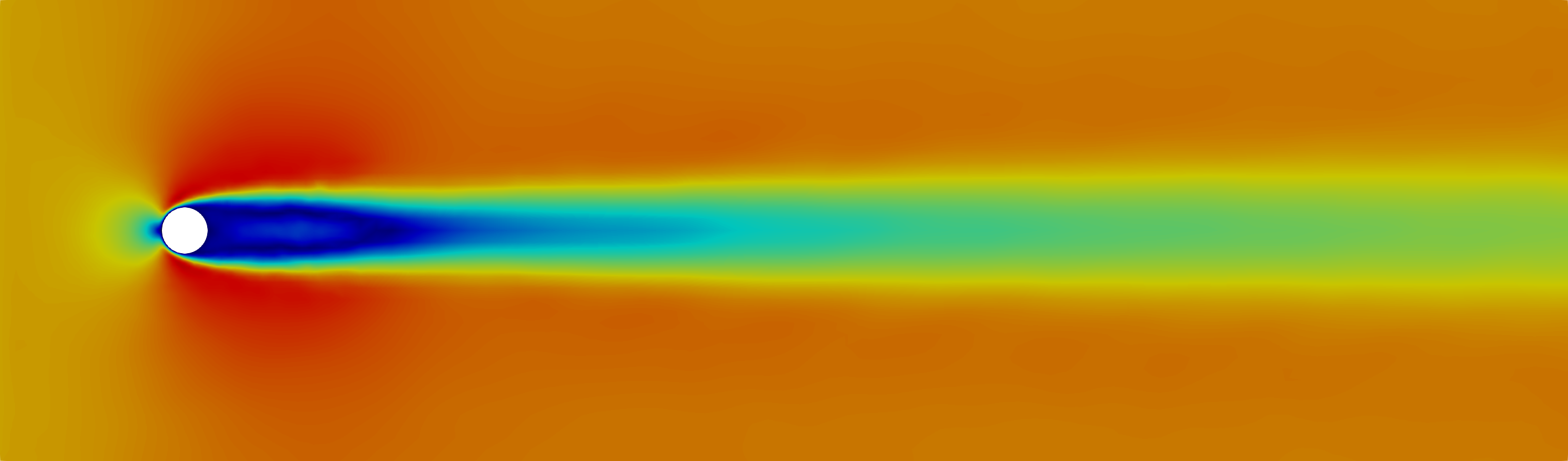};
                \nextgroupplot[title={Truth at $t=10$ for $Re=106$},xlabel={},ylabel={},xtick=\empty,ytick=\empty]
			\addplot graphics [xmin=-4, xmax=30, ymin=-5, ymax=5] {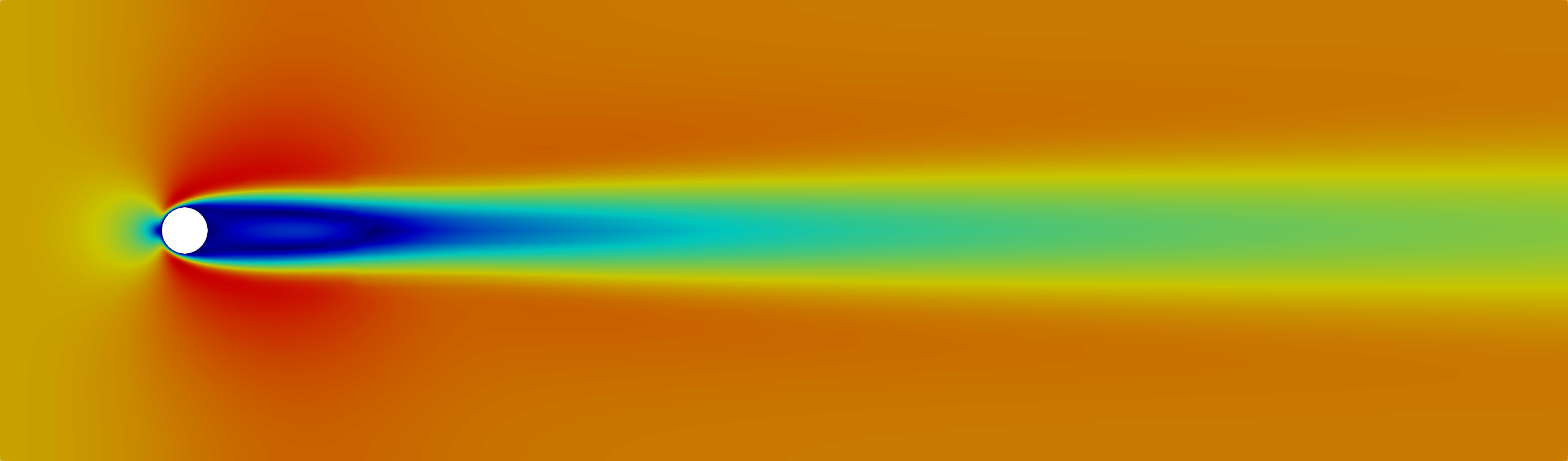};

                \nextgroupplot[title={Reconstruction at $t=10$ for $Re=238$},xlabel={},xtick=\empty]
			\addplot graphics [xmin=-4, xmax=30, ymin=-5, ymax=5] {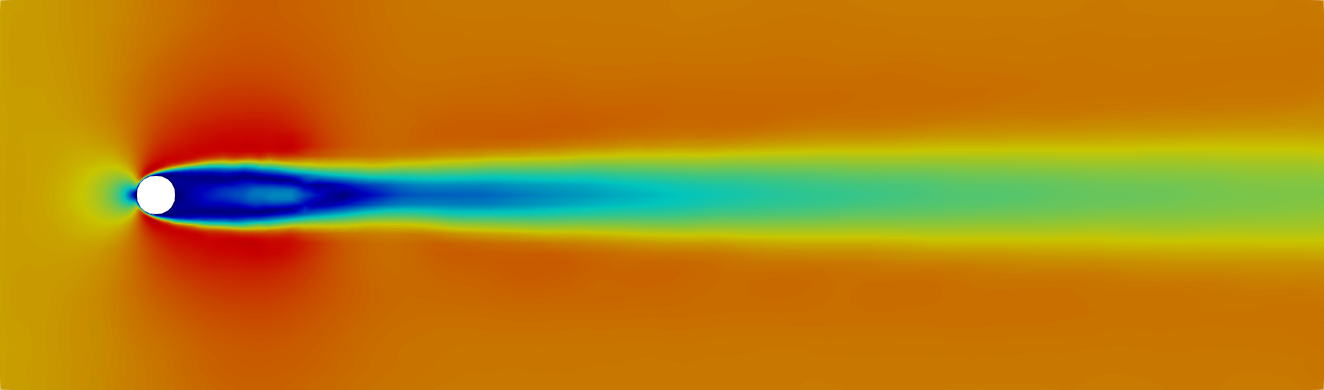};
                \nextgroupplot[title={Truth at $t=10$ for $Re=238$},xlabel={},ylabel={},xtick=\empty,ytick=\empty]
			\addplot graphics [xmin=-4, xmax=30, ymin=-5, ymax=5] {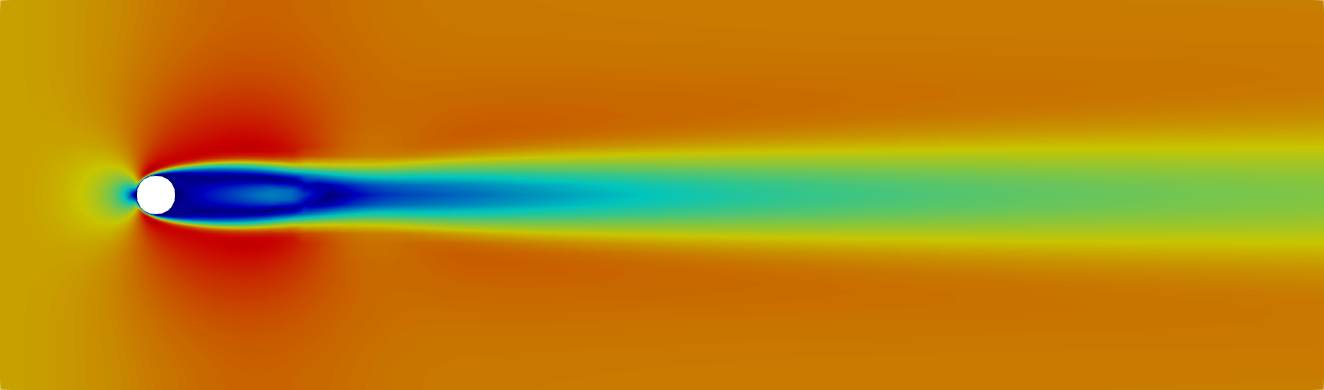};

            \nextgroupplot[title={Reconstruction at $t=10$ for $Re=382$}]
			\addplot graphics [xmin=-4, xmax=30, ymin=-5, ymax=5] {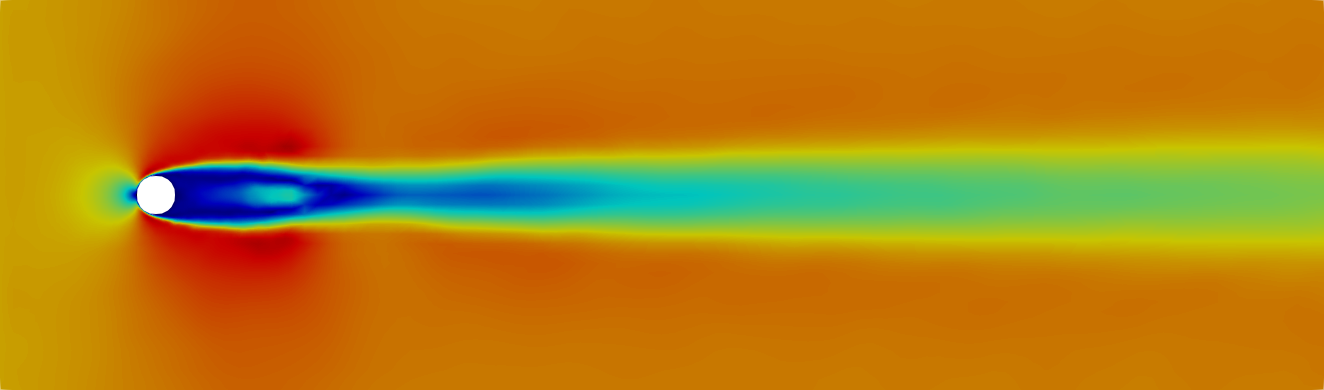};
                \nextgroupplot[title={Truth at $t=10$ for $Re=382$},ylabel={},ytick=\empty]
			\addplot graphics [xmin=-4, xmax=30, ymin=-5, ymax=5] {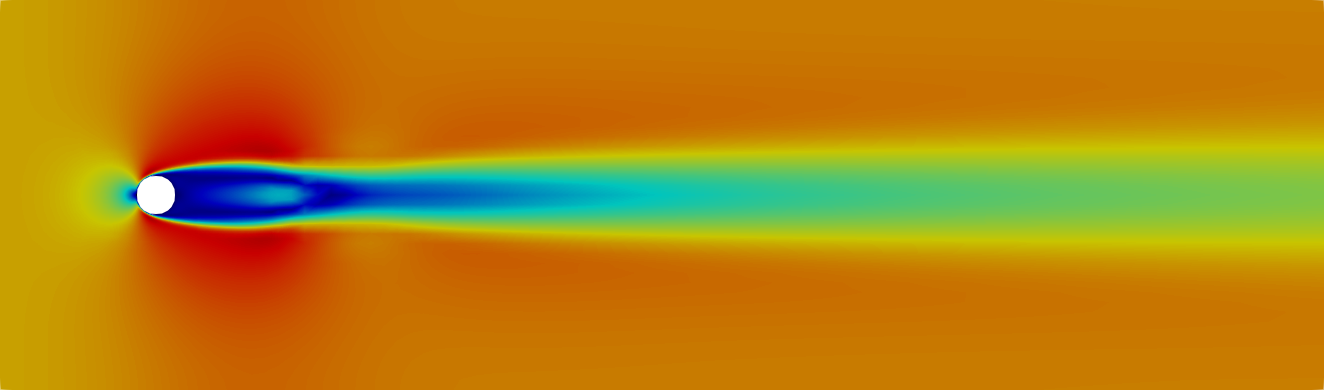};	
		\end{groupplot}
	\end{tikzpicture}
\colorbarMatlabJet{0}{0.4}{0.8}{1.1}{1.5}
	\caption{Velocity magnitude of the viscous flow at three test Reynolds numbers, as reconstructed by the diffusion model and compared with results from DNS.}
	\label{fig:ns_recon_vmag_t10}
\end{figure}

\begin{figure}[htp]
    \centering
    \input{figures/SparseCLCD.tikz}
    \caption{Reconstructed lift coefficients for various Reynolds numbers: comparison between the diffusion model (\ref{line:sparse_cl_diff}) and DNS results (\ref{line:sparse_cl_cfd}).
 }
    \label{fig:Lift_recon}
\end{figure}
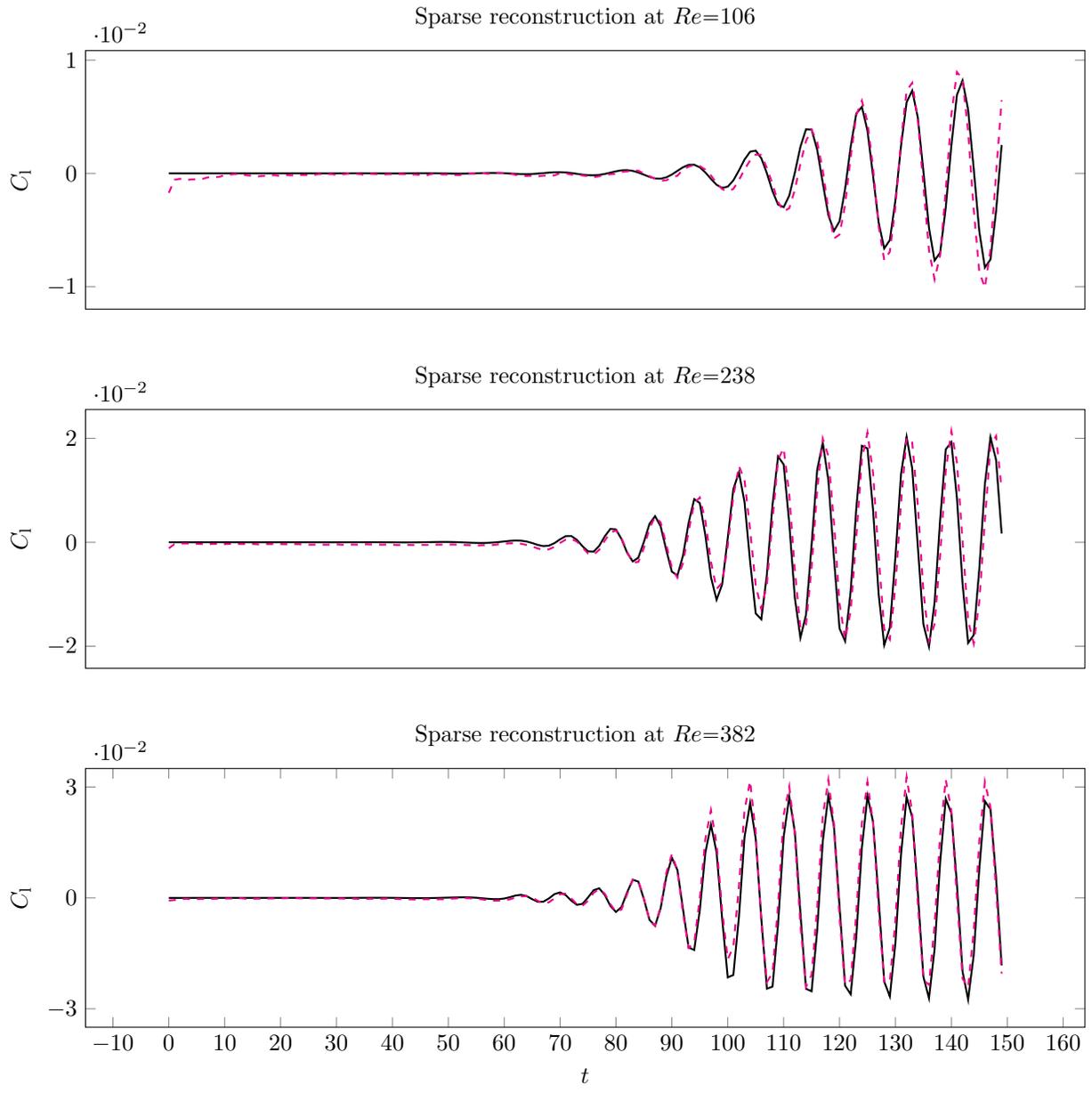

The proposed framework is capable of stably reconstructing the flow field with reasonable accuracy compared to the synthetic truth obtained from DNS. It effectively manages parametric variations that cause significant differences in the flow field from the initial state to the shedding states. Figures~\ref{fig:ns_recon_vmag_t10}, \ref{fig:ns_recon_vmag_t50}, \ref{fig:ns_recon_vmag_t100}, and \ref{fig:ns_recon_vmag_t150} illustrate the flow reconstruction and corresponding mesh at all considered Reynolds numbers. These visualizations confirm that, at all time steps, the reconstructed flow fields align reasonably well with the DNS results. Notably, the zero-mean force coefficient also closely matches the synthetic truth, accurately reflecting the symmetry-breaking phenomenon, as shown in Figure~\ref{fig:Lift_recon}. Overall, the framework's reliably reconstructs the flow fields with high fidelity, despite the sparse observational data. It is particularly beneficial in scenarios where data collection is limited or expensive, enabling accurate flow predictions and analysis. The successful handling of diverse flow conditions further illustrates its potential for other inverse problems in fluid dynamics and related fields.

\subsection{Guided generation with virtual observation for turbulent channel flow}
\label{sec:cflow}
We examine the performance of the framework in turbulent channel flows. Different Reynolds numbers significantly affect the turbulence structures in the spatiotemporal domain. Accurately modeling and predicting turbulence statistics with eddy-resolving simulations require intractable computational resources, positioning the proposed framework as a promising alternative for fast modeling. To this end, the generative framework is applied to learn from high-fidelity DNS data, enabling the efficient generation of realistic turbulent flows at different unseen Reynolds numbers with significant speedup. Our goal is to generate time-coherent, realistic instantaneous velocity fields at cross-sectional (streamwise-wallnormal) spatiotemporal locations. The training data originates from a fully resolved 3D transient DNS of wall-bounded turbulence over a channel, at $Re_\tau \in [180, 605]$. We subsample exclusively during the fully developed phase of flow using a macro time step of $\Delta t = 1$. A simple downsampling (non-trainable) method is applied as the encoder, resulting in spatiotemporal flow sequences at the macro level consisting of 50 snapshots ($\Zbm_{50}(Re_\tau)=[\zbm_0(Re_\tau),...,\zbm_{49}(Re_\tau)]$) with a resolution of $N_{\mathrm{h_r}} \times N_{\mathrm{w_r}} \times N_{\vbm_{\mathrm{r}}} = 32 \times 32 \times 3$ from the micro resolution ($\Ubm_{50}(Re_\tau)=[\ubm_0(Re_\tau),...,\ubm_{49}(Re_\tau)]$) $N_{\mathrm{h}} \times N_{\mathrm{w}} \times N_{\vbm} = 256 \times 256 \times 3$. The focus is on generating sequences of time length equal to 50. During training, 80\% of the database at different $Re_\tau$ is used for training the diffusion model and virtual observations of stress tensor, while the remaining 20\% of unseen $Re_\tau$ is reserved as the test set for conditional generation with virtual observation.

\begin{figure}[htp]
	\centering
	\begin{tikzpicture}
		\begin{groupplot}[
			group style={
				group size=2 by 3,
				horizontal sep=0.5cm
			},
			width=0.5\textwidth,
			axis equal image,
			xlabel={$x$},
			ylabel={$y$},
			xtick = {0, 3.14, 6.28},
			ytick = {0, 1, 2},
			xmin=0, xmax=6.28,
			ymin=0, ymax=2
			]
			\nextgroupplot[title={Diffusion model at $Re_\tau=226$},xlabel={},xtick=\empty]
			\addplot graphics [xmin=0, xmax=6.28, ymin=0, ymax=2] {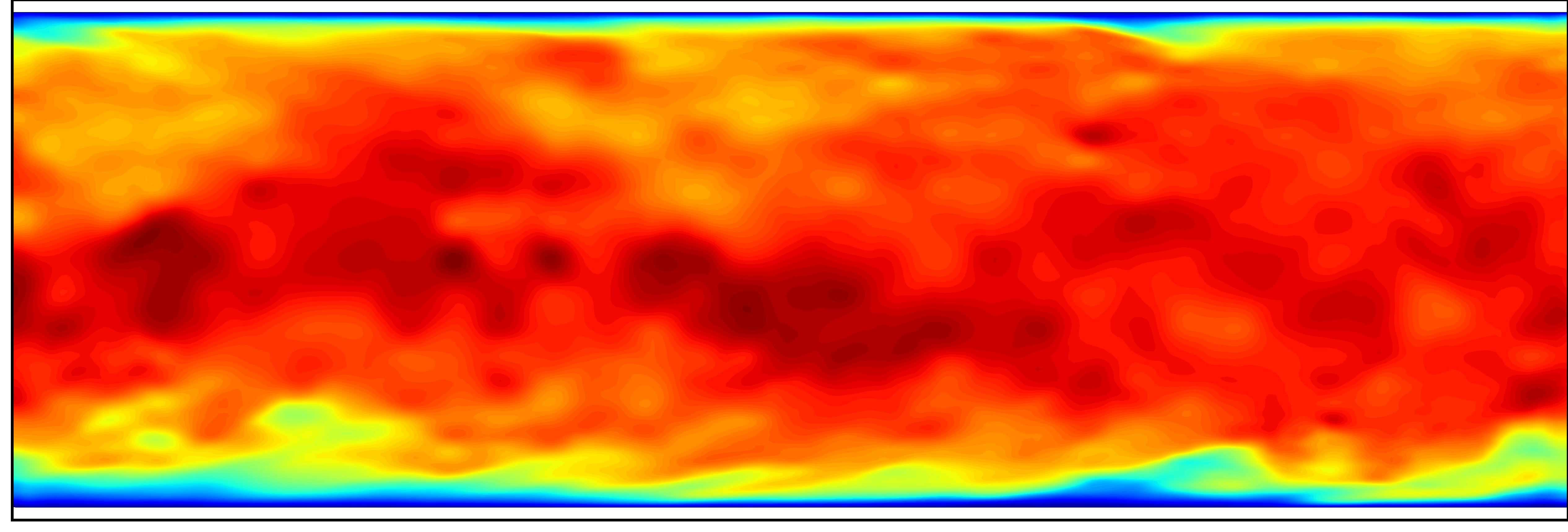};
                \nextgroupplot[title={DNS at $Re_\tau=226$},xlabel={},ylabel={},xtick=\empty,ytick=\empty]
			\addplot graphics [xmin=0, xmax=6.28, ymin=0, ymax=2] {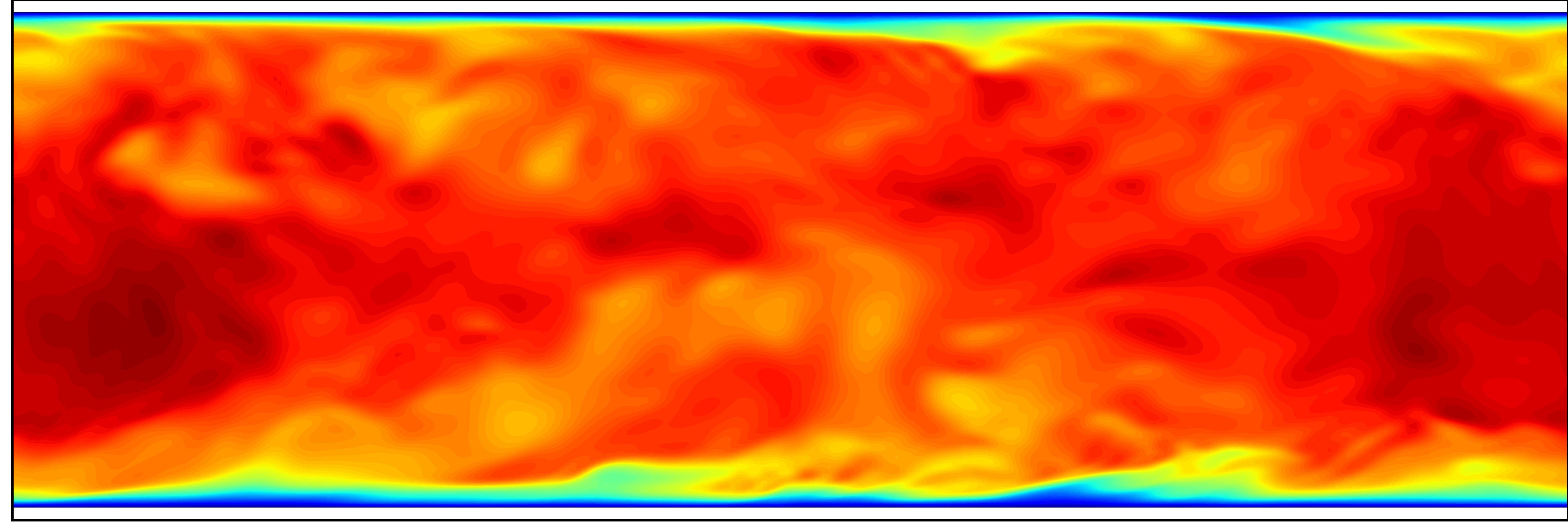};

                \nextgroupplot[title={Diffusion model at $Re_\tau=339$},xlabel={},xtick=\empty]
			\addplot graphics [xmin=0, xmax=6.28, ymin=0, ymax=2] {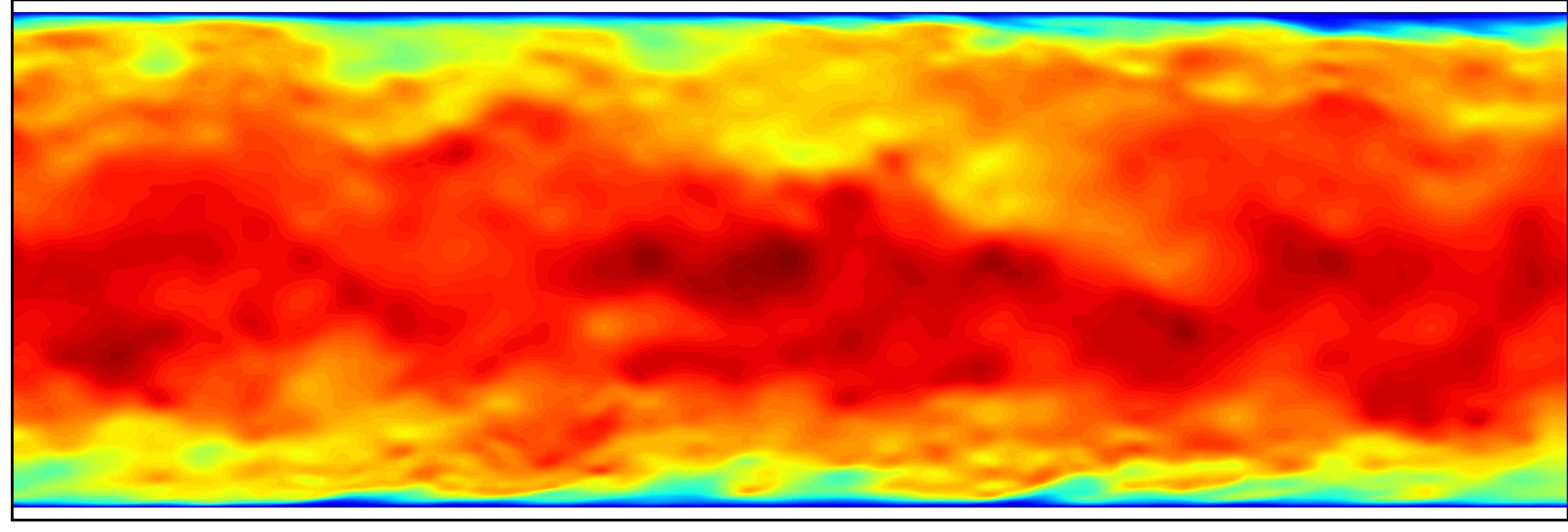};
                \nextgroupplot[title={DNS at $Re_\tau=339$},xlabel={},ylabel={},xtick=\empty,ytick=\empty]
			\addplot graphics [xmin=0, xmax=6.28, ymin=0, ymax=2] {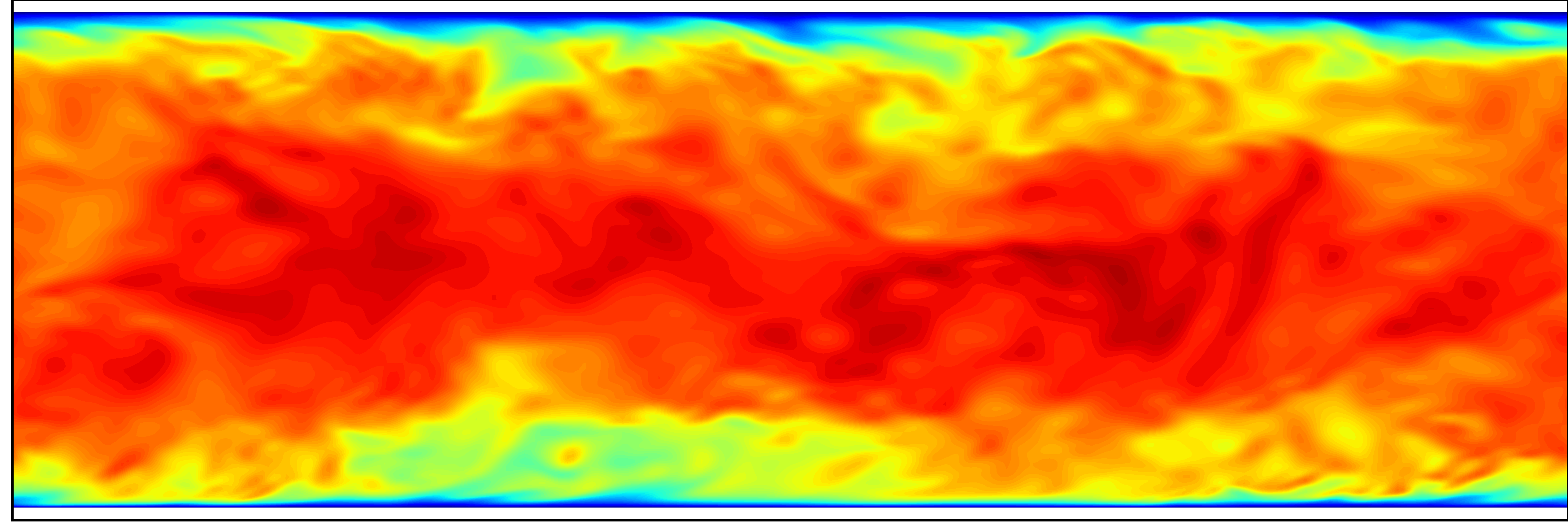};

            \nextgroupplot[title={Diffusion model at $Re_\tau=546$}]
			\addplot graphics [xmin=0, xmax=6.28, ymin=0, ymax=2] {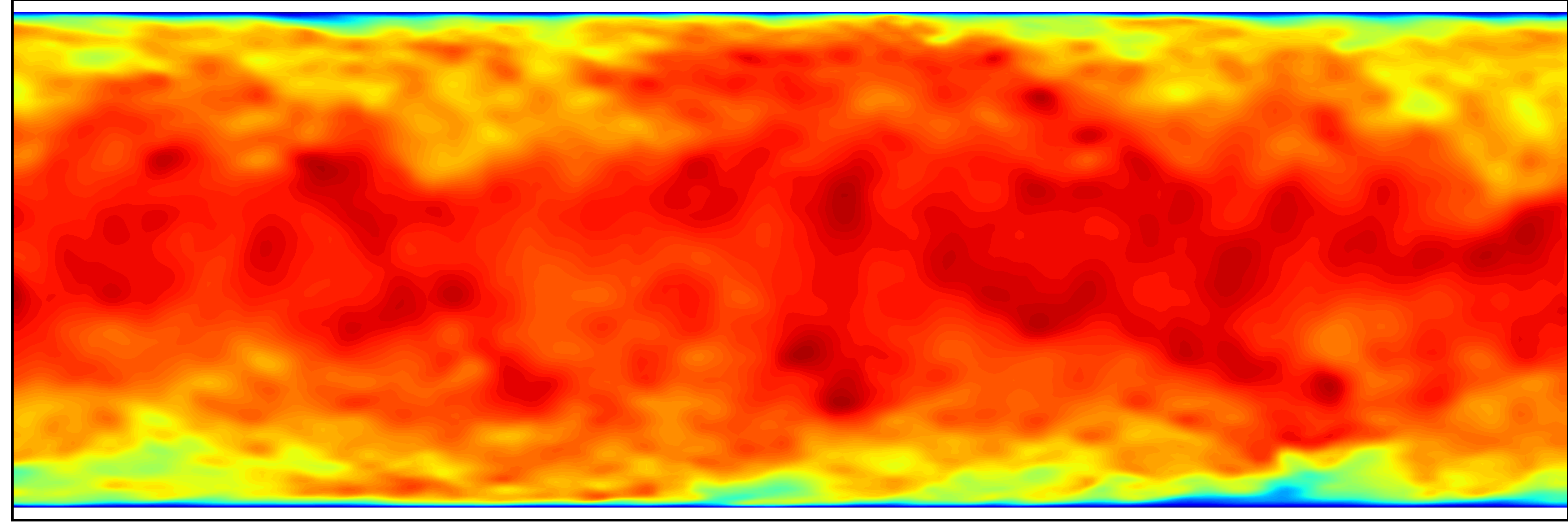};
                \nextgroupplot[title={DNS at $Re_\tau=546$},ylabel={},ytick=\empty]
			\addplot graphics [xmin=0, xmax=6.28, ymin=0, ymax=2] {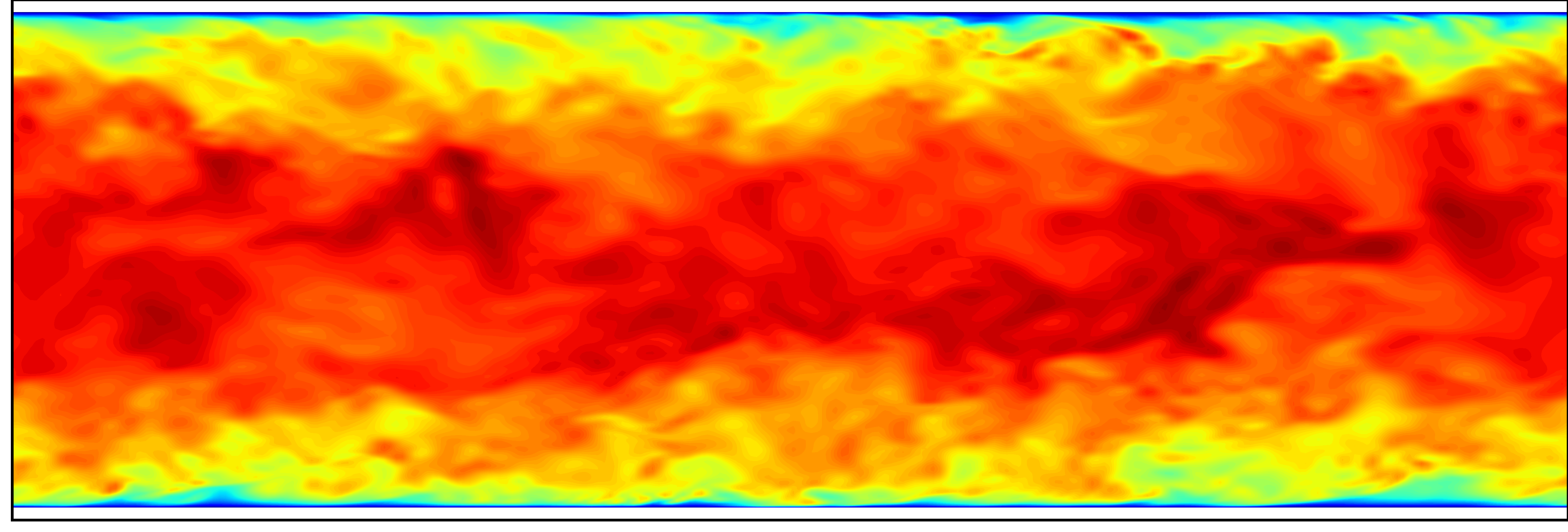};
			
		\end{groupplot}
	\end{tikzpicture}
\colorbarMatlabJet{0}{0.04}{0.08}{0.13}{0.17}
	\caption{Streamwise velocity of the viscous flow at three test Reynolds numbers, as generated by the diffusion model and compared with results from DNS.}
	\label{fig:ns_channel_u_contour}
\end{figure}

Figures~\ref{fig:ns_channel_u_contour}, \ref{fig:ns_channel_v_contour}, and \ref{fig:ns_channel_w_contour} illustrate three instances of flow generated by the proposed framework. These flow trajectory samples are visualized through velocity contours, providing a detailed view of the wall-bounded turbulence characteristics within the domain. Notably, our proposed generative framework demonstrates the capability to effectively generate parametric flow features at previously unseen Reynolds numbers, achieving a computational speedup of over 350 while closely approximating the complex flow fields observed in DNS. As the Reynolds number ($Re_\tau$) increases in turbulent flow, the velocity contours for all three components become more complex and turbulent. Higher $Re_\tau$ leads to increased turbulence intensity, resulting in more pronounced fluctuations in the velocity field. The flow structures become finer and more intricate, with sharper gradients near the wall due to thinner boundary layers. Additionally, enhanced mixing in the flow results in a more uniform velocity distribution away from the wall. Overall, increasing $Re_\tau$ produces more detailed and irregular velocity contours, reflecting the heightened turbulence and energy cascade to smaller scales. Our generative model successfully captures these complex, unseen flow fields.

\begin{figure}[htp]
    \centering
    \input{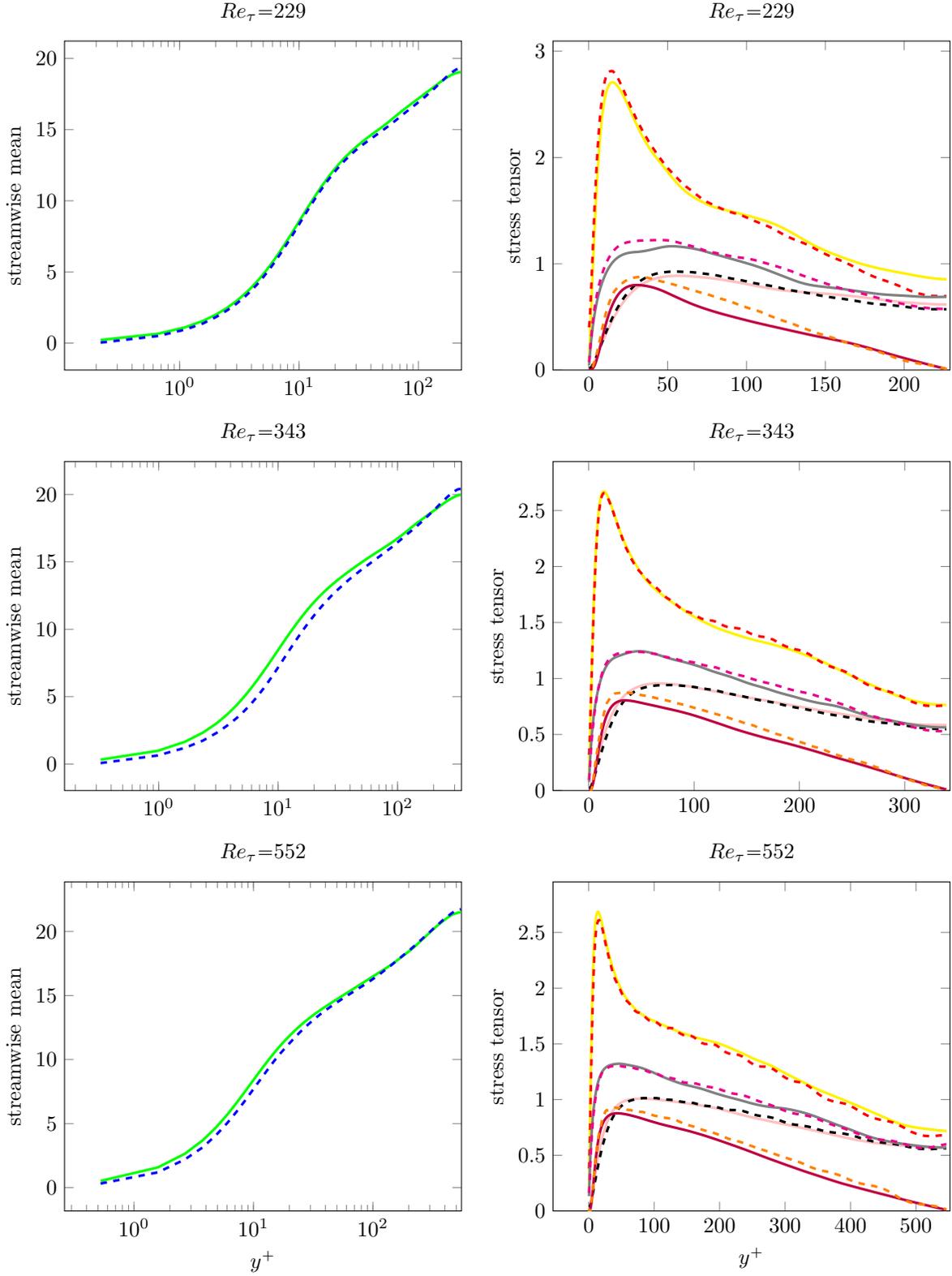}
    \caption{Turbulence statistics: streamwise mean from DNS (\ref{line:idns:mean_u_dns}) and diffusion model (\ref{line:idns:mean_u_diff}); streamwise fluctuation from DNS (\ref{line:idns:urms_dns}) and diffusion model (\ref{line:idns:urms_diff}); wallnormal fluctuation from DNS (\ref{line:idns:vrms_dns}) and diffusion model (\ref{line:idns:vrms_diff}); spanwise fluctuation from DNS (\ref{line:idns:wrms_dns}) and diffusion model (\ref{line:idns:wrms_diff}); non-zero cross fluctuation from DNS (\ref{line:idns:uv_dns}) and diffusion model (\ref{line:idns:uv_diff})
    }
    \label{fig:statistics_mean_rstress}
\end{figure}

The parametric turbulence generation results of the proposed framework are compared with DNS reference in Figure~\ref{fig:statistics_mean_rstress}, illustrating both the fidelity and diversity of the diffusion-generated spatiotemporal velocity field samples. For this assessment, an ensemble of 256 flow sequences, each with 50 snapshots, was synthesized to ensure statistical convergence. As shown in the figure, the turbulence statistics obtained from our model are in reasonably good agreement with those obtained by DNS. We note that these results were obtained for  untrained Reynolds numbers ($Re_\tau$). The mean streamwise velocity profile generated matches the DNS reasonably well, reflecting the expected behavior across the linear viscous sublayer. As the Reynolds number ($Re_\tau$) increases, the mean velocity profile versus $y^+$ (the dimensionless wall-normal distance) shows a more pronounced logarithmic region, indicating higher velocities away from the wall due to increased turbulence. The viscous sublayer becomes thinner, and the buffer layer shifts, reflecting stronger velocity gradients near the wall. Velocity fluctuations for wallnormal and spanwise velocity versus $y^+$ also increase with higher $Re_\tau$, particularly in the near-wall region, due to intensified turbulent activity. These fluctuations extend further into the log layer, indicating more vigorous mixing and energy transfer across the flow. Overall, higher $Re_\tau$ results in steeper mean velocity gradients near the wall and increased turbulence intensity throughout the boundary layer, leading to more complex and energetic flow characteristics. The relative good match in the figure shows that the generative framework is able to parametrically capture these statistics accurately.

\begin{figure}[htp]
    \centering
    \input{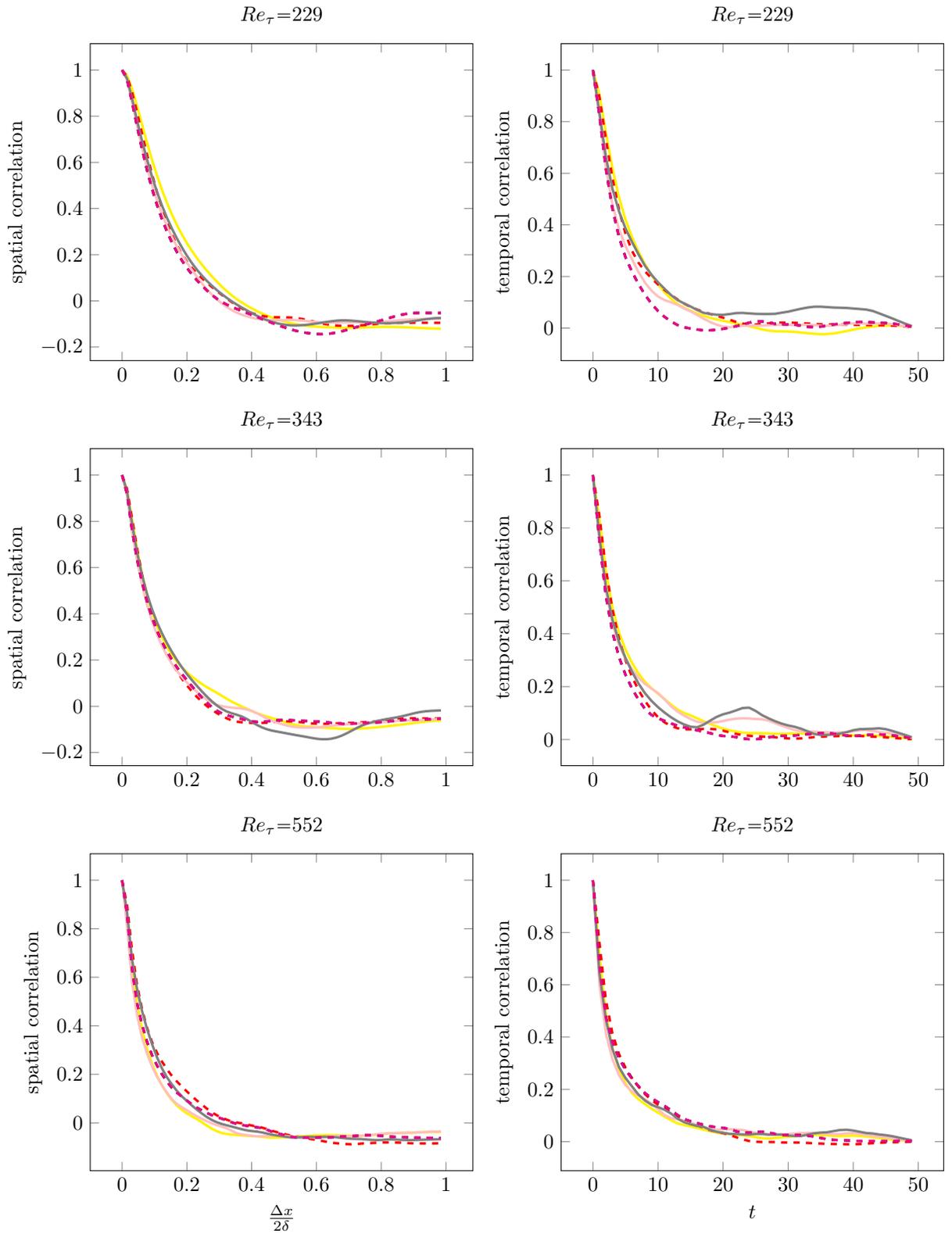}
    \caption{ Streamwise spatial and temporal correlation at $y^+ = 20$ from DNS (\ref{line:idns:dns_y20}) and diffusion model (\ref{line:idns:diff_y20}); at $y^+ = 50$ from DNS (\ref{line:idns:dns_y50}) and diffusion model (\ref{line:idns:diff_y50}); at $y^+ = 100$ from DNS (\ref{line:idns:dns_y100}) and diffusion model (\ref{line:idns:diff_y100}).
    }
    \label{fig:correlations0}
\end{figure}

Figures~\ref{fig:correlations0}, \ref{fig:correlations1}, and \ref{fig:correlations2} present the spatial and temporal correlations at different unseen $Re_\tau$ compared with DNS data. Although not perfect, the generated samples exhibit a correlation decay trend similar to that observed in the DNS data. As the Reynolds number ($Re_\tau$) increases in channel flow, both temporal and spatial correlations undergo significant changes. Temporally, the correlation time scales become shorter due to the increased turbulence intensity, resulting in more rapid decorrelation of flow structures. This behavior reflects the faster dynamics and more frequent changes in the turbulent flow field. Spatially, the correlation lengths decrease, particularly in the wall-normal and spanwise directions. The increased Reynolds number leads to finer turbulent structures and more complex flow patterns, causing the flow to decorrelate over shorter distances. Additionally, the enhanced mixing and energy transfer at higher $Re_\tau$ contribute to the more rapid loss of spatial coherence in the velocity field. Overall, the generative framework effectively captures the relationship between higher $Re_\tau$ values and the resulting shorter temporal and spatial correlations, thereby indicating more dynamic and spatially complex turbulence.

\begin{figure}[htp]
    \centering
    \input{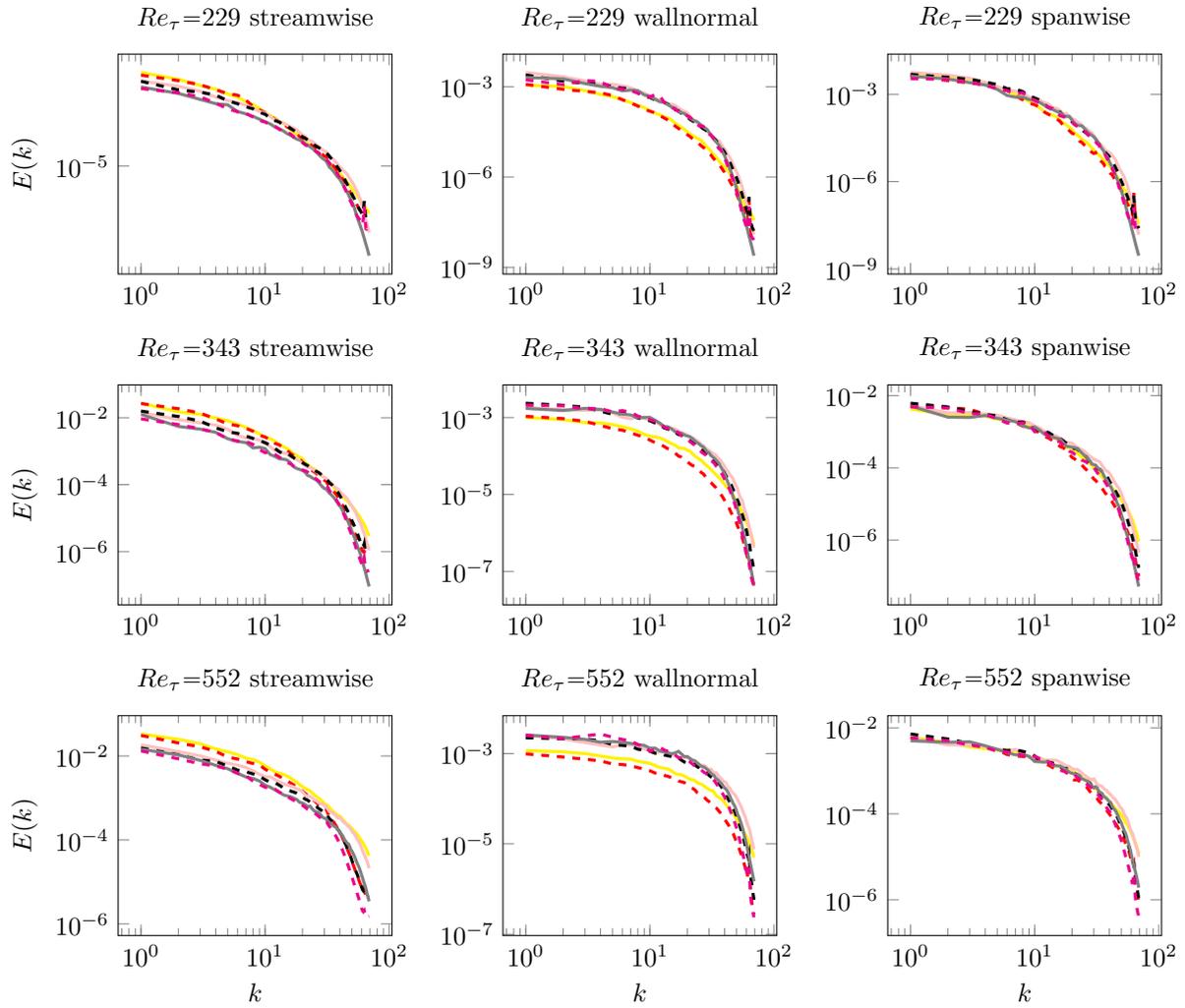}
    \caption{Energy spectrum of velocity at $y^+ = 20$ from DNS (\ref{line:idns:Edns_y20}) and diffusion model (\ref{line:idns:Ediff_y20}); at $y^+ = 50$ from DNS (\ref{line:idns:Edns_y50}) and diffusion model (\ref{line:idns:Ediff_y50}); at $y^+ = 100$ from DNS (\ref{line:idns:Edns_y100}) and diffusion model (\ref{line:idns:Ediff_y100}).
    }
    \label{fig:energy_spectrum}
\end{figure}

Figures~\ref{fig:energy_spectrum} illustrate the energy spectrum for generated sequences and DNS data. Our generative framework successfully captures the changes in the energy spectrum as the Reynolds number ($Re_\tau$) increases. In the wall-normal and spanwise direction, the energy spectrum shows a shift towards higher frequencies, indicating the presence of smaller-scale turbulent structures. This shift is attributed to the increased turbulence intensity, leading to a broader distribution of energy across various scales. The inertial subrange extends, demonstrating a more efficient energy cascade that transfers energy from larger to smaller scales more rapidly. Overall, higher $Re_\tau$ results in a more extensive range of turbulent scales and a more detailed and energetic velocity field, as evidenced by the energy spectrum in both the wall-normal and spanwise directions.

\section{Conclusion}
\label{sec:conclu}
In this work, we present a novel generative learning framework for parametric high-dimensional systems using diffusion models with gradient guidance and virtual observations. Our approach addresses the complexities and computational challenges inherent in simulating large-scale nonlinear systems with high parametric dependencies. The framework employs various encoder-decoder architectures to elevate the generative learning capabilities of diffusion models to the macro level. By integrating both micro and macro-level information, along with virtual observations, our framework achieves versatility and effectiveness in generating complex parametric systems.

We demonstrated the effectiveness of our framework through two case studies: incompressible, two-dimensional laminar flow past a cylinder on an unstructured mesh and incompressible turbulent channel flow on a structured mesh, both parameterized by the Reynolds number. The results illustrate the robustness of the framework across different settings, showcasing its capability to generate flow sequences directly across different parameters. By utilizing multi-level information to guide the generative process, the present framework enhances the fidelity and accuracy of the generated flow sequences. In the flow past a cylinder case, it successfully captured the characteristic flow patterns associated with different Reynolds numbers. In the more challenging turbulent channel flow scenario, the framework accurately generated dynamics that matched DNS results, reflecting the complex and energetic characteristics of turbulent flows at varying Reynolds numbers and underscoring the model's capability to manage high-dimensional and dynamic turbulence phenomena.

Overall, the proposed generative framework showed promising results in efficiently and accurately modeling high-dimensional parametric systems. Its ability to integrate gradient guidance and virtual observations allows it to adapt to different scenarios and parameter variations without the need for retraining. This flexibility and efficiency make it a valuable tool for various applications in fluid dynamics and other fields requiring the solution of high-dimensional, parametric PDEs.

\appendix
\newpage
\section{Additional results for cylinder flow}

\begin{figure}[htp]
	\centering
	\begin{tikzpicture}
		\begin{groupplot}[
			group style={
				group size=2 by 3,
				horizontal sep=0.5cm
			},
			width=0.5\textwidth,
			axis equal image,
			xlabel={$x$},
			ylabel={$y$},
			xtick = {-4, 0, 30},
			ytick = {-5, 0.0, 5},
			xmin=-4, xmax=30,
			ymin=-5, ymax=5
			]
			\nextgroupplot[title={Diffusion model at $Re=106$},xlabel={},xtick=\empty]
			\addplot graphics [xmin=-4, xmax=30, ymin=-5, ymax=5] {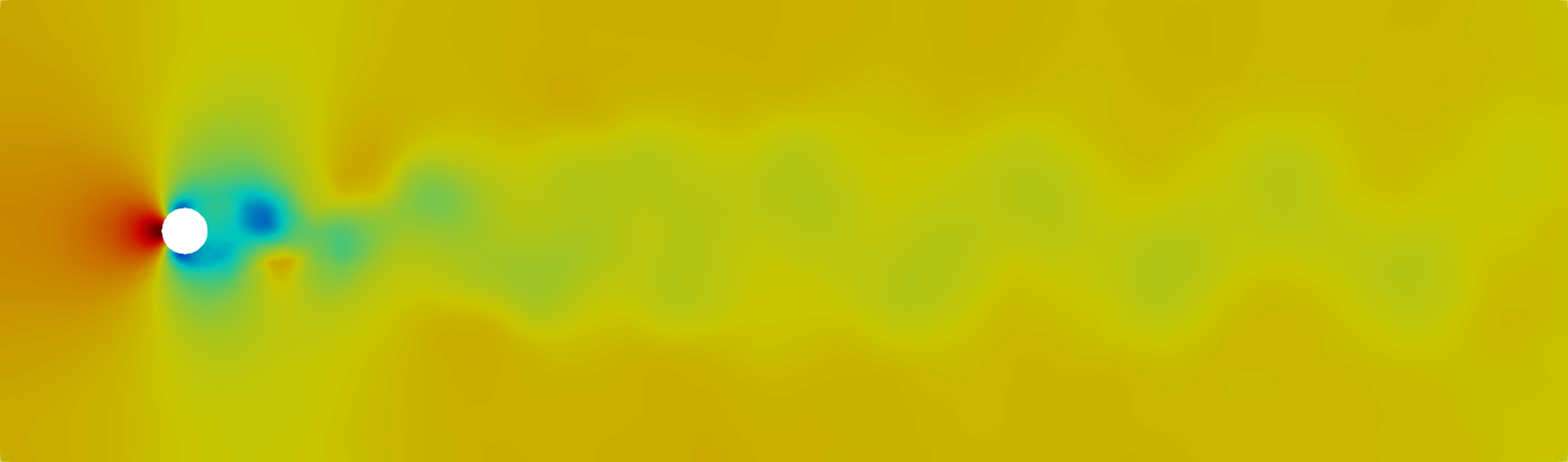};
                \nextgroupplot[title={DNS at $Re=106$},xlabel={},ylabel={},xtick=\empty,ytick=\empty]
			\addplot graphics [xmin=-4, xmax=30, ymin=-5, ymax=5] {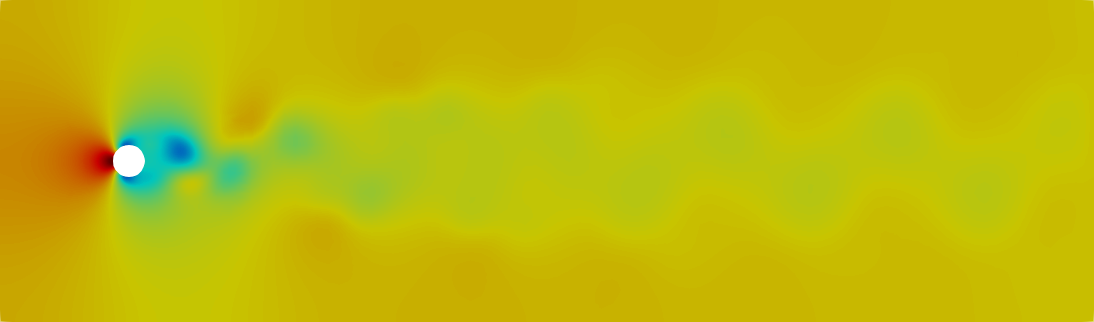};

                \nextgroupplot[title={Diffusion model at $Re=238$},xlabel={},xtick=\empty]
			\addplot graphics [xmin=-4, xmax=30, ymin=-5, ymax=5] {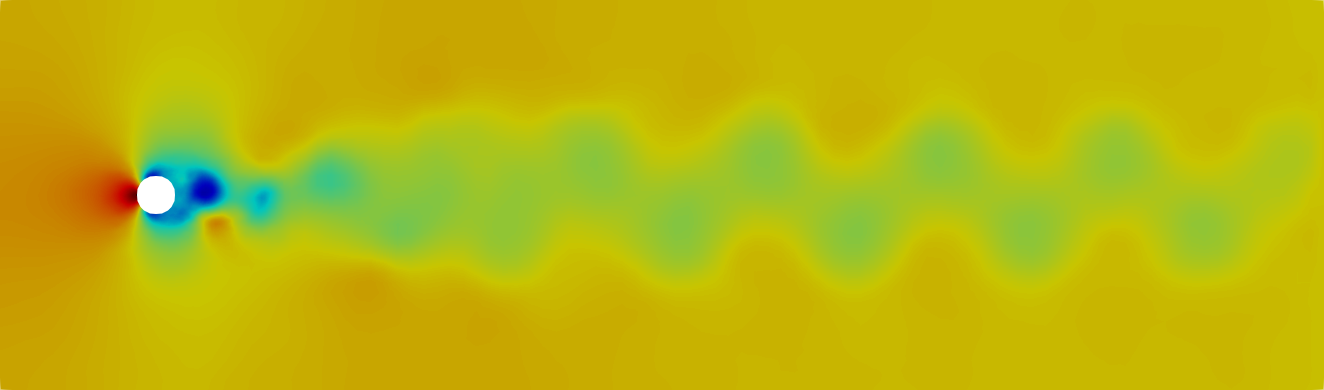};
                \nextgroupplot[title={DNS at $Re=238$},xlabel={},ylabel={},xtick=\empty,ytick=\empty]
			\addplot graphics [xmin=-4, xmax=30, ymin=-5, ymax=5] {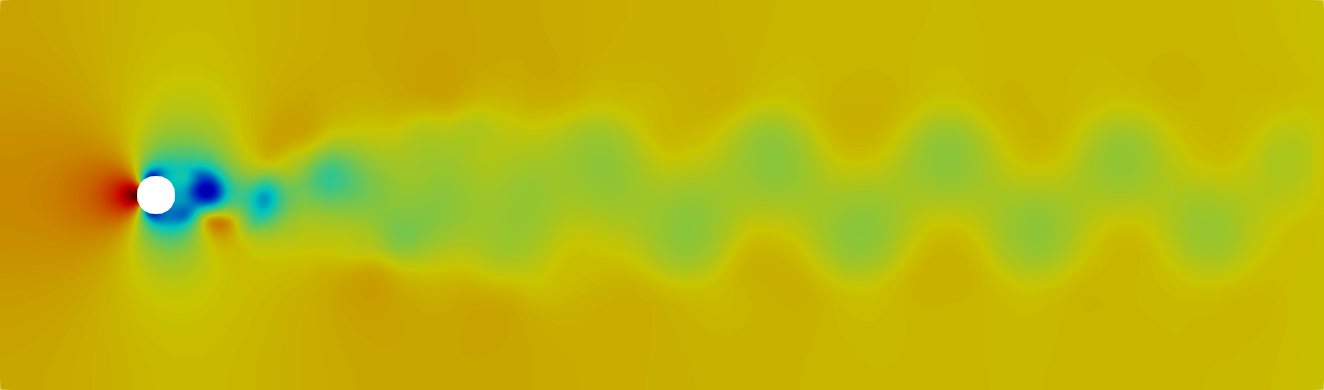};

            \nextgroupplot[title={Diffusion model at $Re=382$}]
			\addplot graphics [xmin=-4, xmax=30, ymin=-5, ymax=5] {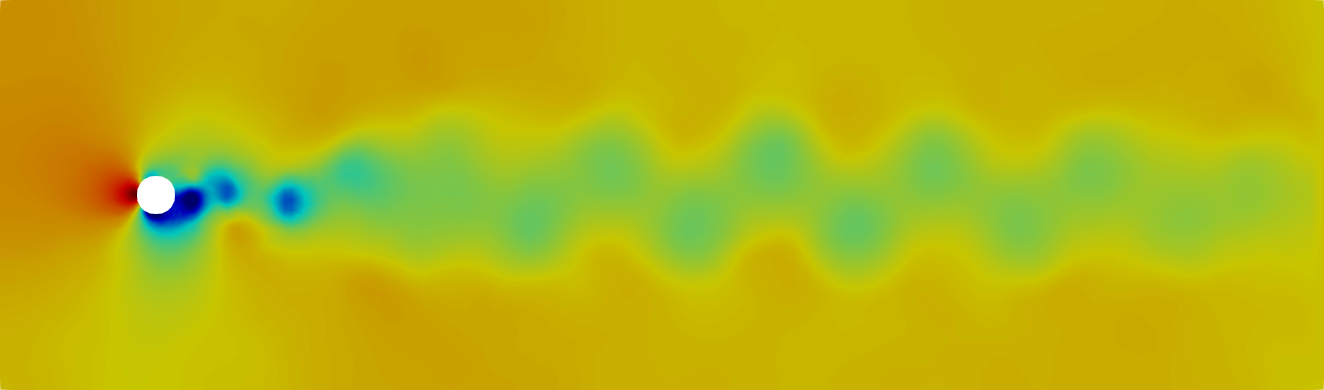};
                \nextgroupplot[title={DNS at $Re=382$},ylabel={},ytick=\empty]
			\addplot graphics [xmin=-4, xmax=30, ymin=-5, ymax=5] {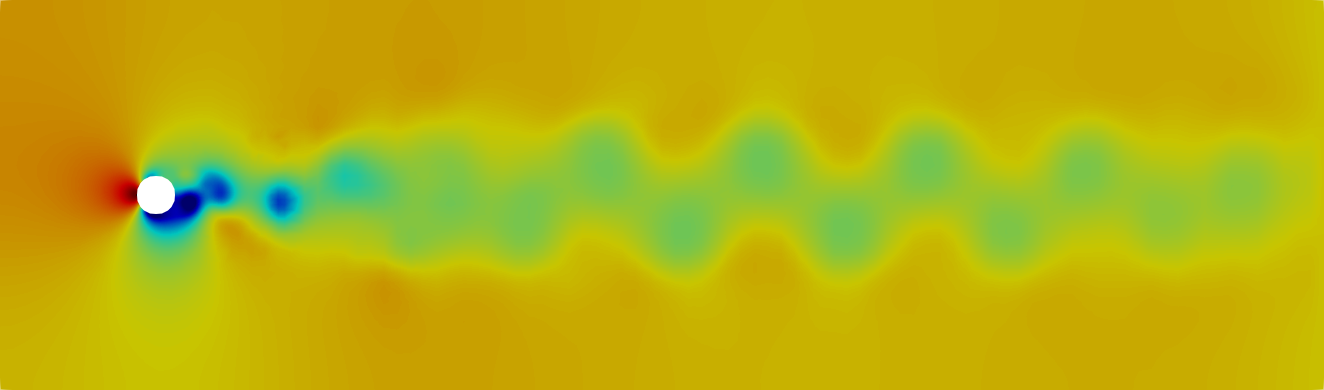};
			
		\end{groupplot}
	\end{tikzpicture}
\colorbarMatlabJet{-0.6}{-0.3}{0}{0.5}{1}
	\caption{Pressure of the viscous flow at three test Reynolds numbers, as generated by the diffusion model and compared with results from DNS.}
	\label{fig:ns_test_pressure}
\end{figure}

\begin{figure}[htp]
	\centering
	\begin{tikzpicture}
		\begin{groupplot}[
			group style={
				group size=2 by 3,
				horizontal sep=0.5cm
			},
			width=0.5\textwidth,
			axis equal image,
			xlabel={$x$},
			ylabel={$y$},
			xtick = {-4, 0, 30},
			ytick = {-5, 0.0, 5},
			xmin=-4, xmax=30,
			ymin=-5, ymax=5
			]
			\nextgroupplot[title={Guided initial state for dynamics at $Re=106$},xlabel={},xtick=\empty]
			\addplot graphics [xmin=-4, xmax=30, ymin=-5, ymax=5] {figures/p_diff_re_id_2.png};
                \nextgroupplot[title={Initial state for dynamics at $Re=106$},xlabel={},ylabel={},xtick=\empty,ytick=\empty]
			\addplot graphics [xmin=-4, xmax=30, ymin=-5, ymax=5] {figures/p_cfd_re_id_2.png};

                \nextgroupplot[title={Guided initial state for dynamics at $Re=238$},xlabel={},xtick=\empty]
			\addplot graphics [xmin=-4, xmax=30, ymin=-5, ymax=5] {figures/p_diff_re_id_24.png};
                \nextgroupplot[title={Initial state for dynamics at $Re=238$},xlabel={},ylabel={},xtick=\empty,ytick=\empty]
			\addplot graphics [xmin=-4, xmax=30, ymin=-5, ymax=5] {figures/p_cfd_re_id_24.png};

            \nextgroupplot[title={Guided initial state for dynamics at $Re=382$}]
			\addplot graphics [xmin=-4, xmax=30, ymin=-5, ymax=5] {figures/p_diff_re_id_48.png};
                \nextgroupplot[title={Initial state for dynamics at $Re=382$},ylabel={},ytick=\empty]
			\addplot graphics [xmin=-4, xmax=30, ymin=-5, ymax=5] {figures/p_cfd_re_id_48.png};
			
		\end{groupplot}
	\end{tikzpicture}
\colorbarMatlabJet{-0.6}{-0.3}{0}{0.5}{1}
	\caption{Initial pressure of the viscous flow at three test Reynolds numbers, as generated by the diffusion model and compared with results from DNS.}
	\label{fig:ns_ini_pressure}
\end{figure}

\begin{figure}
    \centering
    \input{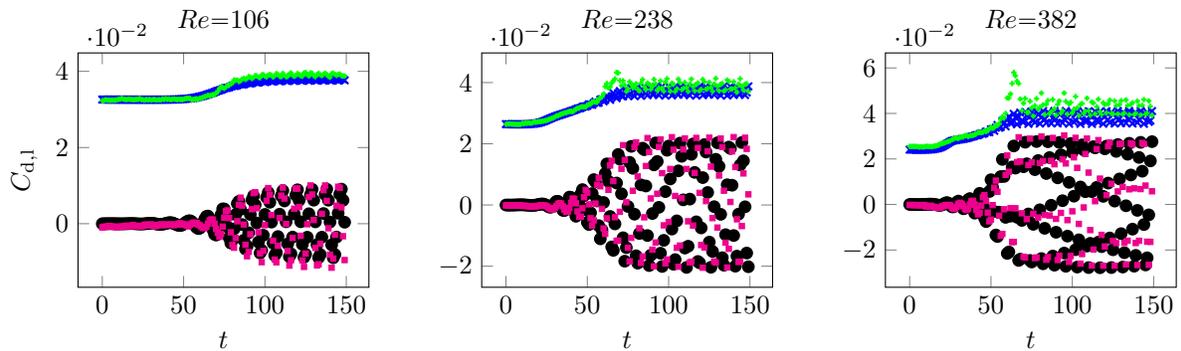}
    \caption{Drag coefficient from DNS (\ref{line:forecast_cd_cfd}), from the diffusion model (\ref{line:forecast_cd_diff}); 
    Lift coefficient from DNS (\ref{line:forecast_cl_cfd}), from the diffusion model (\ref{line:forecast_cl_diff})
    }
    \label{fig:cylinder_forca_clcd}
\end{figure}

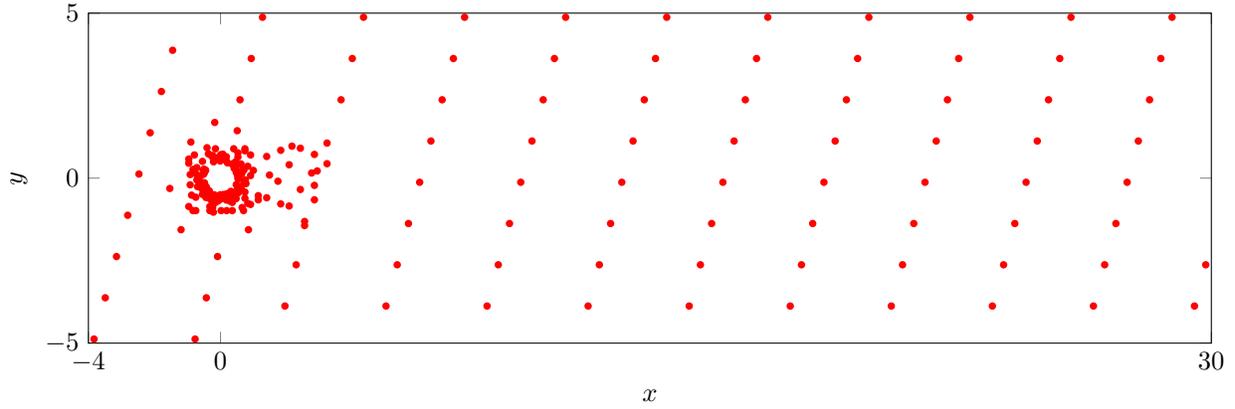
\begin{figure}[htp]
    \centering
    \input{figures/sparse_nodes.tikz}
    \caption{A sparse observation set consisting of only 256 points (fewer than those 1024 pivotal points shown in Figure~\ref{fig:pivotal_nodes} for the encoder) is utilized for the reconstruction of 11,644 cells depicted in Figure~\ref{fig:fpc_mesh}.}
    \label{fig:cylinder_sparse_obspoints}
\end{figure}

\begin{figure}[htp]
	\centering
	\begin{tikzpicture}
		\begin{groupplot}[
			group style={
				group size=2 by 3,
				horizontal sep=0.5cm
			},
			width=0.5\textwidth,
			axis equal image,
			xlabel={$x$},
			ylabel={$y$},
			xtick = {-4, 0, 30},
			ytick = {-5, 0.0, 5},
			xmin=-4, xmax=30,
			ymin=-5, ymax=5
			]
			\nextgroupplot[title={Reconstruction at $t=50$ for $Re=106$},xlabel={},xtick=\empty]
			\addplot graphics [xmin=-4, xmax=30, ymin=-5, ymax=5] {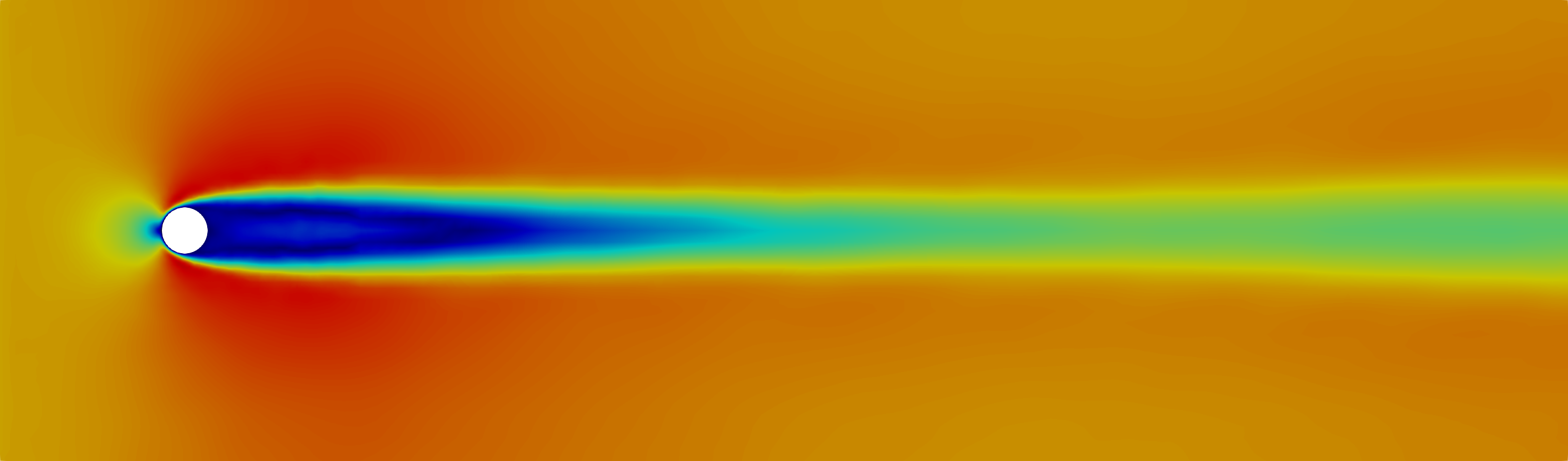};
                \nextgroupplot[title={Truth at $t=50$ for $Re=106$},xlabel={},ylabel={},xtick=\empty,ytick=\empty]
			\addplot graphics [xmin=-4, xmax=30, ymin=-5, ymax=5] {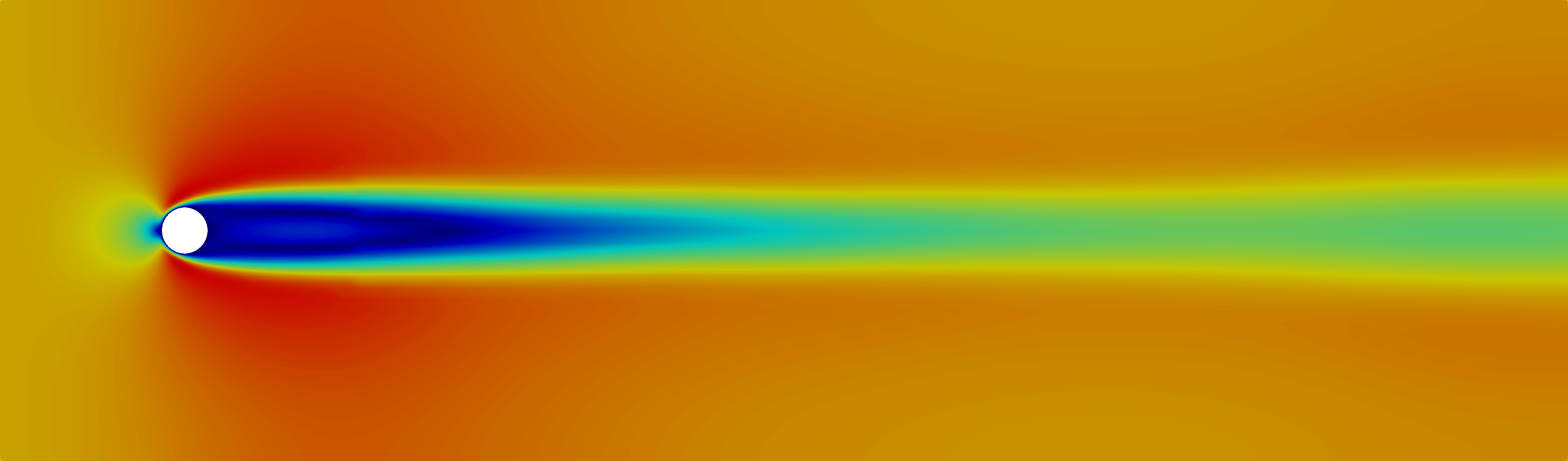};

                \nextgroupplot[title={Reconstruction at $t=50$ for $Re=238$},xlabel={},xtick=\empty]
			\addplot graphics [xmin=-4, xmax=30, ymin=-5, ymax=5] {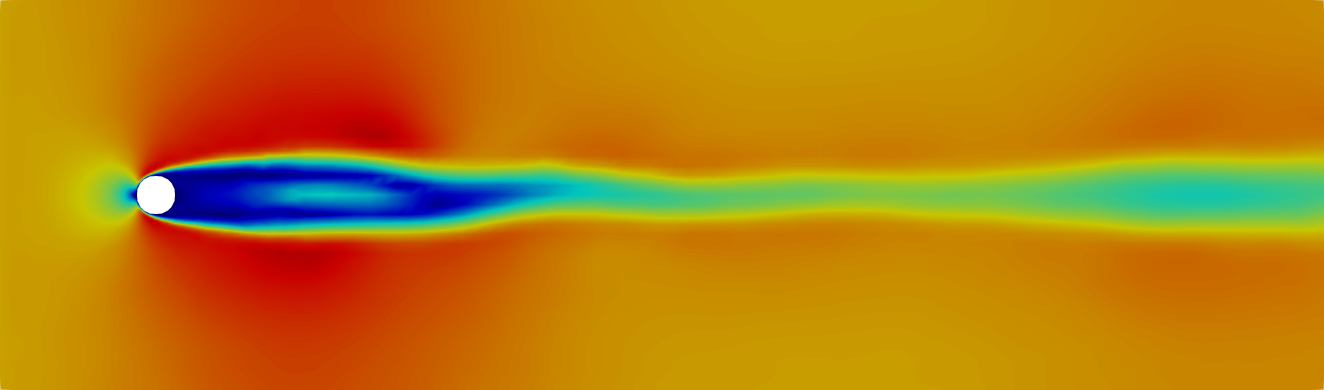};
                \nextgroupplot[title={Truth at $t=50$ for $Re=238$},xlabel={},ylabel={},xtick=\empty,ytick=\empty]
			\addplot graphics [xmin=-4, xmax=30, ymin=-5, ymax=5] {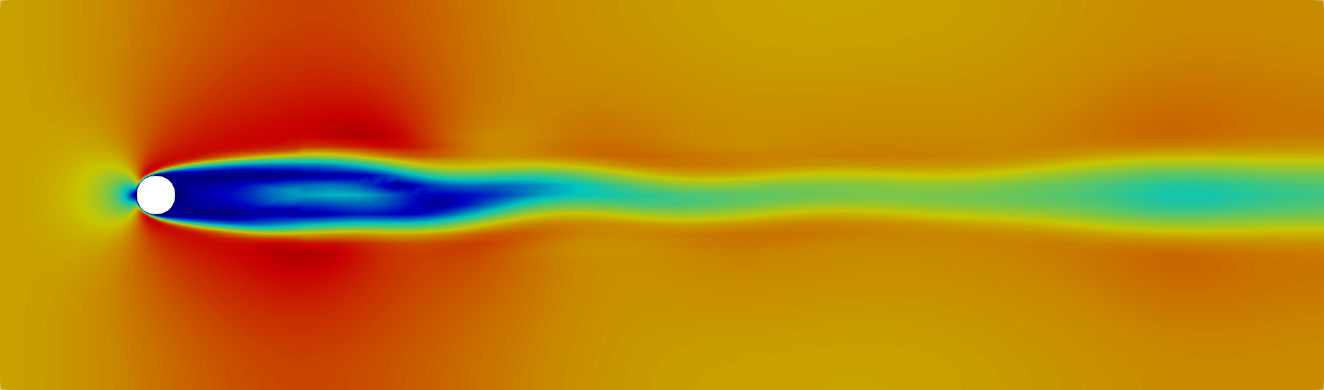};

            \nextgroupplot[title={Reconstruction at $t=50$ for $Re=382$}]
			\addplot graphics [xmin=-4, xmax=30, ymin=-5, ymax=5] {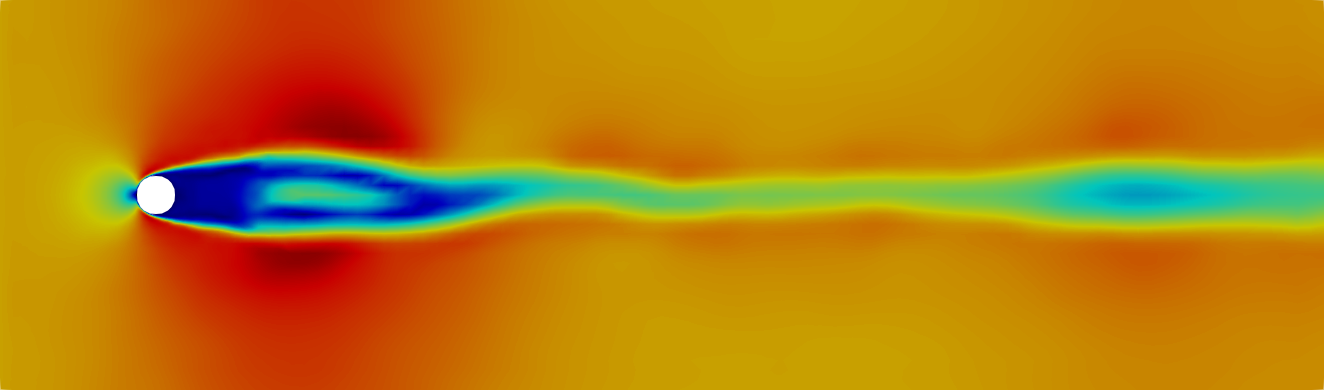};
                \nextgroupplot[title={Truth at $t=50$ for $Re=382$},ylabel={},ytick=\empty]
			\addplot graphics [xmin=-4, xmax=30, ymin=-5, ymax=5] {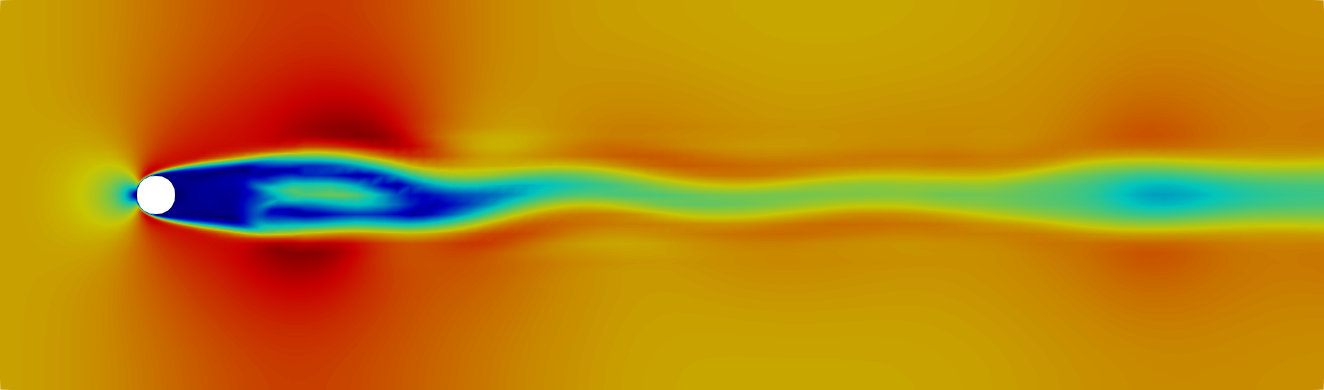};	
		\end{groupplot}
	\end{tikzpicture}
\colorbarMatlabJet{0}{0.4}{0.8}{1.1}{1.5}
	\caption{Velocity magnitude of the viscous flow at three test Reynolds numbers, as reconstructed by the diffusion model and compared with results from DNS.}
	\label{fig:ns_recon_vmag_t50}
\end{figure}

\begin{figure}[htp]
	\centering
	\begin{tikzpicture}
		\begin{groupplot}[
			group style={
				group size=2 by 3,
				horizontal sep=0.5cm
			},
			width=0.5\textwidth,
			axis equal image,
			xlabel={$x$},
			ylabel={$y$},
			xtick = {-4, 0, 30},
			ytick = {-5, 0.0, 5},
			xmin=-4, xmax=30,
			ymin=-5, ymax=5
			]
			\nextgroupplot[title={Reconstruction at $t=100$ for $Re=106$},xlabel={},xtick=\empty]
			\addplot graphics [xmin=-4, xmax=30, ymin=-5, ymax=5] {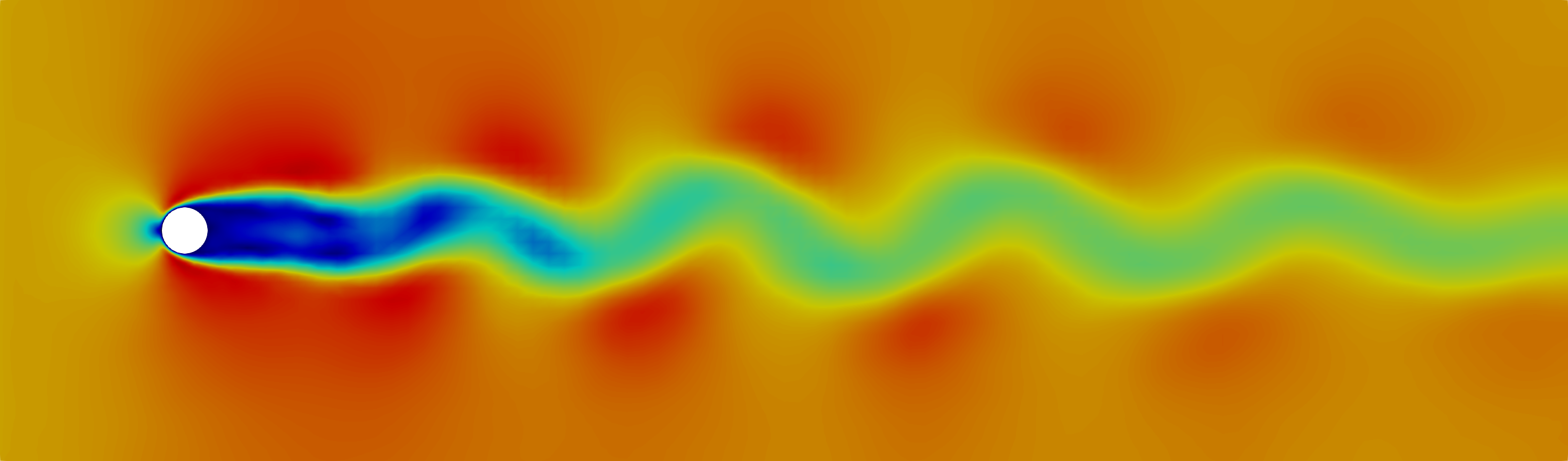};
                \nextgroupplot[title={Truth at $t=100$ for $Re=106$},xlabel={},ylabel={},xtick=\empty,ytick=\empty]
			\addplot graphics [xmin=-4, xmax=30, ymin=-5, ymax=5] {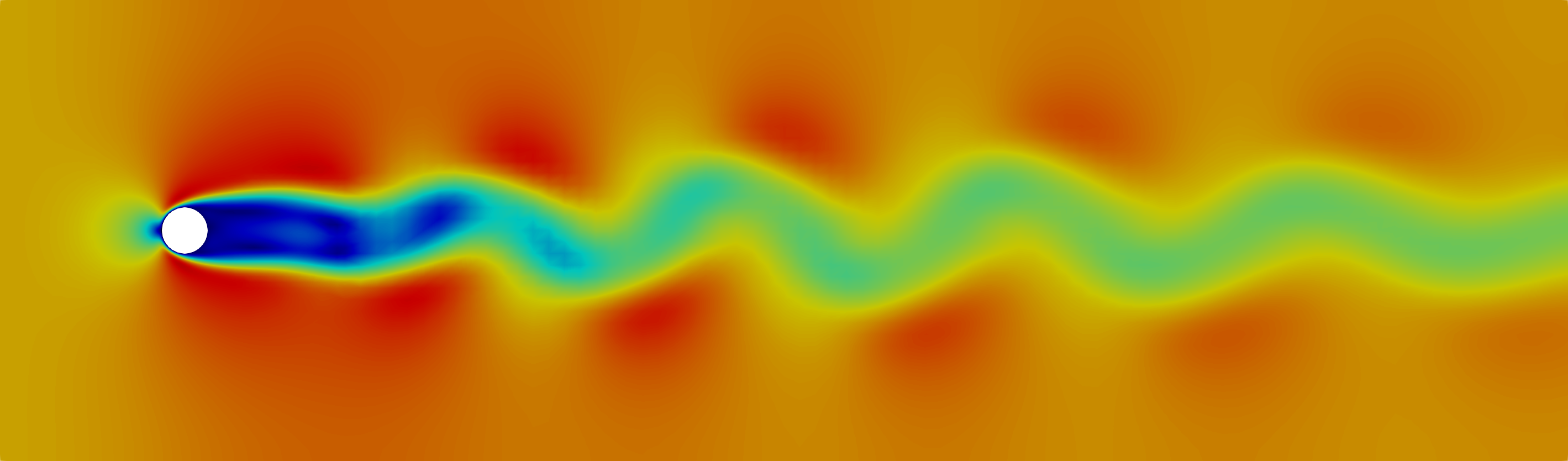};

                \nextgroupplot[title={Reconstruction at $t=100$ for $Re=238$},xlabel={},xtick=\empty]
			\addplot graphics [xmin=-4, xmax=30, ymin=-5, ymax=5] {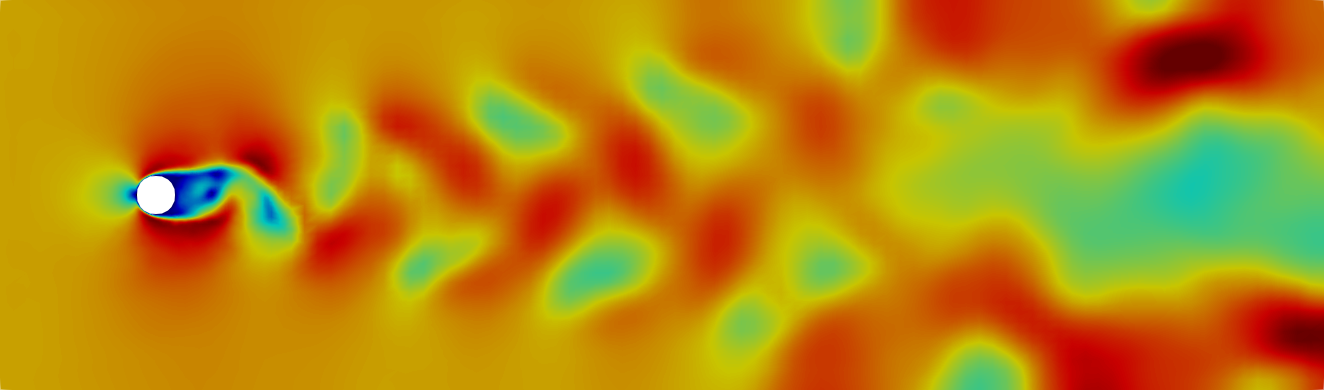};
                \nextgroupplot[title={Truth at $t=100$ for $Re=238$},xlabel={},ylabel={},xtick=\empty,ytick=\empty]
			\addplot graphics [xmin=-4, xmax=30, ymin=-5, ymax=5] {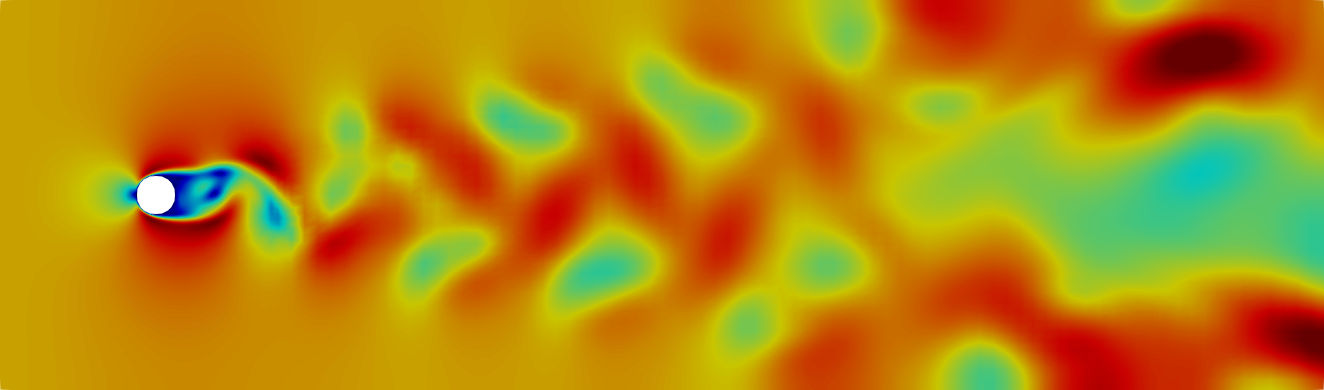};

            \nextgroupplot[title={Reconstruction at $t=100$ for $Re=382$}]
			\addplot graphics [xmin=-4, xmax=30, ymin=-5, ymax=5] {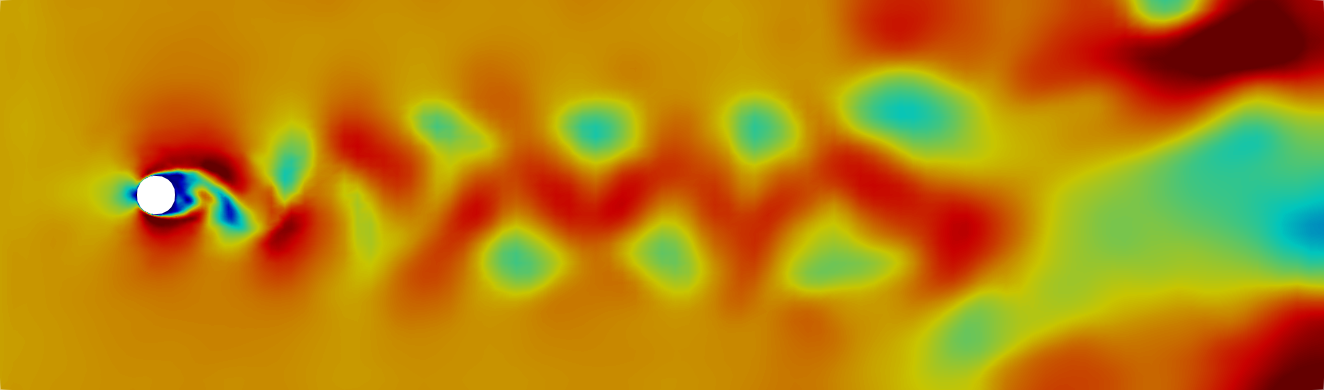};
                \nextgroupplot[title={Truth at $t=100$ for $Re=382$},ylabel={},ytick=\empty]
			\addplot graphics [xmin=-4, xmax=30, ymin=-5, ymax=5] {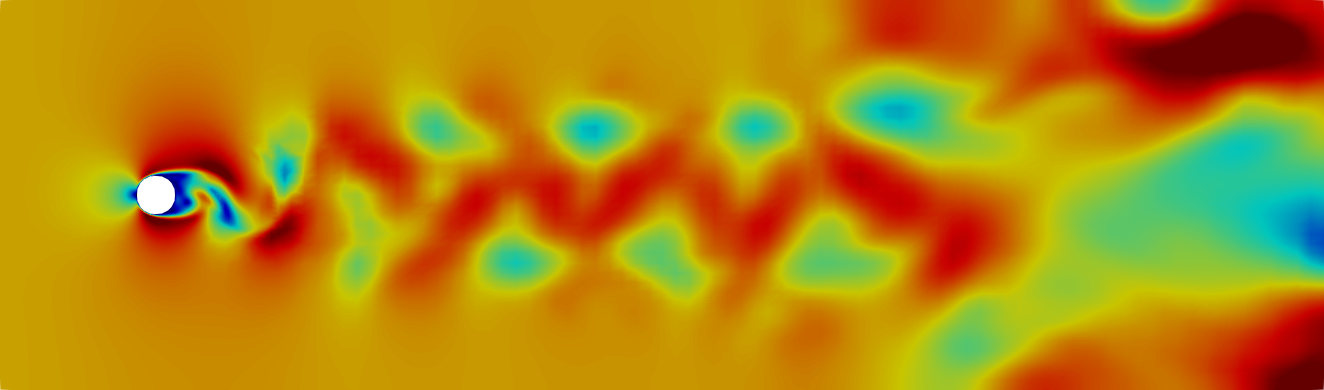};	
		\end{groupplot}
	\end{tikzpicture}
\colorbarMatlabJet{0}{0.4}{0.8}{1.1}{1.5}
	\caption{Velocity magnitude of the viscous flow at three test Reynolds numbers, as reconstructed by the diffusion model and compared with results from DNS.}
	\label{fig:ns_recon_vmag_t100}
\end{figure}

\begin{figure}[htp]
	\centering
	\begin{tikzpicture}
		\begin{groupplot}[
			group style={
				group size=2 by 3,
				horizontal sep=0.5cm
			},
			width=0.5\textwidth,
			axis equal image,
			xlabel={$x$},
			ylabel={$y$},
			xtick = {-4, 0, 30},
			ytick = {-5, 0.0, 5},
			xmin=-4, xmax=30,
			ymin=-5, ymax=5
			]
			\nextgroupplot[title={Reconstruction at $t=150$ for $Re=106$},xlabel={},xtick=\empty]
			\addplot graphics [xmin=-4, xmax=30, ymin=-5, ymax=5] {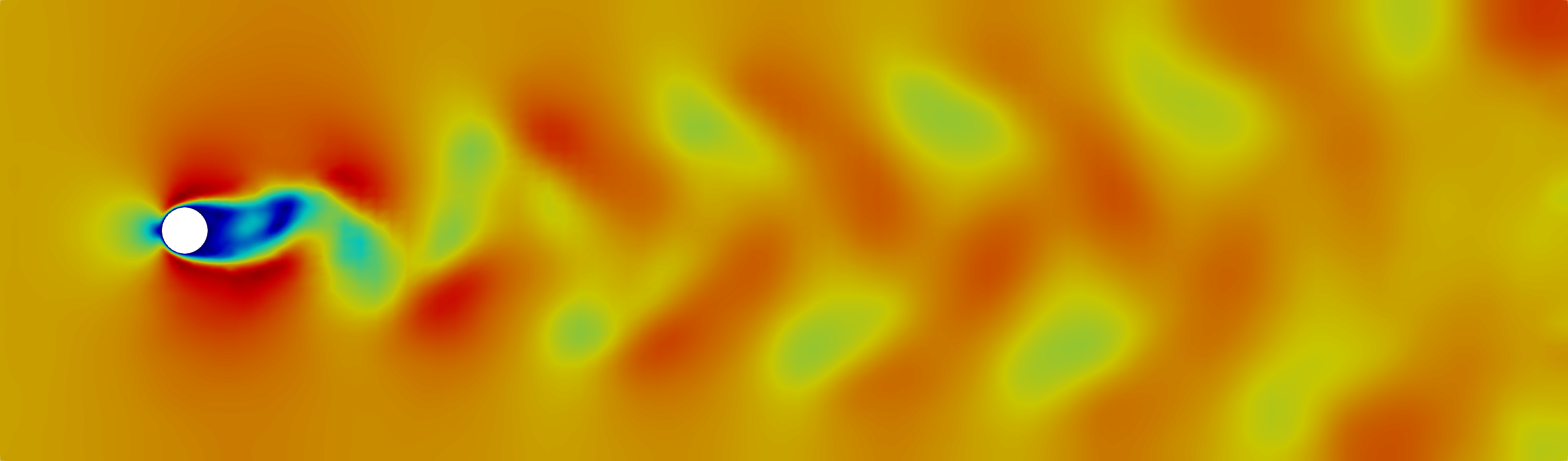};
                \nextgroupplot[title={Truth at $t=150$ for $Re=106$},xlabel={},ylabel={},xtick=\empty,ytick=\empty]
			\addplot graphics [xmin=-4, xmax=30, ymin=-5, ymax=5] {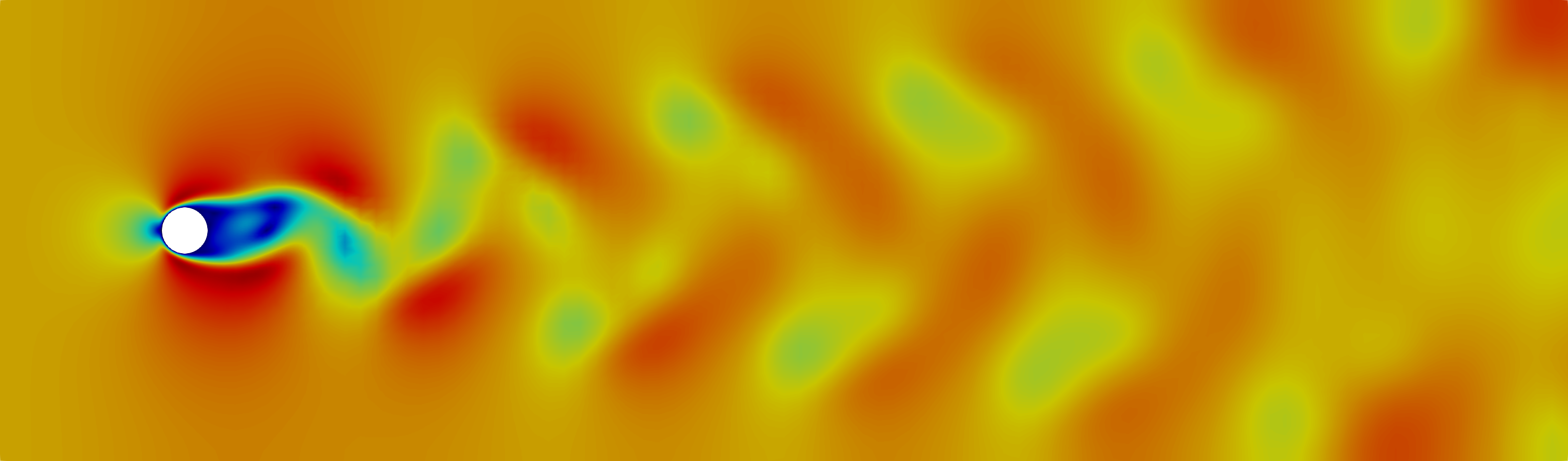};

                \nextgroupplot[title={Reconstruction at $t=150$ for $Re=238$},xlabel={},xtick=\empty]
			\addplot graphics [xmin=-4, xmax=30, ymin=-5, ymax=5] {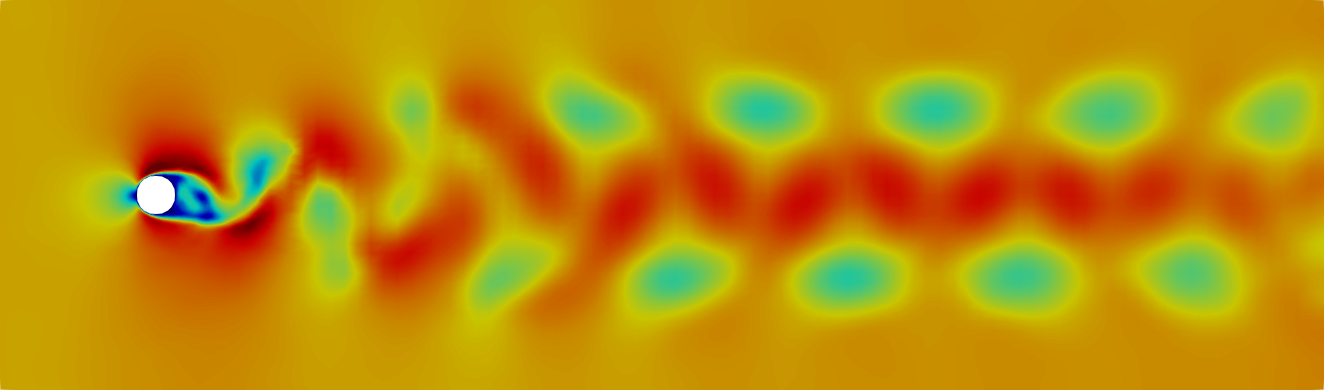};
                \nextgroupplot[title={Truth at $t=150$ for $Re=238$},xlabel={},ylabel={},xtick=\empty,ytick=\empty]
			\addplot graphics [xmin=-4, xmax=30, ymin=-5, ymax=5] {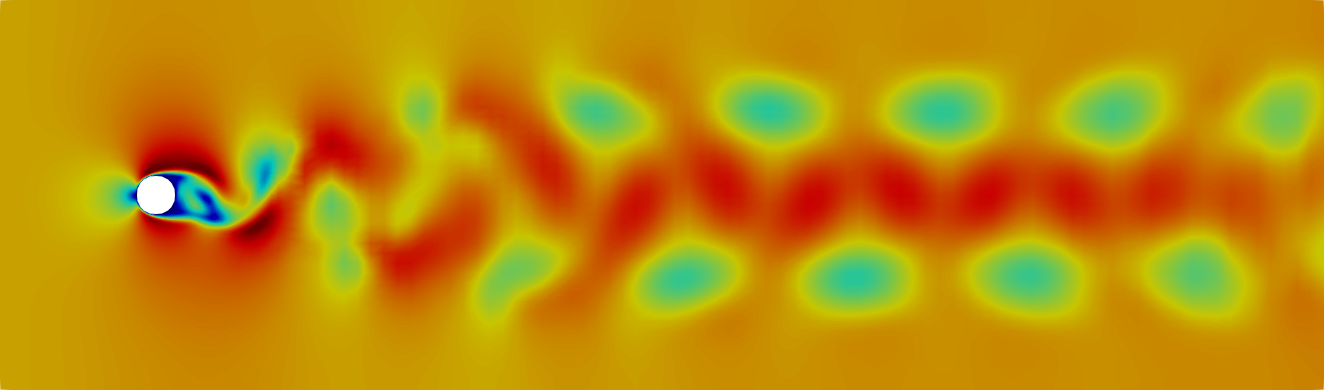};

            \nextgroupplot[title={Reconstruction at $t=150$ for $Re=382$}]
			\addplot graphics [xmin=-4, xmax=30, ymin=-5, ymax=5] {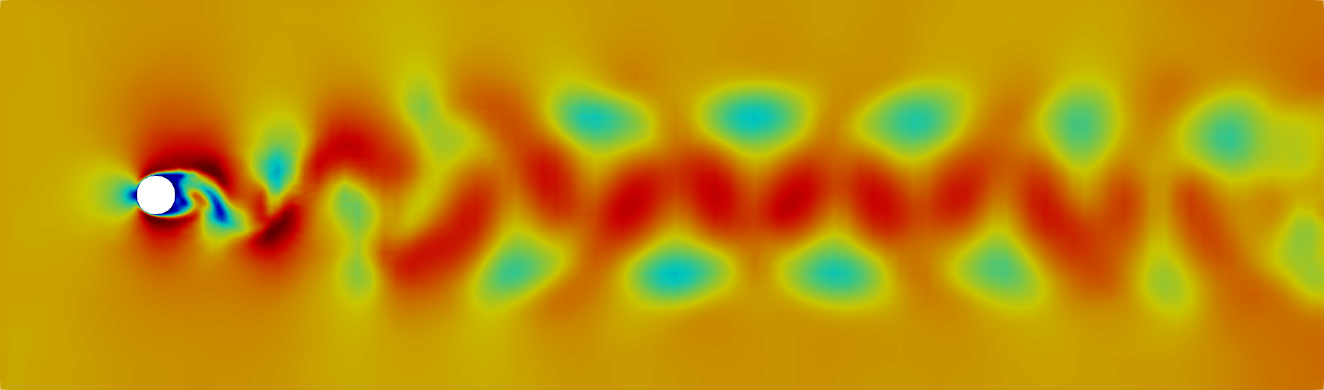};
                \nextgroupplot[title={Truth at $t=150$ for $Re=382$},ylabel={},ytick=\empty]
			\addplot graphics [xmin=-4, xmax=30, ymin=-5, ymax=5] {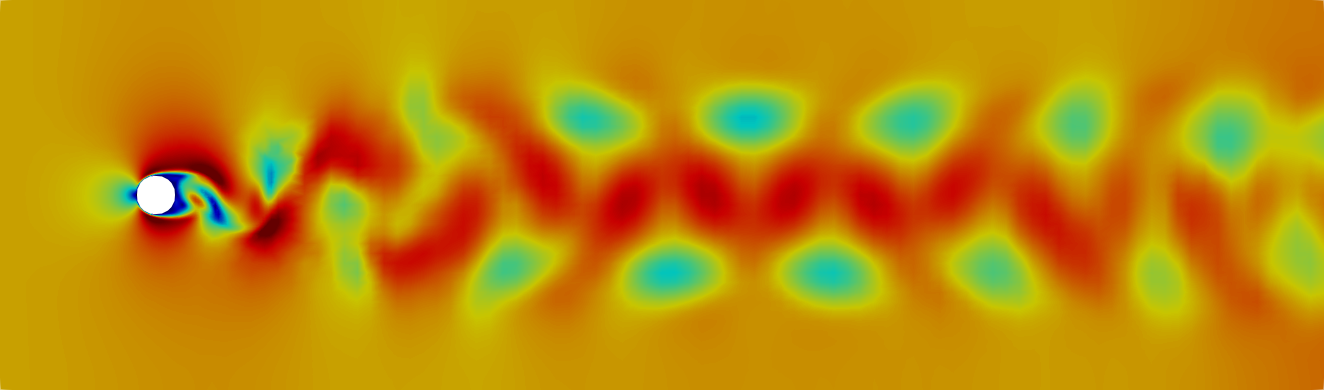};	
		\end{groupplot}
	\end{tikzpicture}
\colorbarMatlabJet{0}{0.4}{0.8}{1.1}{1.5}
	\caption{Velocity magnitude of the viscous flow at three test Reynolds numbers, as reconstructed by the diffusion model and compared with results from DNS.}
	\label{fig:ns_recon_vmag_t150}
\end{figure}

\newpage
\section{Additional results for channel flow}
\begin{figure}[htp]
	\centering
	\begin{tikzpicture}
		\begin{groupplot}[
			group style={
				group size=2 by 3,
				horizontal sep=0.5cm
			},
			width=0.5\textwidth,
			axis equal image,
			xlabel={$x$},
			ylabel={$y$},
			xtick = {0, 3.14, 6.28},
			ytick = {0, 1, 2},
			xmin=0, xmax=6.28,
			ymin=0, ymax=2
			]
			\nextgroupplot[title={Diffusion model at $Re_\tau=226$},xlabel={},xtick=\empty]
			\addplot graphics [xmin=0, xmax=6.28, ymin=0, ymax=2] {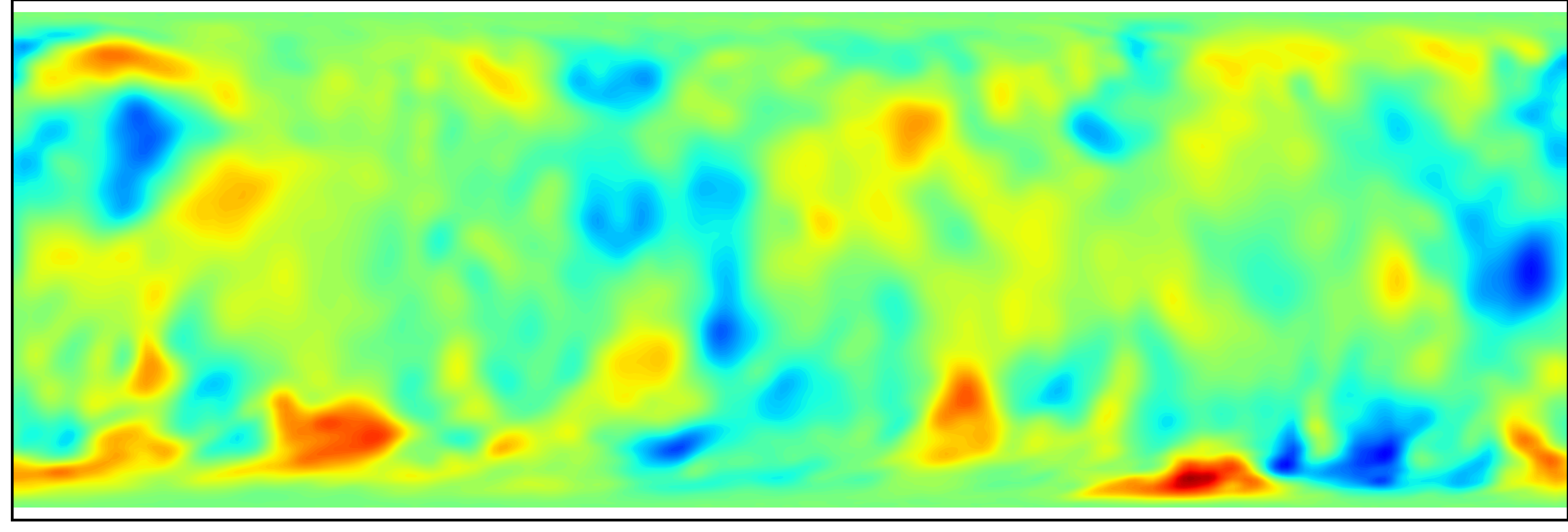};
                \nextgroupplot[title={DNS at $Re_\tau=226$},xlabel={},ylabel={},xtick=\empty,ytick=\empty]
			\addplot graphics [xmin=0, xmax=6.28, ymin=0, ymax=2] {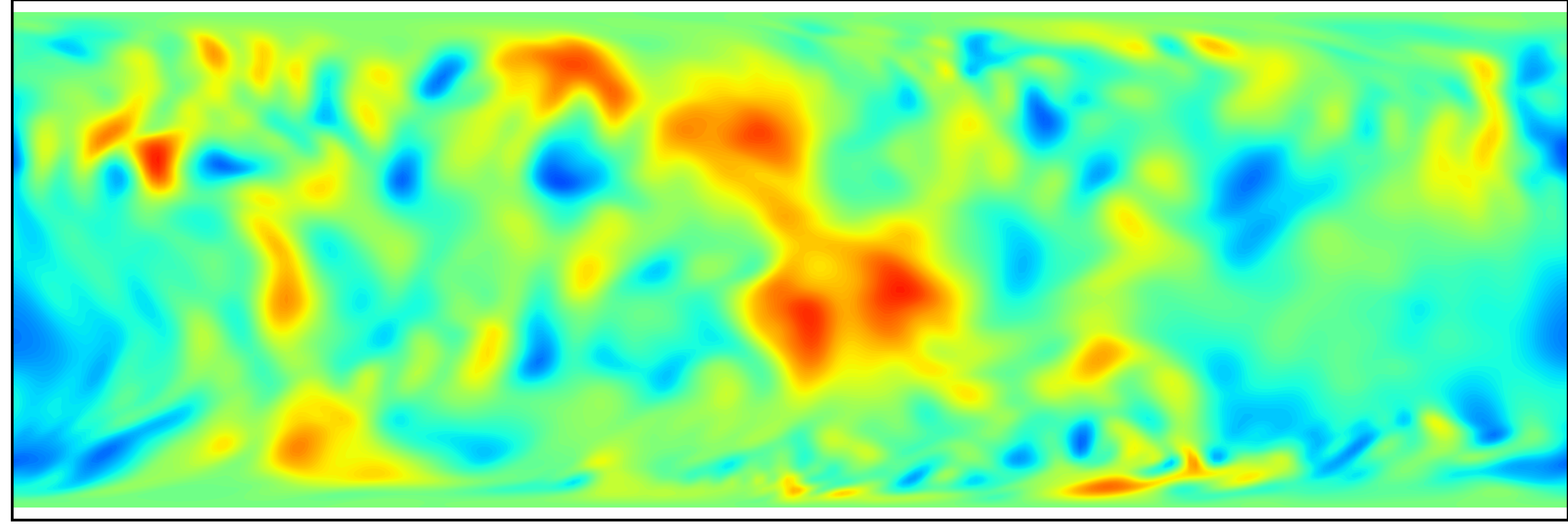};

                \nextgroupplot[title={Diffusion model at $Re_\tau=339$},xlabel={},xtick=\empty]
			\addplot graphics [xmin=0, xmax=6.28, ymin=0, ymax=2] {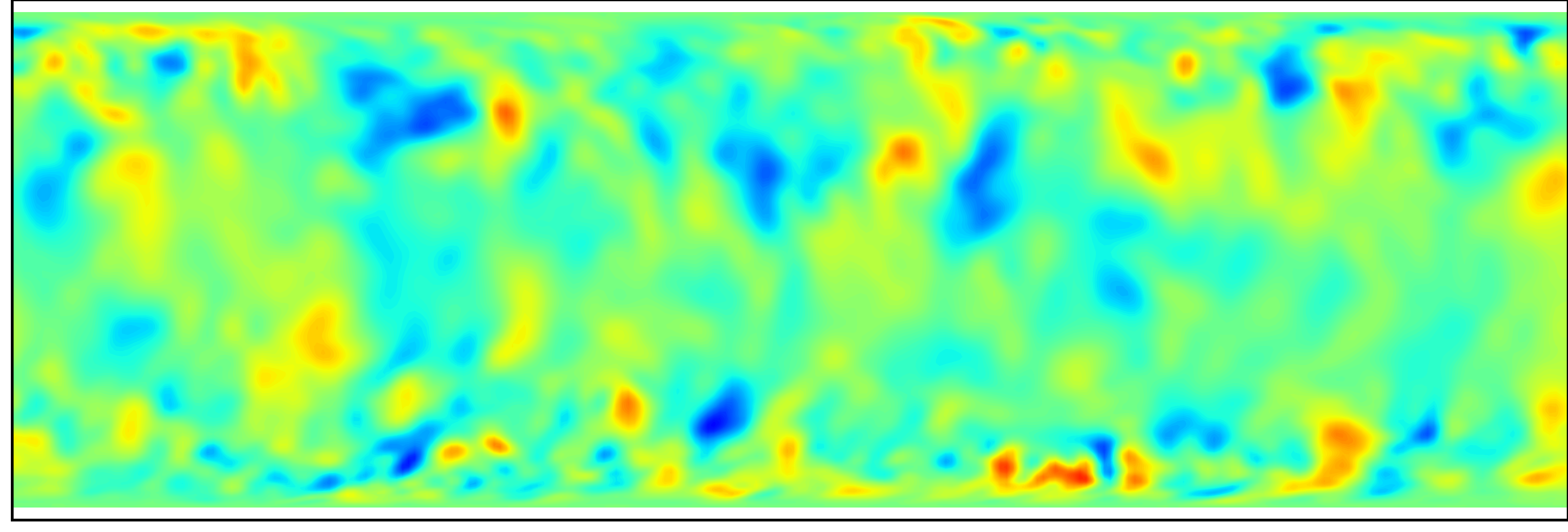};
                \nextgroupplot[title={DNS at $Re_\tau=339$},xlabel={},ylabel={},xtick=\empty,ytick=\empty]
			\addplot graphics [xmin=0, xmax=6.28, ymin=0, ymax=2] {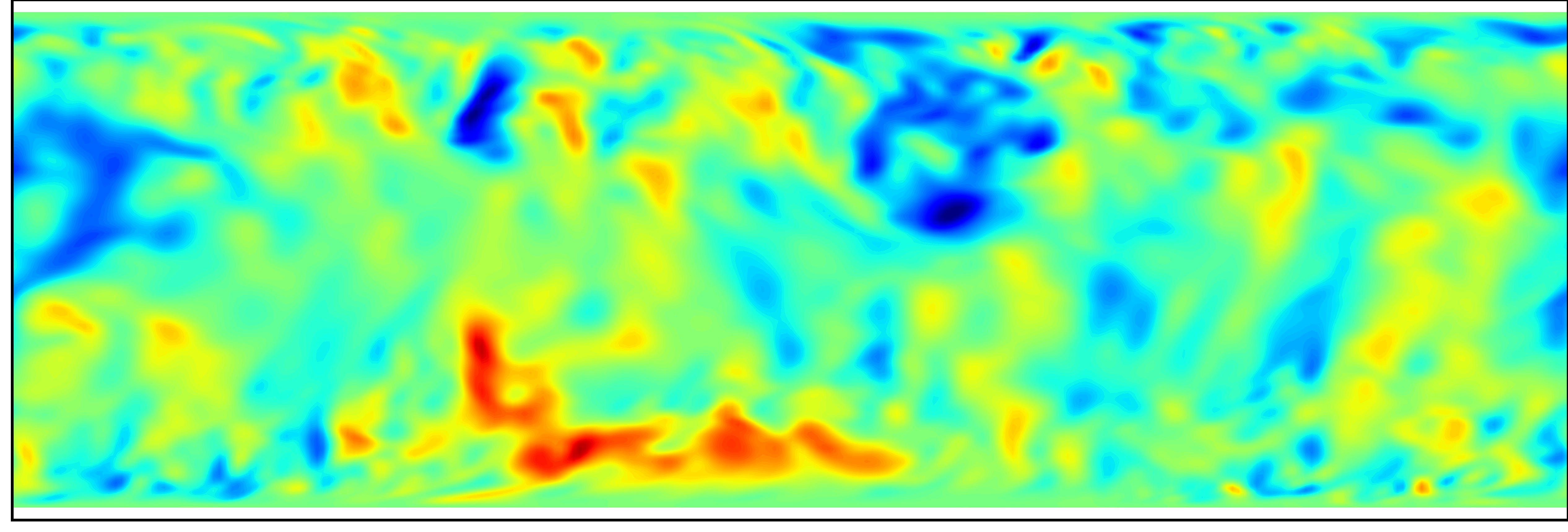};

            \nextgroupplot[title={Diffusion model at $Re_\tau=546$}]
			\addplot graphics [xmin=0, xmax=6.28, ymin=0, ymax=2] {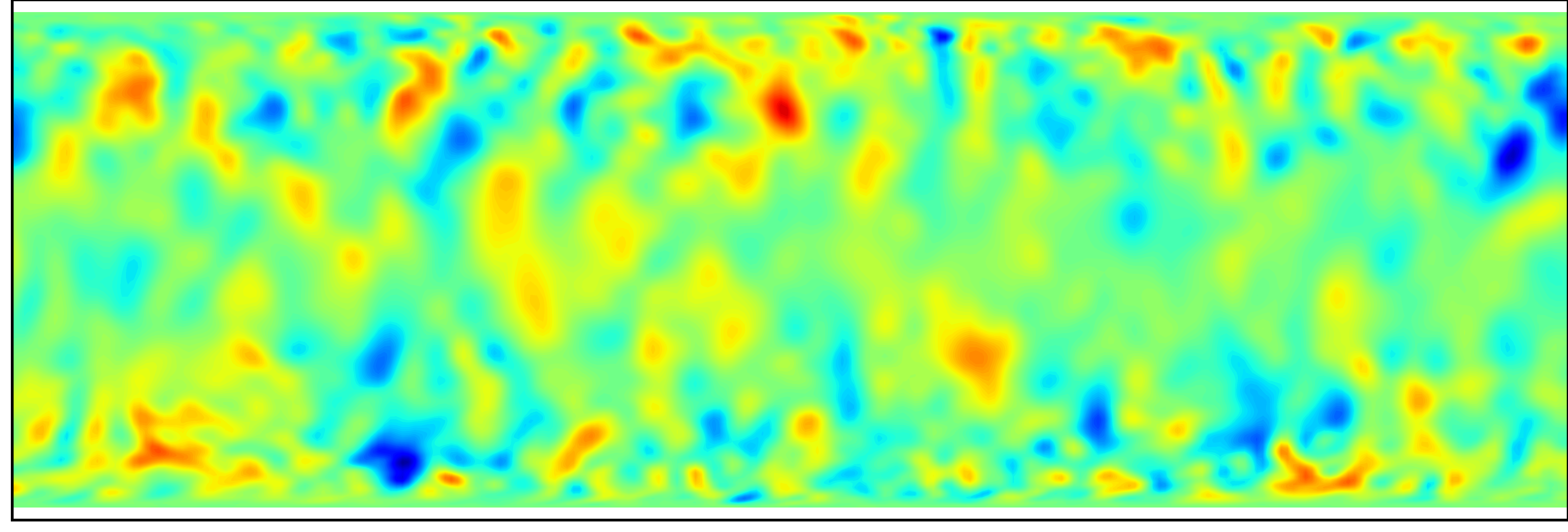};
                \nextgroupplot[title={DNS at $Re_\tau=546$},ylabel={},ytick=\empty]
			\addplot graphics [xmin=0, xmax=6.28, ymin=0, ymax=2] {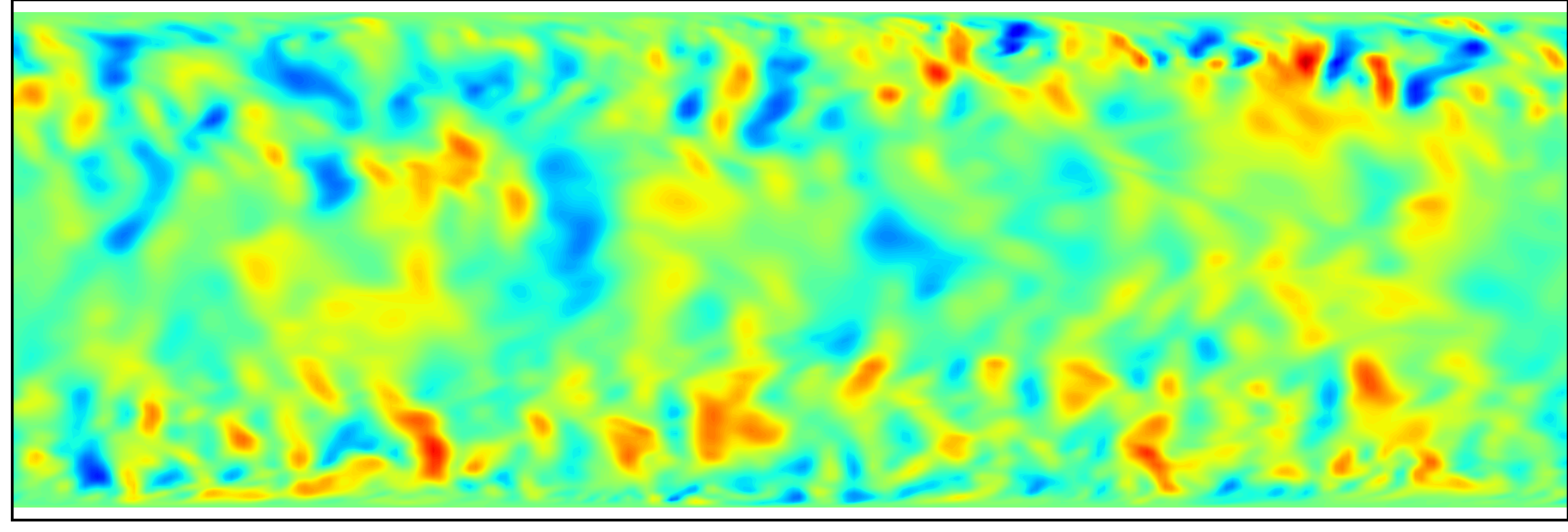};
			
		\end{groupplot}
	\end{tikzpicture}
\colorbarMatlabJet{-0.03}{-0.015}{0}{0.015}{0.03}
	\caption{Wallnormal velocity of the viscous flow at three test Reynolds numbers, as generated by the diffusion model and compared with results from DNS.}
	\label{fig:ns_channel_v_contour}
\end{figure}

\begin{figure}[htp]
	\centering
	\begin{tikzpicture}
		\begin{groupplot}[
			group style={
				group size=2 by 3,
				horizontal sep=0.5cm
			},
			width=0.5\textwidth,
			axis equal image,
			xlabel={$x$},
			ylabel={$y$},
			xtick = {0, 3.14, 6.28},
			ytick = {0, 1, 2},
			xmin=0, xmax=6.28,
			ymin=0, ymax=2
			]
			\nextgroupplot[title={Diffusion model at $Re_\tau=226$},xlabel={},xtick=\empty]
			\addplot graphics [xmin=0, xmax=6.28, ymin=0, ymax=2] {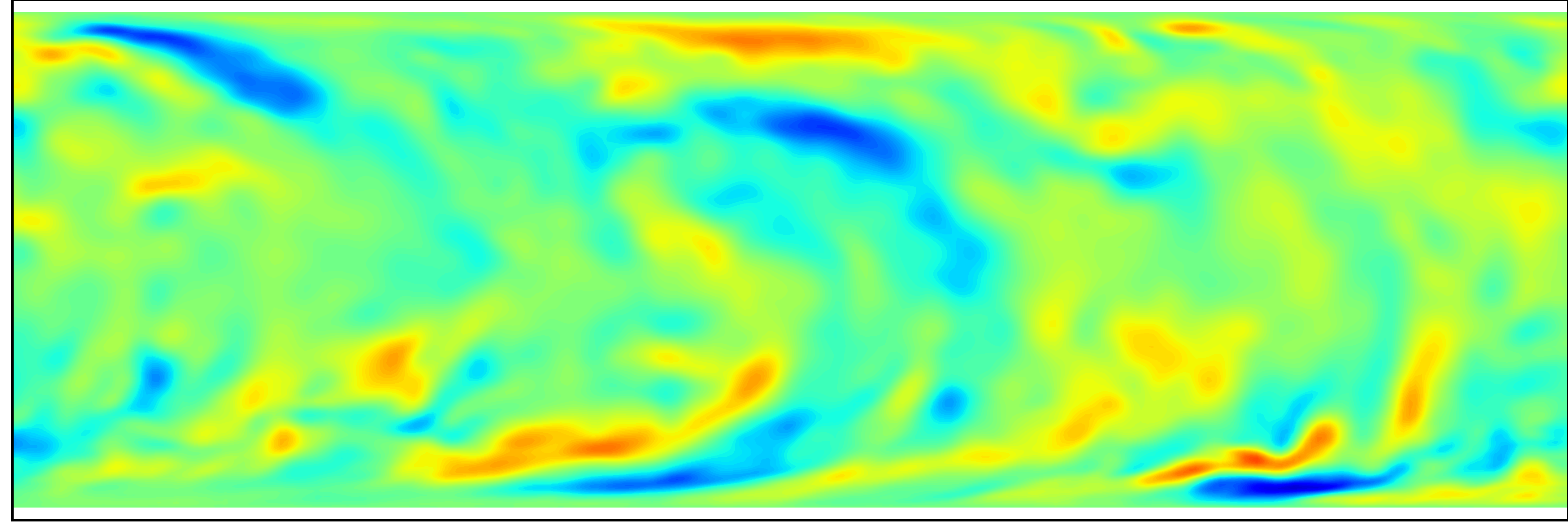};
                \nextgroupplot[title={DNS at $Re_\tau=226$},xlabel={},ylabel={},xtick=\empty,ytick=\empty]
			\addplot graphics [xmin=0, xmax=6.28, ymin=0, ymax=2] {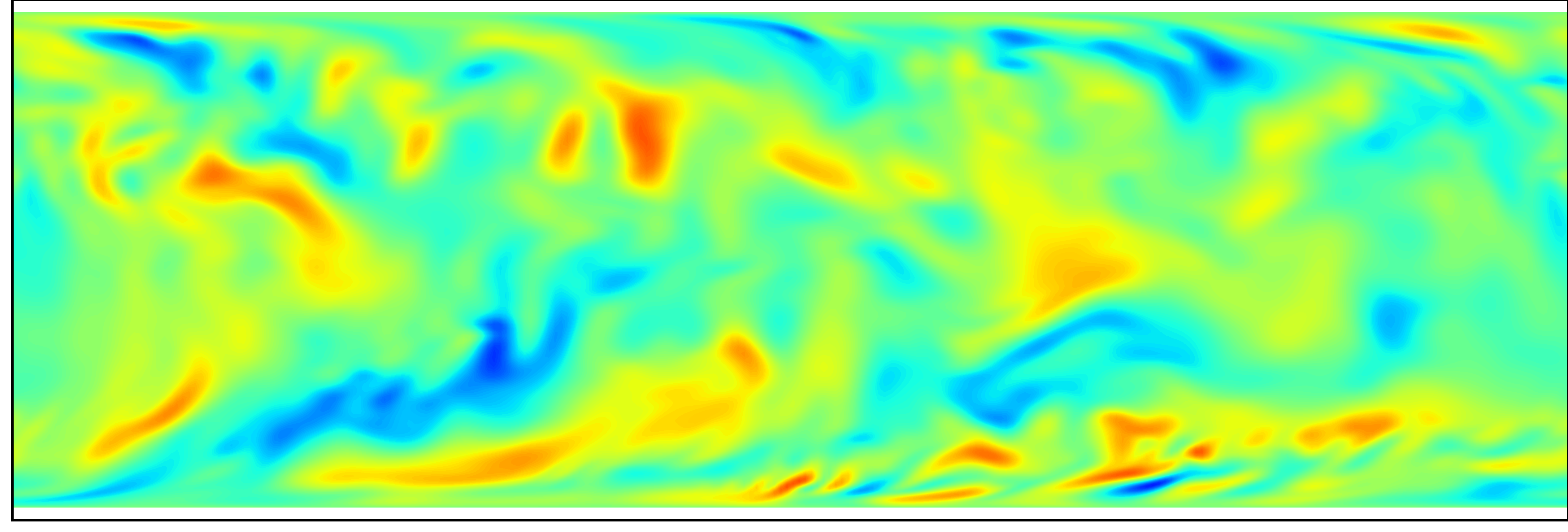};

                \nextgroupplot[title={Diffusion model at $Re_\tau=339$},xlabel={},xtick=\empty]
			\addplot graphics [xmin=0, xmax=6.28, ymin=0, ymax=2] {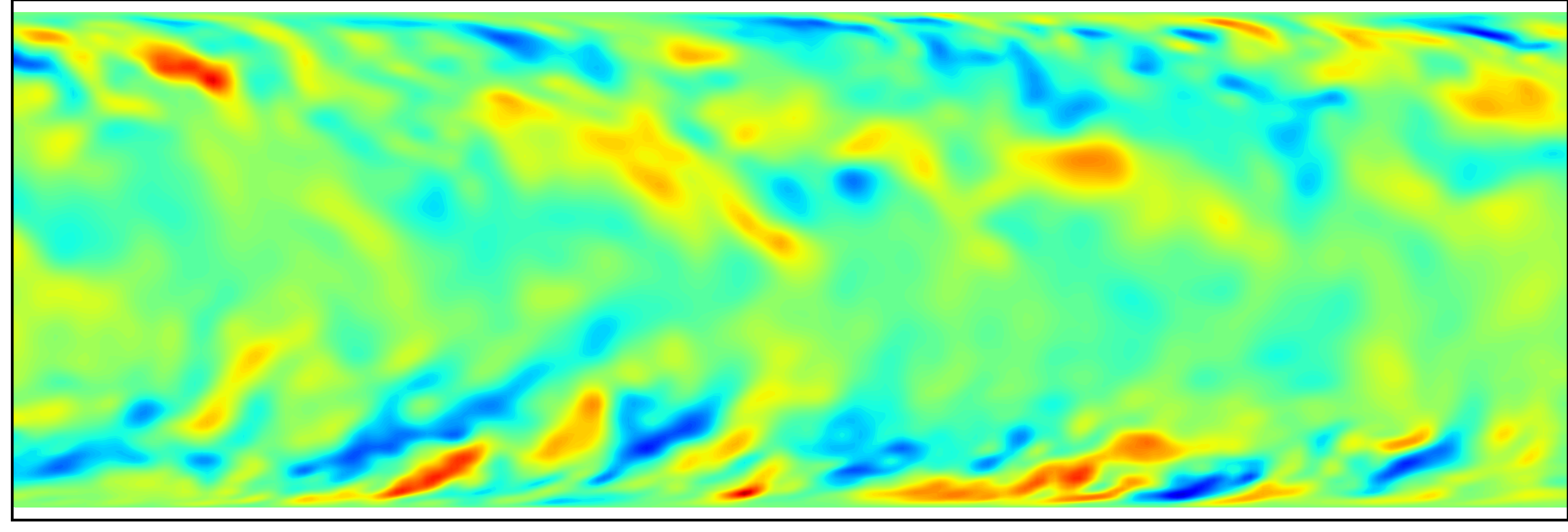};
                \nextgroupplot[title={DNS at $Re_\tau=339$},xlabel={},ylabel={},xtick=\empty,ytick=\empty]
			\addplot graphics [xmin=0, xmax=6.28, ymin=0, ymax=2] {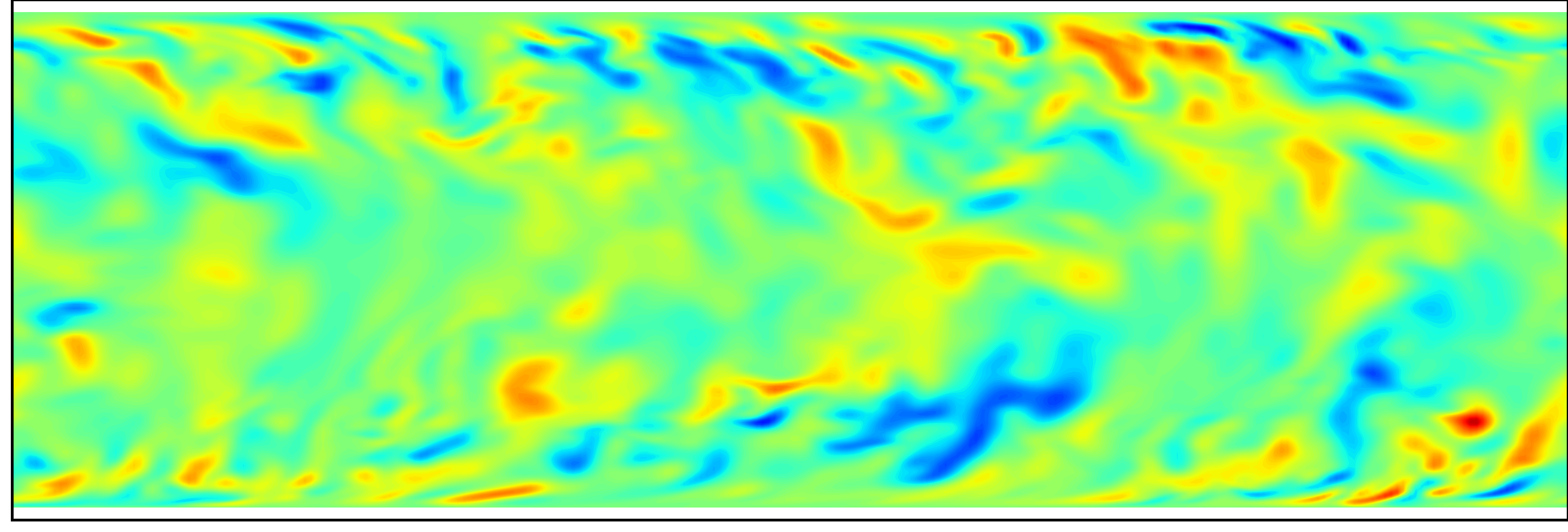};

            \nextgroupplot[title={Diffusion model at $Re_\tau=546$}]
			\addplot graphics [xmin=0, xmax=6.28, ymin=0, ymax=2] {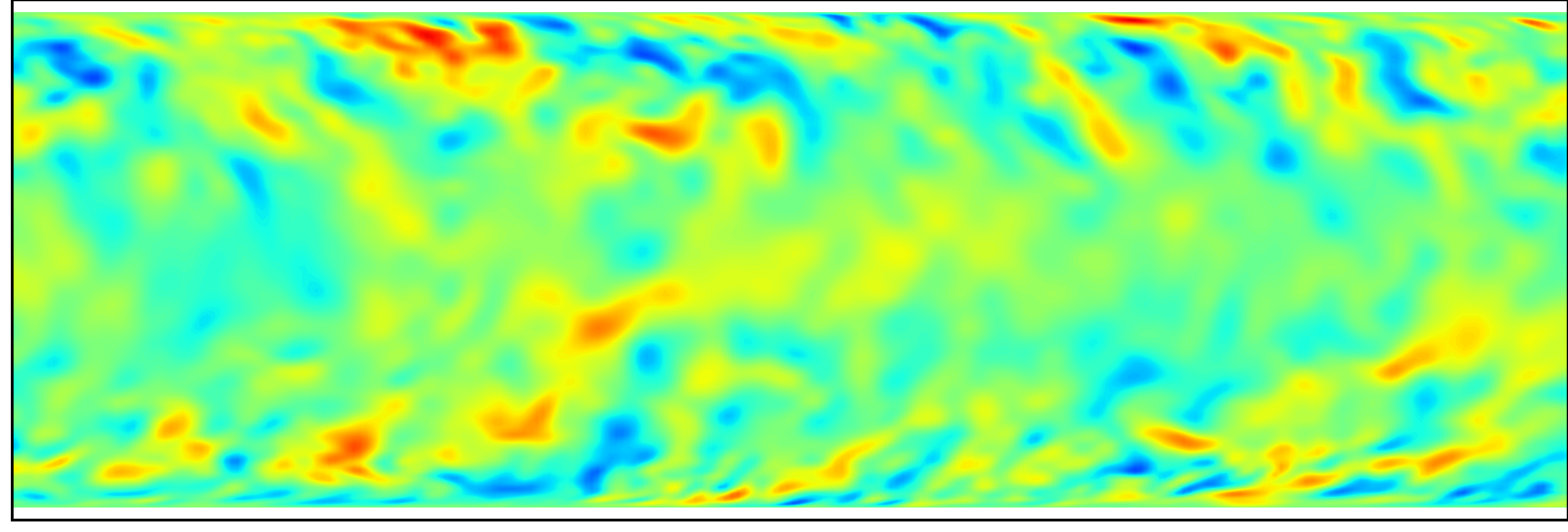};
                \nextgroupplot[title={DNS at $Re_\tau=546$},ylabel={},ytick=\empty]
			\addplot graphics [xmin=0, xmax=6.28, ymin=0, ymax=2] {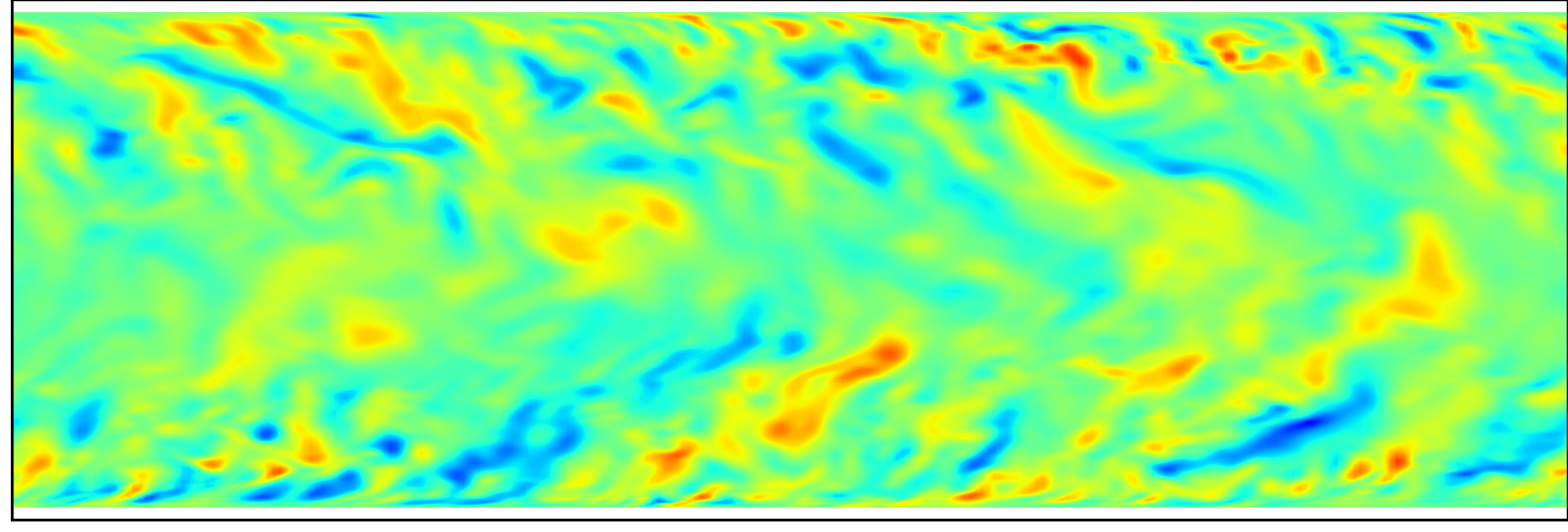};
			
		\end{groupplot}
	\end{tikzpicture}
\colorbarMatlabJet{-0.04}{-0.02}{0}{0.02}{0.04}
	\caption{Spanwise velocity of the viscous flow at three test Reynolds numbers, as generated by the diffusion model and compared with results from DNS.}
	\label{fig:ns_channel_w_contour}
\end{figure}

\begin{figure}[htp]
    \centering
    \input{figures/SpatiaTemporalCorrelation1.tikz}
    \caption{Wallnormal velocity spatial and temporal correlation at $y^+ = 20$ from DNS (\ref{line:idns:dns_y20}) and diffusion model (\ref{line:idns:diff_y20}); at $y^+ = 50$ from DNS (\ref{line:idns:dns_y50}) and diffusion model (\ref{line:idns:diff_y50}); at $y^+ = 100$ from DNS (\ref{line:idns:dns_y100}) and diffusion model (\ref{line:idns:diff_y100}).
    }
   \label{fig:correlations1}
\end{figure}

\begin{figure}[htp]
    \centering
    \input{figures/SpatiaTemporalCorrelation2.tikz}
    \caption{Spanwise velocity spatial and temporal correlation at $y^+ = 20$ from DNS (\ref{line:idns:dns_y20}) and diffusion model (\ref{line:idns:diff_y20}); at $y^+ = 50$ from DNS (\ref{line:idns:dns_y50}) and diffusion model (\ref{line:idns:diff_y50}); at $y^+ = 100$ from DNS (\ref{line:idns:dns_y100}) and diffusion model (\ref{line:idns:diff_y100}).
    }
    \label{fig:correlations2}
\end{figure}

\section{Model settings}
\begin{table}[htp]
\centering
\begin{tabular}{||c | c | c  ||} 	
 \hline
hyperparameters        & Cylinder flow & Channel flow  \\ 
	       \hline 
number of channels in convolution layers & (32,64,128,256) & (32,64,128,256) \\ 
\hline
number of noise step ($N_\mathrm{noise}$) & 20 & 20 \\ 
\hline
noise range in diffusion model  & (0.002,80) & (0.002,80) \\ 
\hline
\end{tabular}
\caption{Important hyperparameters of for the diffusion model.}
\label{tab:hyper-parameters-diff}
\end{table}

\begin{table}[htp]
\centering
\begin{tabular}{||c | c | c  ||} 	
 \hline
hyperparameters        & Cylinder flow & Channel flow  \\ 
	       \hline 
number of channels in encoder layers & (128,128,128) & None \\ 
\hline
number of channels in decoder layers & (128,128,128) & (64,64) \\ 
\hline
virtual observation model  & None &  MLP (1,64,64,64,128) \\ 
\hline
\end{tabular}
\caption{Important hyperparameters of for the encoders and decoders.}
\label{tab:hyper-parameters-ed}
\end{table}

\section*{Acknowledgments}
S.K. and P.K. acknowledge support by The European High Performance Computing Joint Undertaking (EuroHPC) Grant DCoMEX (956201-H2020-JTI-EuroHPC-2019-1).

\bibliographystyle{plain}
\bibliography{biblio}

\end{document}

%% file: preamble.pap.tex
\usepackage[margin=1in]{geometry}

\usepackage{bbm}
\usepackage{graphicx}
\usepackage{pstool}
\usepackage{wrapfig}
\usepackage{caption}
\usepackage{subcaption}
\usepackage[section]{placeins} 

\usepackage{fancyhdr} 
\usepackage{lastpage} 
\usepackage{extramarks} 

\usepackage{xcolor} 
\usepackage{enumerate} 
\usepackage{paralist} 
\usepackage{amsmath, amsthm, amssymb, mathtools}
\usepackage{mathabx, pifont, stmaryrd} 
\usepackage[explicit]{titlesec} 
\usepackage{etoolbox} 
\usepackage{bibentry} 
\makeatletter\let\saved@bibitem\@bibitem\makeatother 
\usepackage[colorlinks, bookmarksopen, bookmarksnumbered,
            citecolor=red,urlcolor=red]{hyperref} 
\makeatletter\let\@bibitem\saved@bibitem\makeatother 

\usepackage{algorithm}
\usepackage{algorithmic}

\theoremstyle{definition}

%% file: preamble.macros.tex
\usepackage{bm} 
\usepackage{amsfonts} 

\newcommand{\argoptunc}[2]{\underset{#1}{\arg\min} ~~ #2}

\newcommand{\pder}[2]{\ensuremath{\frac{\partial #1}{\partial #2}}}

\newcommand{\Ncal}{\ensuremath{\mathcal{N}}}

\newcommand{\Pcal}{\ensuremath{\mathcal{P}}}

\newcommand{\Ucal}{\ensuremath{\mathcal{U}}}
\newcommand{\Vcal}{\ensuremath{\mathcal{V}}}

\newcommand{\Ebb}{\ensuremath{\mathbb{E}}}

\newcommand{\Nbb}{\ensuremath{\mathbb{N} }}

\newcommand{\Rbb}{\ensuremath{\mathbb{R} }}

\newcommand\Ibm{{\ensuremath{\bm{I}}}}

\newcommand\Rbm{{\ensuremath{\bm{R}}}}

\newcommand\Ubm{{\ensuremath{\bm{U}}}}

\newcommand\Wbm{{\ensuremath{\bm{W}}}}
\newcommand\Xbm{{\ensuremath{\bm{X}}}}
\newcommand\Ybm{{\ensuremath{\bm{Y}}}}
\newcommand\Zbm{{\ensuremath{\bm{Z}}}}

\newcommand\bbm{{\ensuremath{\bm{b}}}}

\newcommand\ebm{{\ensuremath{\bm{e}}}}

\newcommand\ubm{{\ensuremath{\bm{u}}}}
\newcommand\vbm{{\ensuremath{\bm{v}}}}

\newcommand\xbm{{\ensuremath{\bm{x}}}}
\newcommand\ybm{{\ensuremath{\bm{y}}}}
\newcommand\zbm{{\ensuremath{\bm{z}}}}

\newcommand\thetabold{{\ensuremath{\boldsymbol{\theta}}}}



%% file: preamble.tikz.tex
\usepackage{tikz}
\usepackage{pgfplots}
\usepackage{pgfplotstable, filecontents, booktabs}
\pgfplotsset{compat=1.9}

\usetikzlibrary{pgfplots.groupplots}
\usepgfplotslibrary{fillbetween}
\usetikzlibrary{calc,fit,matrix,arrows,automata,positioning,shapes}
\usetikzlibrary{arrows.meta}

\pgfplotsset{select coords between index/.style 2 args={
    x filter/.code={
        \ifnum\coordindex<#1\fi
        \ifnum\coordindex>#2\fi
    }
}}

\tikzset{
 invisible/.style={opacity=0},
 visible on/.style={alt={#1{}{invisible}}},
 alt/.code args={<#1>#2#3}{%
   \alt<#1>{\pgfkeysalso{#2}}{\pgfkeysalso{#3}}
 },
}

\newcommand{\colorbarMatlabJet}[5]{
\begin{tikzpicture}
\begin{axis}[
   hide axis, scale only axis,
   height=0pt, width=0pt,
   colormap={jet}{rgb255=(0,0,131) rgb255=(0,0,135) rgb255=(0,0,139) rgb255=(0,0,143) rgb255=(0,0,147) rgb255=(0,0,151) rgb255=(0,0,155) rgb255=(0,0,159) rgb255=(0,0,163) rgb255=(0,0,167) rgb255=(0,0,171) rgb255=(0,0,175) rgb255=(0,0,179) rgb255=(0,0,183) rgb255=(0,0,187) rgb255=(0,0,191) rgb255=(0,0,195) rgb255=(0,0,199) rgb255=(0,0,203) rgb255=(0,0,207) rgb255=(0,0,211) rgb255=(0,0,215) rgb255=(0,0,219) rgb255=(0,0,223) rgb255=(0,0,227) rgb255=(0,0,231) rgb255=(0,0,235) rgb255=(0,0,239) rgb255=(0,0,243) rgb255=(0,0,247) rgb255=(0,0,251) rgb255=(0,0,255) rgb255=(0,4,255) rgb255=(0,8,255) rgb255=(0,12,255) rgb255=(0,16,255) rgb255=(0,20,255) rgb255=(0,24,255) rgb255=(0,28,255) rgb255=(0,32,255) rgb255=(0,36,255) rgb255=(0,40,255) rgb255=(0,44,255) rgb255=(0,48,255) rgb255=(0,52,255) rgb255=(0,56,255) rgb255=(0,60,255) rgb255=(0,64,255) rgb255=(0,68,255) rgb255=(0,72,255) rgb255=(0,76,255) rgb255=(0,80,255) rgb255=(0,84,255) rgb255=(0,88,255) rgb255=(0,92,255) rgb255=(0,96,255) rgb255=(0,100,255) rgb255=(0,104,255) rgb255=(0,108,255) rgb255=(0,112,255) rgb255=(0,116,255) rgb255=(0,120,255) rgb255=(0,124,255) rgb255=(0,128,255) rgb255=(0,131,255) rgb255=(0,135,255) rgb255=(0,139,255) rgb255=(0,143,255) rgb255=(0,147,255) rgb255=(0,151,255) rgb255=(0,155,255) rgb255=(0,159,255) rgb255=(0,163,255) rgb255=(0,167,255) rgb255=(0,171,255) rgb255=(0,175,255) rgb255=(0,179,255) rgb255=(0,183,255) rgb255=(0,187,255) rgb255=(0,191,255) rgb255=(0,195,255) rgb255=(0,199,255) rgb255=(0,203,255) rgb255=(0,207,255) rgb255=(0,211,255) rgb255=(0,215,255) rgb255=(0,219,255) rgb255=(0,223,255) rgb255=(0,227,255) rgb255=(0,231,255) rgb255=(0,235,255) rgb255=(0,239,255) rgb255=(0,243,255) rgb255=(0,247,255) rgb255=(0,251,255) rgb255=(0,255,255) rgb255=(4,255,251) rgb255=(8,255,247) rgb255=(12,255,243) rgb255=(16,255,239) rgb255=(20,255,235) rgb255=(24,255,231) rgb255=(28,255,227) rgb255=(32,255,223) rgb255=(36,255,219) rgb255=(40,255,215) rgb255=(44,255,211) rgb255=(48,255,207) rgb255=(52,255,203) rgb255=(56,255,199) rgb255=(60,255,195) rgb255=(64,255,191) rgb255=(68,255,187) rgb255=(72,255,183) rgb255=(76,255,179) rgb255=(80,255,175) rgb255=(84,255,171) rgb255=(88,255,167) rgb255=(92,255,163) rgb255=(96,255,159) rgb255=(100,255,155) rgb255=(104,255,151) rgb255=(108,255,147) rgb255=(112,255,143) rgb255=(116,255,139) rgb255=(120,255,135) rgb255=(124,255,131) rgb255=(128,255,128) rgb255=(131,255,124) rgb255=(135,255,120) rgb255=(139,255,116) rgb255=(143,255,112) rgb255=(147,255,108) rgb255=(151,255,104) rgb255=(155,255,100) rgb255=(159,255,96) rgb255=(163,255,92) rgb255=(167,255,88) rgb255=(171,255,84) rgb255=(175,255,80) rgb255=(179,255,76) rgb255=(183,255,72) rgb255=(187,255,68) rgb255=(191,255,64) rgb255=(195,255,60) rgb255=(199,255,56) rgb255=(203,255,52) rgb255=(207,255,48) rgb255=(211,255,44) rgb255=(215,255,40) rgb255=(219,255,36) rgb255=(223,255,32) rgb255=(227,255,28) rgb255=(231,255,24) rgb255=(235,255,20) rgb255=(239,255,16) rgb255=(243,255,12) rgb255=(247,255,8) rgb255=(251,255,4) rgb255=(255,255,0) rgb255=(255,251,0) rgb255=(255,247,0) rgb255=(255,243,0) rgb255=(255,239,0) rgb255=(255,235,0) rgb255=(255,231,0) rgb255=(255,227,0) rgb255=(255,223,0) rgb255=(255,219,0) rgb255=(255,215,0) rgb255=(255,211,0) rgb255=(255,207,0) rgb255=(255,203,0) rgb255=(255,199,0) rgb255=(255,195,0) rgb255=(255,191,0) rgb255=(255,187,0) rgb255=(255,183,0) rgb255=(255,179,0) rgb255=(255,175,0) rgb255=(255,171,0) rgb255=(255,167,0) rgb255=(255,163,0) rgb255=(255,159,0) rgb255=(255,155,0) rgb255=(255,151,0) rgb255=(255,147,0) rgb255=(255,143,0) rgb255=(255,139,0) rgb255=(255,135,0) rgb255=(255,131,0) rgb255=(255,128,0) rgb255=(255,124,0) rgb255=(255,120,0) rgb255=(255,116,0) rgb255=(255,112,0) rgb255=(255,108,0) rgb255=(255,104,0) rgb255=(255,100,0) rgb255=(255,96,0) rgb255=(255,92,0) rgb255=(255,88,0) rgb255=(255,84,0) rgb255=(255,80,0) rgb255=(255,76,0) rgb255=(255,72,0) rgb255=(255,68,0) rgb255=(255,64,0) rgb255=(255,60,0) rgb255=(255,56,0) rgb255=(255,52,0) rgb255=(255,48,0) rgb255=(255,44,0) rgb255=(255,40,0) rgb255=(255,36,0) rgb255=(255,32,0) rgb255=(255,28,0) rgb255=(255,24,0) rgb255=(255,20,0) rgb255=(255,16,0) rgb255=(255,12,0) rgb255=(255,8,0) rgb255=(255,4,0) rgb255=(255,0,0) rgb255=(251,0,0) rgb255=(247,0,0) rgb255=(243,0,0) rgb255=(239,0,0) rgb255=(235,0,0) rgb255=(231,0,0) rgb255=(227,0,0) rgb255=(223,0,0) rgb255=(219,0,0) rgb255=(215,0,0) rgb255=(211,0,0) rgb255=(207,0,0) rgb255=(203,0,0) rgb255=(199,0,0) rgb255=(195,0,0) rgb255=(191,0,0) rgb255=(187,0,0) rgb255=(183,0,0) rgb255=(179,0,0) rgb255=(175,0,0) rgb255=(171,0,0) rgb255=(167,0,0) rgb255=(163,0,0) rgb255=(159,0,0) rgb255=(155,0,0) rgb255=(151,0,0) rgb255=(147,0,0) rgb255=(143,0,0) rgb255=(139,0,0) rgb255=(135,0,0) rgb255=(131,0,0) rgb255=(128,0,0) },
   colorbar horizontal,
   point meta min=#1, point meta max=#5,
   colorbar style={width=10cm, xtick={#1, #2, #3, #4, #5}}
]
\addplot [draw=none] coordinates {(0,0)};
\end{axis}
\end{tikzpicture}
}

%% file: figures/pivotal_nodes.tikz
\begin{tikzpicture}
\begin{axis}[
axis equal image,
width=1\textwidth,
xtick={-4, 0, 30},
ytick={-5, 0, 5},
xlabel=$x$,
ymax=5,
xmax=30,
ylabel=$y$,
xmin=-4,
ymin=-5]
\addplot [mark options={solid, thick}, mark=*, mark size=0.3, blue, only marks]
coordinates {
( 3.99000000e+00, -4.37500000e+00)
( 5.01000000e+00, -4.12500000e+00)
( 5.69000000e+00, -4.87500000e+00)
( 6.71000000e+00, -4.62500000e+00)
( 7.73000000e+00, -4.37500000e+00)
( 8.75000000e+00, -4.12500000e+00)
( 9.43000000e+00, -4.87500000e+00)
( 1.04500000e+01, -4.62500000e+00)
( 1.14700000e+01, -4.37500000e+00)
( 1.07900000e+01, -3.62500000e+00)
( 1.24900000e+01, -4.12500000e+00)
( 1.31700000e+01, -4.87500000e+00)
( 1.41900000e+01, -4.62500000e+00)
( 1.52100000e+01, -4.37500000e+00)
( 1.62300000e+01, -4.12500000e+00)
( 1.69100000e+01, -4.87500000e+00)
( 1.79300000e+01, -4.62500000e+00)
( 1.89500000e+01, -4.37500000e+00)
( 1.99700000e+01, -4.12500000e+00)
( 2.06500000e+01, -4.87500000e+00)
( 2.09900000e+01, -3.87500000e+00)
( 2.16700000e+01, -4.62500000e+00)
( 2.26900000e+01, -4.37500000e+00)
( 2.37100000e+01, -4.12500000e+00)
( 2.43900000e+01, -4.87500000e+00)
( 2.54100000e+01, -4.62500000e+00)
( 2.64300000e+01, -4.37500000e+00)
( 2.74500000e+01, -4.12500000e+00)
( 2.81300000e+01, -4.87500000e+00)
( 2.84700000e+01, -3.87500000e+00)
( 2.91500000e+01, -4.62500000e+00)
( 2.94900000e+01, -3.62500000e+00)
( 2.97000000e+00, -4.62500000e+00)
( 4.33000000e+00, -3.37500000e+00)
( 6.03000000e+00, -3.87500000e+00)
( 7.05000000e+00, -3.62500000e+00)
( 8.07000000e+00, -3.37500000e+00)
( 9.09000000e+00, -3.12500000e+00)
( 9.77000000e+00, -3.87500000e+00)
( 1.01100000e+01, -2.87500000e+00)
( 1.18100000e+01, -3.37500000e+00)
( 1.11300000e+01, -2.62500000e+00)
( 1.35100000e+01, -3.87500000e+00)
( 1.28300000e+01, -3.12500000e+00)
( 1.45300000e+01, -3.62500000e+00)
( 1.55500000e+01, -3.37500000e+00)
( 1.65700000e+01, -3.12500000e+00)
( 1.72500000e+01, -3.87500000e+00)
( 1.82700000e+01, -3.62500000e+00)
( 1.92900000e+01, -3.37500000e+00)
( 1.86100000e+01, -2.62500000e+00)
( 2.03100000e+01, -3.12500000e+00)
( 2.20100000e+01, -3.62500000e+00)
( 2.13300000e+01, -2.87500000e+00)
( 2.30300000e+01, -3.37500000e+00)
( 2.40500000e+01, -3.12500000e+00)
( 2.47300000e+01, -3.87500000e+00)
( 2.57500000e+01, -3.62500000e+00)
( 2.67700000e+01, -3.37500000e+00)
( 2.60900000e+01, -2.62500000e+00)
( 2.77900000e+01, -3.12500000e+00)
( 2.88100000e+01, -2.87500000e+00)
( 2.98300000e+01, -2.62500000e+00)
( 2.91500000e+01, -1.87500000e+00)
( 3.31000000e+00, -3.62500000e+00)
( 5.35000000e+00, -3.12500000e+00)
( 6.37000000e+00, -2.87500000e+00)
( 7.39000000e+00, -2.62500000e+00)
( 8.41000000e+00, -2.37500000e+00)
( 9.43000000e+00, -2.12500000e+00)
( 7.73000000e+00, -1.62500000e+00)
( 1.04500000e+01, -1.87500000e+00)
( 1.21500000e+01, -2.37500000e+00)
( 1.14700000e+01, -1.62500000e+00)
( 1.38500000e+01, -2.87500000e+00)
( 1.31700000e+01, -2.12500000e+00)
( 1.48700000e+01, -2.62500000e+00)
( 1.58900000e+01, -2.37500000e+00)
( 1.75900000e+01, -2.87500000e+00)
( 1.69100000e+01, -2.12500000e+00)
( 1.79300000e+01, -1.87500000e+00)
( 1.96300000e+01, -2.37500000e+00)
( 1.89500000e+01, -1.62500000e+00)
( 2.06500000e+01, -2.12500000e+00)
( 2.23500000e+01, -2.62500000e+00)
( 2.16700000e+01, -1.87500000e+00)
( 2.33700000e+01, -2.37500000e+00)
( 2.50700000e+01, -2.87500000e+00)
( 2.43900000e+01, -2.12500000e+00)
( 2.54100000e+01, -1.87500000e+00)
( 2.71100000e+01, -2.37500000e+00)
( 2.64300000e+01, -1.62500000e+00)
( 2.81300000e+01, -2.12500000e+00)
( 2.74500000e+01, -1.37500000e+00)
( 2.84700000e+01, -1.12500000e+00)
( 2.94900000e+01, -8.75000000e-01)
( 3.65000000e+00, -2.62500000e+00)
( 4.67000000e+00, -2.37500000e+00)
( 5.69000000e+00, -2.12500000e+00)
( 6.71000000e+00, -1.87500000e+00)
( 8.75000000e+00, -1.37500000e+00)
( 7.05000000e+00, -8.75000000e-01)
( 9.77000000e+00, -1.12500000e+00)
( 1.07900000e+01, -8.75000000e-01)
( 1.24900000e+01, -1.37500000e+00)
( 1.18100000e+01, -6.25000000e-01)
( 1.41900000e+01, -1.87500000e+00)
( 1.35100000e+01, -1.12500000e+00)
( 1.52100000e+01, -1.62500000e+00)
( 1.62300000e+01, -1.37500000e+00)
( 1.45300000e+01, -8.75000000e-01)
( 1.72500000e+01, -1.12500000e+00)
( 1.82700000e+01, -8.75000000e-01)
( 1.99700000e+01, -1.37500000e+00)
( 1.92900000e+01, -6.25000000e-01)
( 2.09900000e+01, -1.12500000e+00)
( 2.26900000e+01, -1.62500000e+00)
( 2.20100000e+01, -8.75000000e-01)
( 2.37100000e+01, -1.37500000e+00)
( 2.47300000e+01, -1.12500000e+00)
( 2.30300000e+01, -6.25000000e-01)
( 2.57500000e+01, -8.75000000e-01)
( 2.67700000e+01, -6.25000000e-01)
( 2.77900000e+01, -3.75000000e-01)
( 2.88100000e+01, -1.25000000e-01)
( 2.98300000e+01,  1.25000000e-01)
( 2.81300000e+01,  6.25000000e-01)
( 2.91500000e+01,  8.75000000e-01)
( 2.63000000e+00, -2.87500000e+00)
( 3.99000000e+00, -1.62500000e+00)
( 5.01000000e+00, -1.37500000e+00)
( 6.03000000e+00, -1.12500000e+00)
( 8.07000000e+00, -6.25000000e-01)
( 9.09000000e+00, -3.75000000e-01)
( 1.01100000e+01, -1.25000000e-01)
( 1.11300000e+01,  1.25000000e-01)
( 1.28300000e+01, -3.75000000e-01)
( 1.21500000e+01,  3.75000000e-01)
( 1.38500000e+01, -1.25000000e-01)
( 1.55500000e+01, -6.25000000e-01)
( 1.48700000e+01,  1.25000000e-01)
( 1.65700000e+01, -3.75000000e-01)
( 1.58900000e+01,  3.75000000e-01)
( 1.75900000e+01, -1.25000000e-01)
( 1.86100000e+01,  1.25000000e-01)
( 2.03100000e+01, -3.75000000e-01)
( 1.96300000e+01,  3.75000000e-01)
( 2.13300000e+01, -1.25000000e-01)
( 2.23500000e+01,  1.25000000e-01)
( 2.40500000e+01, -3.75000000e-01)
( 2.33700000e+01,  3.75000000e-01)
( 2.50700000e+01, -1.25000000e-01)
( 2.43900000e+01,  6.25000000e-01)
( 2.60900000e+01,  1.25000000e-01)
( 2.71100000e+01,  3.75000000e-01)
( 2.54100000e+01,  8.75000000e-01)
( 2.64300000e+01,  1.12500000e+00)
( 2.74500000e+01,  1.37500000e+00)
( 2.84700000e+01,  1.62500000e+00)
( 2.94900000e+01,  1.87500000e+00)
( 2.97000000e+00, -1.87500000e+00)
( 2.29000000e+00, -3.87500000e+00)
( 4.33000000e+00, -6.25000000e-01)
( 5.35000000e+00, -3.75000000e-01)
( 6.37000000e+00, -1.25000000e-01)
( 7.39000000e+00,  1.25000000e-01)
( 8.41000000e+00,  3.75000000e-01)
( 9.43000000e+00,  6.25000000e-01)
( 1.04500000e+01,  8.75000000e-01)
( 1.31700000e+01,  6.25000000e-01)
( 1.14700000e+01,  1.12500000e+00)
( 1.41900000e+01,  8.75000000e-01)
( 1.52100000e+01,  1.12500000e+00)
( 1.69100000e+01,  6.25000000e-01)
( 1.62300000e+01,  1.37500000e+00)
( 1.79300000e+01,  8.75000000e-01)
( 1.89500000e+01,  1.12500000e+00)
( 2.06500000e+01,  6.25000000e-01)
( 1.99700000e+01,  1.37500000e+00)
( 2.16700000e+01,  8.75000000e-01)
( 2.26900000e+01,  1.12500000e+00)
( 2.09900000e+01,  1.62500000e+00)
( 2.37100000e+01,  1.37500000e+00)
( 2.47300000e+01,  1.62500000e+00)
( 2.57500000e+01,  1.87500000e+00)
( 2.30300000e+01,  2.12500000e+00)
( 2.67700000e+01,  2.12500000e+00)
( 2.77900000e+01,  2.37500000e+00)
( 2.88100000e+01,  2.62500000e+00)
( 2.98300000e+01,  2.87500000e+00)
( 2.71100000e+01,  3.12500000e+00)
( 2.81300000e+01,  3.37500000e+00)
( 3.22500000e+00, -1.06250000e+00)
( 3.22500000e+00, -9.37500000e-01)
( 2.88500000e+00, -1.06250000e+00)
( 5.69000000e+00,  6.25000000e-01)
( 6.71000000e+00,  8.75000000e-01)
( 7.73000000e+00,  1.12500000e+00)
( 8.75000000e+00,  1.37500000e+00)
( 9.77000000e+00,  1.62500000e+00)
( 1.24900000e+01,  1.37500000e+00)
( 1.07900000e+01,  1.87500000e+00)
( 1.35100000e+01,  1.62500000e+00)
( 1.45300000e+01,  1.87500000e+00)
( 1.55500000e+01,  2.12500000e+00)
( 1.72500000e+01,  1.62500000e+00)
( 1.65700000e+01,  2.37500000e+00)
( 1.82700000e+01,  1.87500000e+00)
( 1.92900000e+01,  2.12500000e+00)
( 1.75900000e+01,  2.62500000e+00)
( 2.03100000e+01,  2.37500000e+00)
( 2.20100000e+01,  1.87500000e+00)
( 2.13300000e+01,  2.62500000e+00)
( 2.23500000e+01,  2.87500000e+00)
( 2.40500000e+01,  2.37500000e+00)
( 2.50700000e+01,  2.62500000e+00)
( 2.33700000e+01,  3.12500000e+00)
( 2.60900000e+01,  2.87500000e+00)
( 2.43900000e+01,  3.37500000e+00)
( 2.54100000e+01,  3.62500000e+00)
( 2.91500000e+01,  3.62500000e+00)
( 2.64300000e+01,  3.87500000e+00)
( 2.74500000e+01,  4.12500000e+00)
( 2.84700000e+01,  4.37500000e+00)
( 2.37500000e+00, -1.43750000e+00)
( 2.75750000e+00, -9.68750000e-01)
( 4.67000000e+00,  3.75000000e-01)
( 3.65000000e+00,  1.25000000e-01)
( 6.03000000e+00,  1.62500000e+00)
( 7.05000000e+00,  1.87500000e+00)
( 8.07000000e+00,  2.12500000e+00)
( 9.09000000e+00,  2.37500000e+00)
( 1.18100000e+01,  2.12500000e+00)
( 1.28300000e+01,  2.37500000e+00)
( 1.38500000e+01,  2.62500000e+00)
( 1.48700000e+01,  2.87500000e+00)
( 1.21500000e+01,  3.12500000e+00)
( 1.58900000e+01,  3.12500000e+00)
( 1.69100000e+01,  3.37500000e+00)
( 1.86100000e+01,  2.87500000e+00)
( 1.96300000e+01,  3.12500000e+00)
( 1.79300000e+01,  3.62500000e+00)
( 2.06500000e+01,  3.37500000e+00)
( 2.16700000e+01,  3.62500000e+00)
( 1.89500000e+01,  3.87500000e+00)
( 2.26900000e+01,  3.87500000e+00)
( 2.37100000e+01,  4.12500000e+00)
( 2.47300000e+01,  4.37500000e+00)
( 2.57500000e+01,  4.62500000e+00)
( 2.30300000e+01,  4.87500000e+00)
( 2.67700000e+01,  4.87500000e+00)
( 2.94900000e+01,  4.62500000e+00)
( 2.20100000e+01,  4.62500000e+00)
( 2.09900000e+01,  4.37500000e+00)
( 1.99700000e+01,  4.12500000e+00)
( 1.92900000e+01,  4.87500000e+00)
( 2.75750000e+00, -9.06250000e-01)
( 2.92750000e+00, -5.31250000e-01)
( 3.39500000e+00, -6.25000000e-02)
( 3.05500000e+00, -1.87500000e-01)
( 5.01000000e+00,  1.37500000e+00)
( 6.37000000e+00,  2.62500000e+00)
( 1.01100000e+01,  2.62500000e+00)
( 1.11300000e+01,  2.87500000e+00)
( 9.43000000e+00,  3.37500000e+00)
( 1.31700000e+01,  3.37500000e+00)
( 1.04500000e+01,  3.62500000e+00)
( 1.41900000e+01,  3.62500000e+00)
( 1.52100000e+01,  3.87500000e+00)
( 1.62300000e+01,  4.12500000e+00)
( 1.72500000e+01,  4.37500000e+00)
( 1.45300000e+01,  4.62500000e+00)
( 1.82700000e+01,  4.62500000e+00)
( 1.55500000e+01,  4.87500000e+00)
( 1.35100000e+01,  4.37500000e+00)
( 1.24900000e+01,  4.12500000e+00)
( 1.14700000e+01,  3.87500000e+00)
( 1.18100000e+01,  4.87500000e+00)
( 1.07900000e+01,  4.62500000e+00)
( 9.77000000e+00,  4.37500000e+00)
( 8.41000000e+00,  3.12500000e+00)
( 8.75000000e+00,  4.12500000e+00)
( 7.73000000e+00,  3.87500000e+00)
( 8.07000000e+00,  4.87500000e+00)
( 7.39000000e+00,  2.87500000e+00)
( 7.05000000e+00,  4.62500000e+00)
( 6.71000000e+00,  3.62500000e+00)
( 6.03000000e+00,  4.37500000e+00)
( 2.50250000e+00, -9.06250000e-01)
( 2.67250000e+00, -7.18750000e-01)
( 2.92750000e+00, -1.56250000e-01)
( 3.99000000e+00,  1.12500000e+00)
( 3.39500000e+00,  8.12500000e-01)
( 5.35000000e+00,  2.37500000e+00)
( 4.33000000e+00,  2.12500000e+00)
( 5.69000000e+00,  3.37500000e+00)
( 4.67000000e+00,  3.12500000e+00)
( 2.92750000e+00,  4.06250000e-01)
( 5.01000000e+00,  4.12500000e+00)
( 3.31000000e+00,  1.87500000e+00)
( 2.92750000e+00,  7.81250000e-01)
( 3.65000000e+00,  2.87500000e+00)
( 3.99000000e+00,  3.87500000e+00)
( 2.67250000e+00, -3.43750000e-01)
( 4.33000000e+00,  4.87500000e+00)
( 2.84250000e+00,  7.18750000e-01)
( 2.84250000e+00,  7.81250000e-01)
( 2.75750000e+00,  9.68750000e-01)
( 2.67250000e+00,  5.93750000e-01)
( 2.58750000e+00,  9.37500000e-02)
( 2.71500000e+00,  1.06250000e+00)
( 2.58750000e+00,  2.18750000e-01)
( 2.67250000e+00,  9.68750000e-01)
( 2.50250000e+00, -2.18750000e-01)
( 2.50250000e+00, -9.37500000e-02)
( 2.50250000e+00, -3.12500000e-02)
( 3.31000000e+00,  4.62500000e+00)
( 2.97000000e+00,  3.62500000e+00)
( 2.41750000e+00,  2.18750000e-01)
( 2.41750000e+00,  4.06250000e-01)
( 2.20500000e+00, -1.18750000e+00)
( 2.16250000e+00, -8.43750000e-01)
( 2.24750000e+00, -5.31250000e-01)
( 2.16250000e+00, -7.81250000e-01)
( 2.33250000e+00, -1.56250000e-01)
( 2.16250000e+00, -6.56250000e-01)
( 2.33250000e+00,  2.18750000e-01)
( 2.03500000e+00, -1.18750000e+00)
( 2.24750000e+00,  3.43750000e-01)
( 2.07750000e+00, -5.31250000e-01)
( 2.07750000e+00, -3.43750000e-01)
( 2.24750000e+00,  6.56250000e-01)
( 2.37500000e+00,  1.31250000e+00)
( 1.99250000e+00, -9.06250000e-01)
( 1.95000000e+00, -2.12500000e+00)
( 2.63000000e+00,  2.62500000e+00)
( 2.07750000e+00,  2.81250000e-01)
( 1.99250000e+00, -5.31250000e-01)
( 2.29000000e+00,  1.62500000e+00)
( 2.07750000e+00,  3.43750000e-01)
( 1.99250000e+00,  5.31250000e-01)
( 1.90750000e+00, -9.37500000e-02)
( 1.99250000e+00,  9.06250000e-01)
( 1.90750000e+00,  5.93750000e-01)
( 1.82250000e+00, -4.06250000e-01)
( 1.95000000e+00, -4.87500000e+00)
( 1.82250000e+00,  5.31250000e-01)
( 1.73750000e+00, -4.68750000e-01)
( 1.73750000e+00, -4.06250000e-01)
( 1.65250000e+00, -2.18750000e-01)
( 2.29000000e+00,  4.37500000e+00)
( 1.65250000e+00,  1.56250000e-01)
( 1.56750000e+00, -9.68750000e-01)
( 1.56750000e+00, -8.43750000e-01)
( 1.48250000e+00, -4.68749000e-01)
( 1.48249000e+00, -1.56249000e-01)
( 1.61000000e+00, -3.12500000e+00)
( 1.31250000e+00, -9.68750000e-01)
( 1.31250000e+00, -5.93749000e-01)
( 1.56750000e+00,  7.18750000e-01)
( 1.48250000e+00,  4.06249000e-01)
( 1.56750000e+00,  8.43750000e-01)
( 1.48250000e+00,  4.68750000e-01)
( 1.22750000e+00, -7.18750000e-01)
( 1.39750000e+00,  7.18750000e-01)
( 1.31249000e+00,  3.43747000e-01)
( 1.39750000e+00,  9.06250000e-01)
( 1.18500000e+00, -1.43750000e+00)
( 1.95000000e+00,  3.37500000e+00)
( 1.61000000e+00,  2.37500000e+00)
( 1.31250000e+00,  7.18750000e-01)
( 1.22748000e+00, -3.12492000e-02)
( 1.35500000e+00,  1.18750000e+00)
( 1.14248000e+00, -3.43743000e-01)
( 1.14247000e+00, -2.81242000e-01)
( 1.27000000e+00, -4.12500000e+00)
( 1.14246000e+00,  9.37452000e-02)
( 1.18500000e+00,  1.31250000e+00)
( 9.93750000e-01, -7.96875000e-01)
( 9.93740000e-01, -5.15619000e-01)
( 9.93716000e-01, -3.59359000e-01)
( 1.05750000e+00,  7.81250000e-01)
( 1.05750000e+00,  8.43750000e-01)
( 9.93670000e-01, -4.68694000e-02)
( 9.08749000e-01, -7.65624000e-01)
( 9.51214000e-01, -3.90607000e-01)
( 9.08728000e-01, -4.84361000e-01)
( 8.66249000e-01, -7.65624000e-01)
( 9.08665000e-01, -2.96839000e-01)
( 9.93689000e-01,  2.34356000e-01)
( 9.93702000e-01,  2.96857000e-01)
( 9.08620000e-01, -1.40597000e-01)
( 8.66233000e-01, -5.46863000e-01)
( 9.08607000e-01, -1.56215000e-02)
( 9.30000000e-01, -2.37500000e+00)
( 9.08612000e-01,  7.81081000e-02)
( 9.51241000e-01,  5.46869000e-01)
( 8.66071000e-01, -1.40585000e-01)
( 9.08702000e-01,  3.90600000e-01)
( 9.08712000e-01,  4.21854000e-01)
( 8.66054000e-01, -1.56201000e-02)
( 9.08738000e-01,  5.46866000e-01)
( 1.01500000e+00,  1.43750000e+00)
( 9.51250000e-01,  9.53125000e-01)
( 9.72500000e-01,  1.15625000e+00)
( 1.01500000e+00,  1.68750000e+00)
( 8.66087000e-01,  2.03074000e-01)
( 8.23748000e-01, -7.34373000e-01)
( 8.23743000e-01, -6.40618000e-01)
( 8.02500000e-01, -1.03125000e+00)
( 8.66246000e-01,  6.71872000e-01)
( 8.66249000e-01,  7.65624000e-01)
( 8.66250000e-01,  8.90625000e-01)
( 8.23522000e-01,  2.03051000e-01)
( 1.27000000e+00,  4.12500000e+00)
( 7.81250000e-01, -9.21875000e-01)
( 7.81249000e-01, -7.96873000e-01)
( 7.91560000e-01, -1.32740000e-01)
( 7.38712000e-01, -5.46843000e-01)
( 7.38698000e-01, -5.15583000e-01)
( 7.91537000e-01,  3.90398000e-02)
( 7.48972000e-01, -2.26410000e-01)
( 7.48890000e-01, -3.90303000e-02)
( 7.80982000e-01,  2.65513000e-01)
( 7.81021000e-01,  2.96770000e-01)
( 7.81095000e-01,  3.59289000e-01)
( 7.17500000e-01, -1.03125000e+00)
( 7.48916000e-01,  1.32704000e-01)
( 7.27550000e-01, -1.17065000e-01)
( 8.23750000e-01,  9.53125000e-01)
( 7.48972000e-01,  2.26410000e-01)
( 7.48997000e-01,  2.42038000e-01)
( 7.06253000e-01, -2.10710000e-01)
( 6.96249000e-01, -8.28123000e-01)
( 6.96250000e-01, -9.53125000e-01)
( 7.38741000e-01,  6.71865000e-01)
( 6.85377000e-01, -3.82651000e-01)
( 7.38746000e-01,  7.34370000e-01)
( 7.06414000e-01,  2.73226000e-01)
( 6.85110000e-01, -2.88808000e-01)
( 7.38750000e-01,  9.53125000e-01)
( 6.84710000e-01, -7.80104000e-03)
( 6.84720000e-01,  2.34027000e-02)
( 6.84759000e-01,  7.02006000e-02)
( 6.84757000e-01,  1.01395000e-01)
( 6.85208000e-01,  3.20085000e-01)
( 9.30000000e-01,  3.12500000e+00)
( 6.63609000e-01, -2.57468000e-01)
( 6.41913000e-01, -1.94880000e-01)
( 6.53749000e-01, -8.90624000e-01)
( 6.63311000e-01,  1.32543000e-01)
( 6.41931000e-01, -5.45715000e-02)
( 6.41894000e-01,  3.89833000e-02)
( 6.21722000e-01, -4.76430000e-01)
( 6.21788000e-01, -5.23353000e-01)
( 6.64079000e-01,  3.82616000e-01)
( 6.20664000e-01, -2.41655000e-01)
( 6.42114000e-01,  2.41750000e-01)
( 6.96249000e-01,  8.28123000e-01)
( 6.20353000e-01, -1.63585000e-01)
( 6.20515000e-01, -8.57051000e-02)
( 6.53709000e-01,  5.78082000e-01)
( 6.11199000e-01, -5.78068000e-01)
( 6.11225000e-01, -6.40592000e-01)
( 6.11232000e-01, -6.71850000e-01)
( 6.20377000e-01,  1.79181000e-01)
( 6.20575000e-01,  2.26025000e-01)
( 6.20758000e-01,  2.57290000e-01)
( 6.53749000e-01,  8.90624000e-01)
( 5.98922000e-01, -1.94686000e-01)
( 5.98901000e-01, -1.32356000e-01)
( 6.75000000e-01,  1.43750000e+00)
( 5.99123000e-01,  8.56760000e-02)
( 5.99310000e-01,  2.57185000e-01)
( 6.75000000e-01,  1.68750000e+00)
( 5.77506000e-01, -2.10190000e-01)
( 5.77314000e-01, -1.63400000e-01)
( 5.77517000e-01,  1.16742000e-01)
( 5.77314000e-01,  1.63400000e-01)
( 5.77741000e-01,  2.41440000e-01)
( 5.55822000e-01, -1.47751000e-01)
( 5.56183000e-01, -1.01150000e-01)
( 5.56780000e-01, -3.03884000e-01)
( 5.78547000e-01,  3.51004000e-01)
( 5.68735000e-01, -7.03101000e-01)
( 5.79240000e-01,  5.07678000e-01)
( 5.79266000e-01,  5.23323000e-01)
( 5.57929000e-01,  4.91994000e-01)
( 5.68750000e-01, -9.53124000e-01)
( 5.58040000e-01,  5.54587000e-01)
( 5.68735000e-01,  7.03101000e-01)
( 5.34541000e-01, -1.32186000e-01)
( 5.35903000e-01, -3.66465000e-01)
( 5.28168000e-01,  1.16775000e-02)
( 5.20646000e-01, -5.85411000e-02)
( 5.20954000e-01,  3.52216000e-02)
( 5.18897000e-01, -1.20591000e-01)
( 5.18091000e-01, -1.67020000e-01)
( 5.19372000e-01,  7.39607000e-02)
( 5.18897000e-01,  1.20591000e-01)
( 5.18085000e-01,  1.59237000e-01)
( 5.14667000e-01, -3.82046000e-01)
( 5.14797000e-01, -3.97748000e-01)
( 5.15516000e-01, -5.54546000e-01)
( 5.47500000e-01,  1.15625000e+00)
( 5.13154000e-01,  2.09896000e-01)
( 5.13710000e-01,  2.72355000e-01)
( 5.10254000e-01,  2.73867000e-02)
( 5.14531000e-01,  3.66348000e-01)
( 5.10268000e-01,  5.85793000e-02)
( 5.14797000e-01,  3.97748000e-01)
( 5.26247000e-01,  8.28119000e-01)
( 5.05315000e-01, -3.74115000e-03)
( 5.08187000e-01, -1.05010000e-01)
( 5.07694000e-01, -9.71330000e-02)
( 5.15537000e-01,  5.70193000e-01)
( 5.15552000e-01,  5.85834000e-01)
( 4.92690000e-01,  3.19159000e-01)
( 5.26249000e-01,  8.90622000e-01)
( 4.96864000e-01, -1.90304000e-01)
( 4.86302000e-01,  1.59258000e-01)
( 4.85279000e-01,  1.50801000e-01)
( 4.76279000e-01,  2.37183000e-01)
( 4.75718000e-01, -1.90323000e-01)
( 4.71521000e-01,  3.34745000e-01)
( 4.76401000e-01, -2.52787000e-01)
( 4.72516000e-01,  4.44704000e-01)
( 4.72849000e-01,  5.07472000e-01)
( 5.90000000e-01,  2.12500000e+00)
( 4.71334000e-01, -3.19051000e-01)
( 4.65869000e-01, -2.45055000e-01)
( 4.72787000e-01, -4.91793000e-01)
( 4.65489000e-01, -2.29310000e-01)
( 4.65846000e-01, -2.52810000e-01)
( 4.72980000e-01, -5.54478000e-01)
( 4.57216000e-01,  2.07181000e-01)
( 4.54854000e-01,  2.68101000e-01)
( 5.05000000e-01, -1.43750000e+00)
( 4.55292000e-01, -2.37293000e-01)
( 4.55075000e-01, -2.60469000e-01)
( 4.51819000e-01,  6.32708000e-01)
( 4.50772000e-01, -3.97423000e-01)
( 4.51567000e-01, -5.07410000e-01)
( 4.37086000e-01, -2.46669000e-01)
( 4.29641000e-01,  4.13094000e-01)
( 4.23077000e-01,  2.99200000e-01)
( 4.41248000e-01,  8.90621000e-01)
( 4.22834000e-01,  2.91255000e-01)
( 4.13301000e-01,  3.15355000e-01)
( 4.33281000e-01, -2.75633000e-01)
( 4.22893000e-01, -2.75597000e-01)
( 4.22098000e-01, -2.82900000e-01)
( 4.29641000e-01, -4.13094000e-01)
( 3.87175000e-01,  4.13067000e-01)
( 4.11827000e-01, -2.98662000e-01)
( 4.41226000e-01, -7.03076000e-01)
( 3.86579000e-01,  3.81270000e-01)
( 4.07757000e-01, -3.65686000e-01)
( 4.08377000e-01, -4.13049000e-01)
( 4.02878000e-01, -3.23250000e-01)
( 3.83486000e-01,  3.24212000e-01)
( 4.41250000e-01, -9.53124000e-01)
( 4.08943000e-01, -4.91584000e-01)
( 4.09057000e-01, -5.22940000e-01)
( 3.95123000e-01, -3.09533000e-01)
( 4.09303000e-01, -6.32668000e-01)
( 3.91699000e-01, -3.53840000e-01)
( 3.86579000e-01, -3.81270000e-01)
( 3.70367000e-01,  3.45973000e-01)
( 3.87552000e-01, -4.60191000e-01)
( 3.70201000e-01,  3.61308000e-01)
( 3.66586000e-01,  5.54134000e-01)
( 3.66295000e-01,  4.44558000e-01)
( 3.66804000e-01,  6.48278000e-01)
( 3.88086000e-01, -6.79600000e-01)
( 3.62205000e-01, -3.47509000e-01)
( 3.45592000e-01,  6.95228000e-01)
( 3.45580000e-01,  6.79578000e-01)
( 3.45450000e-01,  6.01180000e-01)
( 3.38826000e-01,  3.77389000e-01)
( 3.24348000e-01,  7.10866000e-01)
( 3.24309000e-01,  6.63896000e-01)
( 3.35000000e-01,  1.31250000e+00)
( 3.13749000e-01,  9.53123000e-01)
( 3.35000000e-01,  1.56250000e+00)
( 3.08333000e-01,  4.25588000e-01)
( 3.60189000e-01, -3.85289000e-01)
( 3.02639000e-01,  5.38026000e-01)
( 3.49499000e-01, -3.61470000e-01)
( 3.38826000e-01, -3.77389000e-01)
( 3.45090000e-01, -4.91255000e-01)
( 3.29363000e-01, -3.94176000e-01)
( 2.86624000e-01,  4.40412000e-01)
( 2.81643000e-01,  6.00978000e-01)
( 3.77500000e-01, -1.15625000e+00)
( 3.56250000e-01, -9.84373000e-01)
( 2.60602000e-01,  7.26489000e-01)
( 3.02717000e-01, -5.53786000e-01)
( 2.60317000e-01,  5.85141000e-01)
( 2.60243000e-01,  5.69356000e-01)
( 3.03104000e-01, -7.26500000e-01)
( 2.54125000e-01,  4.54812000e-01)
( 5.90000000e-01,  4.87500000e+00)
( 3.35000000e-01, -1.31250000e+00)
( 2.81157000e-01, -5.06315000e-01)
( 2.81582000e-01, -5.85215000e-01)
( 2.65016000e-01, -4.39779000e-01)
( 2.39225000e-01,  6.32357000e-01)
( 2.28749000e-01,  9.21871000e-01)
( 2.28748000e-01,  8.90618000e-01)
( 2.07500000e-01,  1.21875000e+00)
( 2.18062000e-01,  6.79464000e-01)
( 2.32857000e-01,  4.46979000e-01)
( 2.17929000e-01,  6.16565000e-01)
( 2.17656000e-01,  5.53390000e-01)
( 2.22326000e-01,  4.62465000e-01)
( 2.17321000e-01,  5.05844000e-01)
( 2.11904000e-01,  4.93922000e-01)
( 2.11838000e-01,  4.85996000e-01)
( 1.96859000e-01,  7.42119000e-01)
( 2.01330000e-01,  4.93957000e-01)
( 1.96574000e-01,  5.84953000e-01)
( 1.90643000e-01,  4.78160000e-01)
( 1.80327000e-01,  5.09956000e-01)
( 2.71239000e-01, -7.65584000e-01)
( 2.71243000e-01, -7.96850000e-01)
( 2.60581000e-01, -6.95171000e-01)
( 2.32935000e-01, -4.62509000e-01)
( 2.39290000e-01, -6.63782000e-01)
( 2.39313000e-01, -6.79476000e-01)
( 2.11665000e-01, -4.70100000e-01)
( 2.50000000e-01,  3.12500000e+00)
( 2.17929000e-01, -6.16550000e-01)
( 2.18010000e-01, -6.48037000e-01)
( 2.01330000e-01, -4.93956000e-01)
( 2.28749000e-01, -9.21870000e-01)
( 1.54346000e-01,  7.10784000e-01)
( 1.96348000e-01, -5.37527000e-01)
( 1.96573000e-01, -5.84938000e-01)
( 1.96843000e-01, -7.10796000e-01)
( 1.01250000e-01,  9.53122000e-01)
( 1.32974000e-01,  6.16407000e-01)
( 1.11863000e-01,  7.42098000e-01)
( 1.38038000e-01,  5.25704000e-01)
( 1.11560000e-01,  5.37441000e-01)
( 9.05584000e-02,  6.47891000e-01)
( 9.53979000e-02,  4.94526000e-01)
( 6.91783000e-02,  5.37394000e-01)
( 3.75000000e-02,  1.03125000e+00)
( 4.81045000e-02,  6.79341000e-01)
( 1.68832000e-01, -4.77949000e-01)
( 4.80656000e-02,  6.16296000e-01)
( 5.32795000e-02,  5.33356000e-01)
( 5.32630000e-02,  5.25484000e-01)
( 1.48621000e-01, -5.10135000e-01)
( 1.54273000e-01, -6.47937000e-01)
( 4.24033000e-02,  5.01866000e-01)
( 1.32972000e-01, -6.16375000e-01)
( 9.56376000e-02, -5.17829000e-01)
( 1.43750000e-01, -9.53121000e-01)
( 1.11851000e-01, -7.10747000e-01)
( 8.53049000e-02, -4.95793000e-01)
( 9.05377000e-02, -6.32086000e-01)
( 7.45890000e-02, -5.03009000e-01)
( 1.01248000e-01, -8.28102000e-01)
( 5.62495000e-03,  7.42086000e-01)
( 5.62468000e-03,  7.10740000e-01)
( 5.90000000e-01, -4.12500000e+00)
( 1.09021000e-02,  5.25541000e-01)
( 6.93639000e-02, -7.26409000e-01)
( 3.10306000e-04,  5.03100000e-01)
(-1.02818000e-02,  5.10006000e-01)
(-6.87496000e-02,  9.21870000e-01)
(-1.56152000e-02,  6.79329000e-01)
(-3.68675000e-02,  7.26420000e-01)
(-1.55840000e-02,  5.68836000e-01)
(-6.87490000e-02,  8.59362000e-01)
(-6.87483000e-02,  8.28105000e-01)
(-4.20431000e-02,  5.25442000e-01)
(-4.20142000e-02,  5.09516000e-01)
(-5.79815000e-02,  5.68854000e-01)
(-7.93062000e-02,  6.47854000e-01)
(-7.47546000e-02,  4.96946000e-01)
(-1.00600000e-01,  7.10752000e-01)
(-9.49918000e-02,  5.09964000e-01)
(-1.00337000e-01,  5.53060000e-01)
( 2.68037000e-02, -5.53008000e-01)
( 1.09031000e-02, -5.09982000e-01)
( 1.09014000e-02, -5.17722000e-01)
( 2.68559000e-02, -6.47799000e-01)
( 5.87497000e-02, -9.21869000e-01)
( 5.61854000e-03, -5.84599000e-01)
( 5.62500000e-03, -7.10714000e-01)
(-1.55784000e-02, -5.37260000e-01)
( 5.62493000e-03, -7.57728000e-01)
(-2.08654000e-02, -5.17621000e-01)
(-1.56114000e-02, -6.63549000e-01)
(-1.56195000e-02, -7.10713000e-01)
(-4.20132000e-02, -5.09507000e-01)
(-3.68052000e-02, -6.00408000e-01)
(-1.21855000e-01,  7.26433000e-01)
(-5.19471000e-02, -5.01636000e-01)
(-2.62491000e-02, -8.28101000e-01)
(-5.26184000e-02, -5.17612000e-01)
(-1.75000000e-01,  1.31250000e+00)
(-1.53750000e-01,  9.53122000e-01)
(-1.53748000e-01,  8.59362000e-01)
(-2.17500000e-01,  1.09375000e+00)
(-1.43043000e-01,  6.63650000e-01)
(-1.17495000e-01,  4.87595000e-01)
(-1.64358000e-01,  7.42103000e-01)
(-1.37202000e-01,  5.09409000e-01)
(-1.96249000e-01,  9.53122000e-01)
(-1.64125000e-01,  6.00564000e-01)
(-1.58373000e-01,  5.09410000e-01)
(-9.00000000e-02,  4.12500000e+00)
(-2.06850000e-01,  7.26461000e-01)
(-2.06712000e-01,  6.32241000e-01)
(-5.80184000e-02, -6.00418000e-01)
(-8.46931000e-02, -4.95887000e-01)
(-9.47878000e-02, -4.94614000e-01)
(-7.93042000e-02, -6.47811000e-01)
(-7.93454000e-02, -6.95024000e-01)
(-1.05548000e-01, -5.25441000e-01)
(-7.93674000e-02, -7.57731000e-01)
(-1.00514000e-01, -6.32051000e-01)
(-1.00539000e-01, -6.47827000e-01)
(-1.00576000e-01, -6.79319000e-01)
(-4.75000000e-02, -1.15625000e+00)
(-1.26603000e-01, -4.93734000e-01)
(-1.21514000e-01, -5.52993000e-01)
( 2.50000000e-01, -3.12500000e+00)
(-1.36592000e-01, -4.85476000e-01)
(-1.37201000e-01, -5.09404000e-01)
(-2.11319000e-01,  4.93916000e-01)
(-2.11265000e-01,  4.86047000e-01)
(-4.30000000e-01,  2.37500000e+00)
(-2.70592000e-01,  7.10833000e-01)
(-2.27634000e-01,  5.53309000e-01)
(-2.49057000e-01,  5.85043000e-01)
(-2.32362000e-01,  4.47222000e-01)
(-2.91742000e-01,  6.32448000e-01)
(-2.54930000e-01,  4.32920000e-01)
(-2.91604000e-01,  5.85248000e-01)
(-2.69775000e-01,  4.59368000e-01)
(-3.12772000e-01,  5.53879000e-01)
(-3.34341000e-01,  6.95219000e-01)
(-3.34329000e-01,  6.79566000e-01)
(-3.08035000e-01,  4.18255000e-01)
(-1.42892000e-01, -6.00497000e-01)
(-3.08156000e-01,  4.10617000e-01)
(-1.91360000e-01, -4.63946000e-01)
(-1.90117000e-01, -4.93858000e-01)
(-1.85047000e-01, -5.37233000e-01)
(-2.00630000e-01, -4.78174000e-01)
(-1.85296000e-01, -5.84769000e-01)
(-1.64351000e-01, -7.26435000e-01)
(-1.64362000e-01, -7.57743000e-01)
(-2.27441000e-01, -5.21663000e-01)
(-2.06849000e-01, -7.26453000e-01)
(-3.33996000e-01,  4.60278000e-01)
(-2.28034000e-01, -6.63739000e-01)
(-1.96250000e-01, -9.84372000e-01)
(-2.48985000e-01, -5.69239000e-01)
(-3.66243000e-01,  7.96854000e-01)
(-2.69789000e-01, -4.74902000e-01)
(-2.75271000e-01, -4.48001000e-01)
(-2.69939000e-01, -5.06289000e-01)
(-4.72500000e-01,  1.09375000e+00)
(-4.08747000e-01,  8.59368000e-01)
(-3.55358000e-01,  5.54130000e-01)
(-3.55275000e-01,  5.22759000e-01)
(-4.51241000e-01,  7.65605000e-01)
(-3.48352000e-01,  3.69141000e-01)
(-3.59851000e-01,  3.85603000e-01)
(-3.80643000e-01,  3.54004000e-01)
(-3.96791000e-01,  3.81536000e-01)
(-4.01750000e-01,  3.46182000e-01)
(-4.01665000e-01,  3.38339000e-01)
(-4.61821000e-01,  6.17089000e-01)
(-4.61771000e-01,  5.70146000e-01)
(-2.70023000e-01, -5.22037000e-01)
(-2.97298000e-01, -4.10517000e-01)
(-2.96792000e-01, -4.48477000e-01)
(-4.22391000e-01,  2.91362000e-01)
(-3.18667000e-01, -4.02657000e-01)
(-2.38749000e-01, -8.90617000e-01)
(-3.18373000e-01, -4.33461000e-01)
(-2.91703000e-01, -6.16722000e-01)
(-3.29053000e-01, -4.02373000e-01)
(-2.81237000e-01, -7.65579000e-01)
(-3.12829000e-01, -5.69611000e-01)
(-3.39524000e-01, -4.10005000e-01)
(-3.48352000e-01, -3.69141000e-01)
(-9.00000000e-02, -2.12500000e+00)
(-4.21827000e-01,  2.83131000e-01)
(-3.61903000e-01, -3.47817000e-01)
(-3.59578000e-01, -3.77509000e-01)
(-3.60261000e-01, -4.01654000e-01)
(-3.23742000e-01, -7.96849000e-01)
(-7.70000000e-01,  4.12500000e+00)
(-5.36250000e-01,  9.84374000e-01)
(-5.36249000e-01,  8.90623000e-01)
(-5.36248000e-01,  8.59371000e-01)
(-5.78750000e-01,  9.84375000e-01)
(-4.82901000e-01,  5.07549000e-01)
(-4.61016000e-01,  4.13286000e-01)
(-4.82586000e-01,  4.44804000e-01)
(-4.43749000e-01,  2.75933000e-01)
(-4.46183000e-01,  2.30609000e-01)
(-5.03882000e-01,  4.44862000e-01)
(-4.65035000e-01,  1.98010000e-01)
(-4.75620000e-01,  1.90536000e-01)
(-5.02893000e-01,  3.34865000e-01)
(-3.80586000e-01, -3.61729000e-01)
(-5.02668000e-01,  2.72437000e-01)
(-3.55358000e-01, -5.54129000e-01)
(-3.76474000e-01, -4.75964000e-01)
(-4.86186000e-01,  1.59428000e-01)
(-3.76682000e-01, -5.69916000e-01)
(-3.23748000e-01, -8.90618000e-01)
(-4.96840000e-01,  1.98328000e-01)
(-3.97889000e-01, -5.38644000e-01)
(-3.66239000e-01, -7.65593000e-01)
(-4.18798000e-01, -4.44698000e-01)
(-4.96581000e-01,  1.67126000e-01)
(-4.54436000e-01, -2.21334000e-01)
(-4.65401000e-01, -2.14027000e-01)
(-4.65298000e-01, -2.21754000e-01)
(-4.59920000e-01, -2.72241000e-01)
(-4.08727000e-01, -7.03070000e-01)
(-5.24318000e-01,  3.35017000e-01)
(-1.11000000e+00,  3.12500000e+00)
(-6.42500000e-01,  1.09375000e+00)
(-6.21249000e-01,  8.59373000e-01)
(-6.63750000e-01,  9.21874000e-01)
(-5.46677000e-01,  4.91996000e-01)
(-5.89296000e-01,  5.38982000e-01)
(-5.24210000e-01,  3.19357000e-01)
(-5.67451000e-01,  3.82308000e-01)
(-5.23610000e-01,  2.10107000e-01)
(-5.23169000e-01,  1.63274000e-01)
(-5.01245000e-01,  5.10549000e-02)
(-5.01344000e-01,  4.36157000e-02)
(-5.17216000e-01,  9.70441000e-02)
(-5.05146000e-01,  2.66490000e-02)
(-5.04429000e-01,  3.86136000e-03)
(-5.04429000e-01, -3.86136000e-03)
(-5.28218000e-01,  1.20472000e-01)
(-4.88697000e-01, -1.13987000e-01)
(-4.85136000e-01, -1.43150000e-01)
(-5.14572000e-01,  1.16140000e-02)
(-5.09878000e-01, -5.08441000e-02)
(-4.96408000e-01, -1.51517000e-01)
(-4.60651000e-01, -3.81771000e-01)
(-4.40355000e-01, -5.07451000e-01)
(-4.81334000e-01, -2.72354000e-01)
(-4.86702000e-01, -2.37415000e-01)
(-4.96729000e-01, -1.90503000e-01)
(-4.40547000e-01, -6.01417000e-01)
(-5.17558000e-01, -8.93356000e-02)
(-5.66973000e-01,  3.03989000e-01)
(-4.08750000e-01, -9.84374000e-01)
(-5.29626000e-01, -6.62397000e-02)
(-8.55000000e-01,  1.31250000e+00)
(-7.06249000e-01,  8.59374000e-01)
(-8.12500000e-01,  1.03125000e+00)
(-7.06243000e-01,  7.03117000e-01)
(-5.88808000e-01,  3.82398000e-01)
(-5.66628000e-01,  2.41499000e-01)
(-6.63636000e-01,  4.84281000e-01)
(-6.09863000e-01,  3.19851000e-01)
(-5.44141000e-01,  3.89030000e-02)
(-5.87298000e-01,  1.32317000e-01)
(-5.44141000e-01, -3.89030000e-02)
(-5.28019000e-01, -1.04861000e-01)
(-5.64177000e-01,  7.77461000e-03)
(-5.44244000e-01, -1.01047000e-01)
(-5.23904000e-01, -2.41352000e-01)
(-5.44552000e-01, -1.63339000e-01)
(-5.03037000e-01, -3.50556000e-01)
(-6.08786000e-01,  1.32387000e-01)
(-5.65767000e-01, -1.01115000e-01)
(-5.03768000e-01, -4.29152000e-01)
(-5.24444000e-01, -3.50697000e-01)
(-5.04056000e-01, -4.76245000e-01)
(-5.45479000e-01, -2.88241000e-01)
(-6.08688000e-01,  1.01227000e-01)
(-3.87500000e-01, -1.21875000e+00)
(-4.93726000e-01, -6.71830000e-01)
(-5.25556000e-01, -5.70222000e-01)
(-6.07960000e-01, -2.33638000e-02)
(-4.93747000e-01, -8.28118000e-01)
(-5.87810000e-01, -2.10339000e-01)
(-4.93749000e-01, -9.21873000e-01)
(-5.67637000e-01, -4.13675000e-01)
(-1.45000000e+00,  4.87500000e+00)
(-8.33749000e-01,  7.65624000e-01)
(-7.91244000e-01,  6.71869000e-01)
(-7.06087000e-01,  4.21761000e-01)
(-7.91222000e-01,  5.46853000e-01)
(-6.73799000e-01,  2.73168000e-01)
(-6.73616000e-01,  2.26277000e-01)
(-6.73412000e-01,  1.79406000e-01)
(-7.37794000e-01,  2.73295000e-01)
(-6.08786000e-01, -1.32387000e-01)
(-7.16272000e-01,  1.95129000e-01)
(-6.94617000e-01,  1.17009000e-01)
(-6.72819000e-01, -3.89867000e-02)
(-6.09712000e-01, -2.88569000e-01)
(-6.73106000e-01, -1.01367000e-01)
(-6.31036000e-01, -2.73024000e-01)
(-5.68039000e-01, -5.38971000e-01)
(-5.89144000e-01, -4.60741000e-01)
(-5.36236000e-01, -7.03100000e-01)
(-5.89223000e-01, -4.92050000e-01)
(-6.10219000e-01, -3.98138000e-01)
(-5.89277000e-01, -5.23341000e-01)
(-6.73412000e-01, -1.79406000e-01)
(-7.15867000e-01,  5.46174000e-02)
(-6.73549000e-01, -2.10651000e-01)
(-6.94617000e-01, -1.17009000e-01)
(-7.37547000e-01,  1.63921000e-01)
(-7.37511000e-01,  1.48303000e-01)
(-7.15826000e-01, -3.90123000e-02)
(-6.21241000e-01, -7.03111000e-01)
(-6.63589000e-01, -4.53001000e-01)
(-5.57500000e-01, -1.15625000e+00)
(-1.79000000e+00,  3.87500000e+00)
(-1.45000000e+00,  2.12500000e+00)
(-9.18750000e-01,  8.59375000e-01)
(-8.76248000e-01,  7.03123000e-01)
(-9.18749000e-01,  7.34374000e-01)
(-8.33710000e-01,  4.84348000e-01)
(-1.06750000e+00,  1.03125000e+00)
(-7.37380000e-01,  8.58516000e-02)
(-7.80106000e-01,  7.02673000e-02)
(-7.58676000e-01, -2.34184000e-02)
(-8.76158000e-01,  3.28084000e-01)
(-8.01435000e-01,  7.02743000e-02)
(-7.58769000e-01, -8.58694000e-02)
(-7.16408000e-01, -2.42007000e-01)
(-7.37585000e-01, -1.79541000e-01)
(-8.01391000e-01,  7.80746000e-03)
(-7.80159000e-01, -1.17116000e-01)
(-7.06126000e-01, -4.53033000e-01)
(-6.21249000e-01, -8.28122000e-01)
(-7.48560000e-01, -3.59265000e-01)
(-7.06204000e-01, -5.46835000e-01)
(-7.90968000e-01, -2.34279000e-01)
(-8.76075000e-01,  2.03079000e-01)
(-8.33384000e-01, -4.68551000e-02)
(-6.42500000e-01, -1.03125000e+00)
(-8.33454000e-01, -1.40571000e-01)
(-9.18705000e-01,  3.90602000e-01)
(-7.06250000e-01, -9.21875000e-01)
(-8.75982000e-01,  1.56204000e-02)
(-7.91230000e-01, -5.78108000e-01)
(-9.61241000e-01,  5.46869000e-01)
(-9.18607000e-01,  1.71845000e-01)
(-1.06750000e+00,  7.81250000e-01)
(-1.06750000e+00,  7.18749000e-01)
(-1.23750000e+00,  8.43750000e-01)
(-1.00372000e+00,  3.90613000e-01)
(-9.18556000e-01, -1.56217000e-02)
(-1.00365000e+00,  1.09364000e-01)
(-9.61111000e-01, -4.68682000e-02)
(-1.06744000e+00,  1.56240000e-01)
(-8.33698000e-01, -4.53091000e-01)
(-8.76174000e-01, -3.59337000e-01)
(-7.91248000e-01, -7.65623000e-01)
(-7.91249000e-01, -7.96874000e-01)
(-9.61209000e-01, -3.59357000e-01)
(-7.91250000e-01, -8.90625000e-01)
(-1.00369000e+00, -2.65608000e-01)
(-1.15246000e+00,  1.56245000e-01)
(-1.00373000e+00, -4.21864000e-01)
(-8.12500000e-01, -1.09375000e+00)
(-8.76250000e-01, -9.21875000e-01)
(-8.76250000e-01, -9.53125000e-01)
(-9.61247000e-01, -6.40623000e-01)
(-9.61249000e-01, -7.34374000e-01)
(-1.15245000e+00,  3.12486000e-02)
(-1.00375000e+00, -7.03124000e-01)
(-8.55000000e-01, -1.43750000e+00)
(-4.30000000e-01, -3.12500000e+00)
(-1.23748000e+00,  2.81245000e-01)
(-1.23747000e+00,  2.18746000e-01)
(-1.36500000e+00,  8.12500000e-01)
(-9.00000000e-02, -4.87500000e+00)
(-1.02500000e+00, -1.56250000e+00)
(-1.36498000e+00, -6.24990000e-02)
(-2.13000000e+00,  2.87500000e+00)
(-2.47000000e+00,  4.62500000e+00)
(-1.79000000e+00,  1.12500000e+00)
(-2.81000000e+00,  3.62500000e+00)
(-2.47000000e+00,  1.87500000e+00)
(-1.23750000e+00, -1.03125000e+00)
(-1.53500000e+00, -6.87500000e-01)
(-1.53500000e+00, -1.06250000e+00)
(-2.13000000e+00,  1.25000000e-01)
(-1.53500000e+00, -1.56250000e+00)
(-1.11000000e+00, -2.37500000e+00)
(-1.79000000e+00, -1.62500000e+00)
(-2.81000000e+00,  8.75000000e-01)
(-7.70000000e-01, -4.12500000e+00)
(-3.15000000e+00,  2.62500000e+00)
(-1.45000000e+00, -3.37500000e+00)
(-2.47000000e+00, -8.75000000e-01)
(-2.13000000e+00, -2.62500000e+00)
(-3.49000000e+00,  4.37500000e+00)
(-1.79000000e+00, -4.37500000e+00)
(-3.15000000e+00, -1.25000000e-01)
(-3.49000000e+00,  1.62500000e+00)
(-2.81000000e+00, -1.87500000e+00)
(-2.47000000e+00, -3.62500000e+00)
(-3.83000000e+00,  3.37500000e+00)
(-3.49000000e+00, -1.12500000e+00)
(-3.83000000e+00,  6.25000000e-01)
(-3.15000000e+00, -2.87500000e+00)
(-2.81000000e+00, -4.62500000e+00)
(-3.83000000e+00, -2.12500000e+00)
(-3.49000000e+00, -3.87500000e+00)
(-3.83000000e+00, -4.87500000e+00)};\label{line:pivotal_node}

\end{axis}
\end{tikzpicture}

%% file: figures/SparseCLCD.tikz
\begin{tikzpicture}
\begin{groupplot} [
group style={group size = 1 by 3, horizontal sep = 1.5cm, vertical sep = 1.5cm}]
\nextgroupplot[width=1\textwidth, xtick=\empty, height=0.33\textwidth, title=Sparse reconstruction at $Re\text{$=$}106$, ylabel=$C_{\mathrm{l}}$]
\addplot [thick, color=black]
coordinates {
( 0.00000000e+00, -9.49395600e-07)
( 1.00000000e+00, -1.72179500e-06)
( 2.00000000e+00, -2.94429500e-06)
( 3.00000000e+00, -3.59654300e-06)
( 4.00000000e+00, -3.69908200e-06)
( 5.00000000e+00, -3.33277000e-06)
( 6.00000000e+00, -2.67679200e-06)
( 7.00000000e+00, -1.83402900e-06)
( 8.00000000e+00, -1.04265900e-06)
( 9.00000000e+00, -3.39471800e-07)
( 1.00000000e+01, -1.91203700e-08)
( 1.10000000e+01, -4.36948400e-08)
( 1.20000000e+01, -4.57986700e-07)
( 1.30000000e+01, -1.13443400e-06)
( 1.40000000e+01, -1.95607500e-06)
( 1.50000000e+01, -2.60682400e-06)
( 1.60000000e+01, -2.99016500e-06)
( 1.70000000e+01, -2.89736100e-06)
( 1.80000000e+01, -2.36580000e-06)
( 1.90000000e+01, -1.48729600e-06)
( 2.00000000e+01, -4.57334900e-07)
( 2.10000000e+01,  4.21297600e-07)
( 2.20000000e+01,  8.57522600e-07)
( 2.30000000e+01,  6.79726700e-07)
( 2.40000000e+01, -1.54363100e-07)
( 2.50000000e+01, -1.50082300e-06)
( 2.60000000e+01, -3.03169100e-06)
( 2.70000000e+01, -4.30734600e-06)
( 2.80000000e+01, -4.90149500e-06)
( 2.90000000e+01, -4.51725200e-06)
( 3.00000000e+01, -3.13177800e-06)
( 3.10000000e+01, -9.84218800e-07)
( 3.20000000e+01,  1.39319300e-06)
( 3.30000000e+01,  3.33215600e-06)
( 3.40000000e+01,  4.12498900e-06)
( 3.50000000e+01,  3.37001700e-06)
( 3.60000000e+01,  1.02378900e-06)
( 3.70000000e+01, -2.46840800e-06)
( 3.80000000e+01, -6.25836500e-06)
( 3.90000000e+01, -9.24075800e-06)
( 4.00000000e+01, -1.03868600e-05)
( 4.10000000e+01, -8.98660500e-06)
( 4.20000000e+01, -5.02839300e-06)
( 4.30000000e+01,  6.97283500e-07)
( 4.40000000e+01,  6.79808500e-06)
( 4.50000000e+01,  1.15235500e-05)
( 4.60000000e+01,  1.31342600e-05)
( 4.70000000e+01,  1.06385900e-05)
( 4.80000000e+01,  3.98160200e-06)
( 4.90000000e+01, -5.39653600e-06)
( 5.00000000e+01, -1.53502700e-05)
( 5.10000000e+01, -2.29000200e-05)
( 5.20000000e+01, -2.53015300e-05)
( 5.30000000e+01, -2.07875200e-05)
( 5.40000000e+01, -9.83101600e-06)
( 5.50000000e+01,  5.90540900e-06)
( 5.60000000e+01,  2.20847000e-05)
( 5.70000000e+01,  3.42004200e-05)
( 5.80000000e+01,  3.76314600e-05)
( 5.90000000e+01,  2.98599600e-05)
( 6.00000000e+01,  1.13886500e-05)
( 6.10000000e+01, -1.43846300e-05)
( 6.20000000e+01, -4.09324300e-05)
( 6.30000000e+01, -6.03181700e-05)
( 6.40000000e+01, -6.54398200e-05)
( 6.50000000e+01, -5.23049500e-05)
( 6.60000000e+01, -2.12573100e-05)
( 6.70000000e+01,  2.09861100e-05)
( 6.80000000e+01,  6.45242800e-05)
( 6.90000000e+01,  9.59325200e-05)
( 7.00000000e+01,  1.03846800e-04)
( 7.10000000e+01,  8.11631900e-05)
( 7.20000000e+01,  3.00578400e-05)
( 7.30000000e+01, -4.03131200e-05)
( 7.40000000e+01, -1.11273200e-04)
( 7.50000000e+01, -1.62972200e-04)
( 7.60000000e+01, -1.74408600e-04)
( 7.70000000e+01, -1.36753100e-04)
( 7.80000000e+01, -5.11524500e-05)
( 7.90000000e+01,  6.46386900e-05)
( 8.00000000e+01,  1.82316300e-04)
( 8.10000000e+01,  2.65814000e-04)
( 8.20000000e+01,  2.83799400e-04)
( 8.30000000e+01,  2.18257800e-04)
( 8.40000000e+01,  7.54721300e-05)
( 8.50000000e+01, -1.17857100e-04)
( 8.60000000e+01, -3.11519700e-04)
( 8.70000000e+01, -4.50194800e-04)
( 8.80000000e+01, -4.71222100e-04)
( 8.90000000e+01, -3.53808800e-04)
( 9.00000000e+01, -1.04943100e-04)
( 9.10000000e+01,  2.21794700e-04)
( 9.20000000e+01,  5.41831100e-04)
( 9.30000000e+01,  7.51989900e-04)
( 9.40000000e+01,  7.62646300e-04)
( 9.50000000e+01,  5.29460800e-04)
( 9.60000000e+01,  7.92157700e-05)
( 9.70000000e+01, -4.91594000e-04)
( 9.80000000e+01, -1.00774900e-03)
( 9.90000000e+01, -1.28392700e-03)
( 1.00000000e+02, -1.17168600e-03)
( 1.01000000e+02, -6.25258900e-04)
( 1.02000000e+02,  2.59440400e-04)
( 1.03000000e+02,  1.22076500e-03)
( 1.04000000e+02,  1.91371700e-03)
( 1.05000000e+02,  2.00744600e-03)
( 1.06000000e+02,  1.33366500e-03)
( 1.07000000e+02,  2.47470800e-06)
( 1.08000000e+02, -1.56936100e-03)
( 1.09000000e+02, -2.76027200e-03)
( 1.10000000e+02, -2.98334900e-03)
( 1.11000000e+02, -1.96828500e-03)
( 1.12000000e+02,  5.89292900e-05)
( 1.13000000e+02,  2.35197900e-03)
( 1.14000000e+02,  3.89235000e-03)
( 1.15000000e+02,  3.86459100e-03)
( 1.16000000e+02,  2.08639900e-03)
( 1.17000000e+02, -8.43122600e-04)
( 1.18000000e+02, -3.69237900e-03)
( 1.19000000e+02, -5.08178900e-03)
( 1.20000000e+02, -4.23371400e-03)
( 1.21000000e+02, -1.35961900e-03)
( 1.22000000e+02,  2.37564300e-03)
( 1.23000000e+02,  5.27653400e-03)
( 1.24000000e+02,  5.87134200e-03)
( 1.25000000e+02,  3.76371900e-03)
( 1.26000000e+02, -2.22911900e-04)
( 1.27000000e+02, -4.34618600e-03)
( 1.28000000e+02, -6.63436400e-03)
( 1.29000000e+02, -5.89206700e-03)
( 1.30000000e+02, -2.37662300e-03)
( 1.31000000e+02,  2.40979900e-03)
( 1.32000000e+02,  6.27825000e-03)
( 1.33000000e+02,  7.30655600e-03)
( 1.34000000e+02,  4.93410300e-03)
( 1.35000000e+02,  1.98961000e-04)
( 1.36000000e+02, -4.81054100e-03)
( 1.37000000e+02, -7.69624800e-03)
( 1.38000000e+02, -6.99573700e-03)
( 1.39000000e+02, -3.03320200e-03)
( 1.40000000e+02,  2.45025300e-03)
( 1.41000000e+02,  6.92555700e-03)
( 1.42000000e+02,  8.19479800e-03)
( 1.43000000e+02,  5.62322700e-03)
( 1.44000000e+02,  4.18467100e-04)
( 1.45000000e+02, -5.10510000e-03)
( 1.46000000e+02, -8.31315100e-03)
( 1.47000000e+02, -7.60700000e-03)
( 1.48000000e+02, -3.36319300e-03)
( 1.49000000e+02,  2.50403800e-03)};\label{line:sparse_cl_cfd}

\addplot [thick, dashed, color=magenta]
coordinates {
( 0.00000000e+00, -1.72249100e-03)
( 1.00000000e+00, -5.70034400e-04)
( 2.00000000e+00, -4.83116700e-04)
( 3.00000000e+00, -5.98216100e-04)
( 4.00000000e+00, -5.39920000e-04)
( 5.00000000e+00, -5.17369800e-04)
( 6.00000000e+00, -4.90690800e-04)
( 7.00000000e+00, -3.49815000e-04)
( 8.00000000e+00, -3.53276000e-04)
( 9.00000000e+00, -3.10973800e-04)
( 1.00000000e+01, -1.46441800e-04)
( 1.10000000e+01, -1.04376300e-04)
( 1.20000000e+01, -5.23928300e-05)
( 1.30000000e+01, -9.93482100e-05)
( 1.40000000e+01, -1.75150800e-04)
( 1.50000000e+01, -2.11317500e-04)
( 1.60000000e+01, -2.61966700e-04)
( 1.70000000e+01, -2.25789800e-04)
( 1.80000000e+01, -1.74783900e-04)
( 1.90000000e+01, -1.66779200e-04)
( 2.00000000e+01, -2.24948200e-04)
( 2.10000000e+01, -1.57155600e-04)
( 2.20000000e+01, -1.73935000e-04)
( 2.30000000e+01, -1.61497900e-04)
( 2.40000000e+01, -1.38365800e-04)
( 2.50000000e+01, -1.44328200e-04)
( 2.60000000e+01, -1.27247000e-04)
( 2.70000000e+01, -5.29517600e-05)
( 2.80000000e+01, -6.34491600e-05)
( 2.90000000e+01, -5.37431800e-05)
( 3.00000000e+01, -1.36491600e-04)
( 3.10000000e+01, -1.04163700e-04)
( 3.20000000e+01, -1.46927500e-05)
( 3.30000000e+01, -4.40873200e-05)
( 3.40000000e+01, -4.86111400e-05)
( 3.50000000e+01, -1.33306000e-04)
( 3.60000000e+01, -2.70776500e-05)
( 3.70000000e+01, -5.58214200e-05)
( 3.80000000e+01, -1.19922400e-04)
( 3.90000000e+01, -8.27179300e-05)
( 4.00000000e+01, -6.97777500e-05)
( 4.10000000e+01, -4.93610800e-05)
( 4.20000000e+01, -6.32863200e-05)
( 4.30000000e+01, -1.30283000e-04)
( 4.40000000e+01, -9.70332500e-05)
( 4.50000000e+01, -1.34862900e-04)
( 4.60000000e+01, -1.94529800e-04)
( 4.70000000e+01, -4.49006100e-05)
( 4.80000000e+01, -5.23313300e-05)
( 4.90000000e+01, -9.68798700e-05)
( 5.00000000e+01, -9.96717600e-05)
( 5.10000000e+01, -1.43815100e-04)
( 5.20000000e+01, -1.55458000e-04)
( 5.30000000e+01, -1.61072600e-05)
( 5.40000000e+01,  2.45703600e-06)
( 5.50000000e+01, -3.13208600e-05)
( 5.60000000e+01,  1.78974900e-05)
( 5.70000000e+01, -3.16263000e-05)
( 5.80000000e+01, -2.21762500e-05)
( 5.90000000e+01,  4.03754200e-05)
( 6.00000000e+01,  2.91067400e-05)
( 6.10000000e+01, -6.55335900e-06)
( 6.20000000e+01, -1.30546100e-04)
( 6.30000000e+01, -1.87709500e-04)
( 6.40000000e+01, -2.35016800e-04)
( 6.50000000e+01, -1.98561100e-04)
( 6.60000000e+01, -2.65982300e-04)
( 6.70000000e+01, -2.00142400e-04)
( 6.80000000e+01, -1.94005900e-04)
( 6.90000000e+01, -1.76131700e-04)
( 7.00000000e+01,  1.40876900e-05)
( 7.10000000e+01,  9.43560200e-06)
( 7.20000000e+01, -1.06347200e-04)
( 7.30000000e+01, -1.03436500e-04)
( 7.40000000e+01, -1.84952000e-04)
( 7.50000000e+01, -2.47931200e-04)
( 7.60000000e+01, -3.37995100e-04)
( 7.70000000e+01, -2.53634600e-04)
( 7.80000000e+01, -1.68102500e-04)
( 7.90000000e+01, -7.57628400e-05)
( 8.00000000e+01, -3.52730200e-05)
( 8.10000000e+01,  1.27789700e-04)
( 8.20000000e+01,  1.77469000e-04)
( 8.30000000e+01,  2.48721300e-04)
( 8.40000000e+01,  2.33799400e-04)
( 8.50000000e+01, -1.60876200e-05)
( 8.60000000e+01, -2.48785200e-04)
( 8.70000000e+01, -4.70390300e-04)
( 8.80000000e+01, -6.56032100e-04)
( 8.90000000e+01, -6.10577200e-04)
( 9.00000000e+01, -3.61580000e-04)
( 9.10000000e+01, -2.09168700e-04)
( 9.20000000e+01,  3.48207700e-04)
( 9.30000000e+01,  5.89785600e-04)
( 9.40000000e+01,  7.42768200e-04)
( 9.50000000e+01,  6.39418300e-04)
( 9.60000000e+01,  3.27438000e-04)
( 9.70000000e+01, -1.86458900e-04)
( 9.80000000e+01, -8.62983800e-04)
( 9.90000000e+01, -1.39202400e-03)
( 1.00000000e+02, -1.58210600e-03)
( 1.01000000e+02, -1.37121700e-03)
( 1.02000000e+02, -6.02119100e-04)
( 1.03000000e+02,  3.45467000e-04)
( 1.04000000e+02,  1.14536200e-03)
( 1.05000000e+02,  1.66173700e-03)
( 1.06000000e+02,  1.62035600e-03)
( 1.07000000e+02,  4.66658100e-04)
( 1.08000000e+02, -9.43724600e-04)
( 1.09000000e+02, -2.52901600e-03)
( 1.10000000e+02, -3.37042500e-03)
( 1.11000000e+02, -3.07615700e-03)
( 1.12000000e+02, -1.45773800e-03)
( 1.13000000e+02,  8.52842600e-04)
( 1.14000000e+02,  2.86255900e-03)
( 1.15000000e+02,  3.84110700e-03)
( 1.16000000e+02,  2.76820200e-03)
( 1.17000000e+02,  2.81336900e-04)
( 1.18000000e+02, -3.21540400e-03)
( 1.19000000e+02, -5.74791300e-03)
( 1.20000000e+02, -5.40352500e-03)
( 1.21000000e+02, -2.76532000e-03)
( 1.22000000e+02,  1.47072100e-03)
( 1.23000000e+02,  5.12911000e-03)
( 1.24000000e+02,  6.41177000e-03)
( 1.25000000e+02,  4.54808200e-03)
( 1.26000000e+02,  4.81691000e-04)
( 1.27000000e+02, -4.64314300e-03)
( 1.28000000e+02, -7.66746200e-03)
( 1.29000000e+02, -6.91670900e-03)
( 1.30000000e+02, -2.94114300e-03)
( 1.31000000e+02,  2.83392000e-03)
( 1.32000000e+02,  7.29938700e-03)
( 1.33000000e+02,  7.98660500e-03)
( 1.34000000e+02,  4.76826700e-03)
( 1.35000000e+02, -1.07155900e-03)
( 1.36000000e+02, -6.79349500e-03)
( 1.37000000e+02, -9.35735700e-03)
( 1.38000000e+02, -7.21552200e-03)
( 1.39000000e+02, -1.50856100e-03)
( 1.40000000e+02,  5.14894600e-03)
( 1.41000000e+02,  8.94233100e-03)
( 1.42000000e+02,  8.32453400e-03)
( 1.43000000e+02,  3.65296400e-03)
( 1.44000000e+02, -3.15900400e-03)
( 1.45000000e+02, -8.67487100e-03)
( 1.46000000e+02, -1.00895600e-02)
( 1.47000000e+02, -6.74903800e-03)
( 1.48000000e+02, -9.81795800e-05)
( 1.49000000e+02,  6.45402800e-03)};\label{line:sparse_cl_diff}

\nextgroupplot[width=1\textwidth, xtick=\empty, height=0.33\textwidth, title=Sparse reconstruction at $Re\text{$=$}238$, ylabel=$C_{\mathrm{l}}$]
\addplot [thick, color=black, forget plot]
coordinates {
( 0.00000000e+00, -1.04417000e-06)
( 1.00000000e+00, -2.72545300e-06)
( 2.00000000e+00, -5.44592600e-06)
( 3.00000000e+00, -6.14386100e-06)
( 4.00000000e+00, -4.85416100e-06)
( 5.00000000e+00, -2.58948900e-06)
( 6.00000000e+00, -1.18861900e-07)
( 7.00000000e+00,  1.83497000e-06)
( 8.00000000e+00,  3.22514400e-06)
( 9.00000000e+00,  3.56980500e-06)
( 1.00000000e+01,  3.20157000e-06)
( 1.10000000e+01,  2.18922000e-06)
( 1.20000000e+01,  7.28687700e-07)
( 1.30000000e+01, -5.50421900e-07)
( 1.40000000e+01, -1.83982800e-06)
( 1.50000000e+01, -2.04685400e-06)
( 1.60000000e+01, -1.73299300e-06)
( 1.70000000e+01, -1.00883800e-06)
( 1.80000000e+01,  2.62132000e-07)
( 1.90000000e+01,  1.52684400e-06)
( 2.00000000e+01,  2.80823900e-06)
( 2.10000000e+01,  3.85423400e-06)
( 2.20000000e+01,  4.32167300e-06)
( 2.30000000e+01,  4.03472100e-06)
( 2.40000000e+01,  2.98651200e-06)
( 2.50000000e+01,  1.15325900e-06)
( 2.60000000e+01, -1.21964300e-06)
( 2.70000000e+01, -3.65079200e-06)
( 2.80000000e+01, -5.63597200e-06)
( 2.90000000e+01, -6.51288000e-06)
( 3.00000000e+01, -5.73096600e-06)
( 3.10000000e+01, -3.56585800e-06)
( 3.20000000e+01,  4.89630100e-07)
( 3.30000000e+01,  5.72163900e-06)
( 3.40000000e+01,  1.10309800e-05)
( 3.50000000e+01,  1.48130000e-05)
( 3.60000000e+01,  1.58855700e-05)
( 3.70000000e+01,  1.33612200e-05)
( 3.80000000e+01,  6.93579400e-06)
( 3.90000000e+01, -2.57923600e-06)
( 4.00000000e+01, -1.35341200e-05)
( 4.10000000e+01, -2.34682600e-05)
( 4.20000000e+01, -2.95893300e-05)
( 4.30000000e+01, -2.93088200e-05)
( 4.40000000e+01, -2.16991500e-05)
( 4.50000000e+01, -6.46828700e-06)
( 4.60000000e+01,  1.40218000e-05)
( 4.70000000e+01,  3.59806800e-05)
( 4.80000000e+01,  5.39435800e-05)
( 4.90000000e+01,  6.30041200e-05)
( 5.00000000e+01,  5.84494800e-05)
( 5.10000000e+01,  3.74766000e-05)
( 5.20000000e+01,  2.53815900e-06)
( 5.30000000e+01, -4.20621900e-05)
( 5.40000000e+01, -8.67437900e-05)
( 5.50000000e+01, -1.19962200e-04)
( 5.60000000e+01, -1.33165600e-04)
( 5.70000000e+01, -1.11524100e-04)
( 5.80000000e+01, -5.38315200e-05)
( 5.90000000e+01,  3.86100800e-05)
( 6.00000000e+01,  1.54142900e-04)
( 6.10000000e+01,  2.65037800e-04)
( 6.20000000e+01,  3.32045400e-04)
( 6.30000000e+01,  3.04640600e-04)
( 6.40000000e+01,  1.50143200e-04)
( 6.50000000e+01, -1.43953800e-04)
( 6.60000000e+01, -5.06333200e-04)
( 6.70000000e+01, -7.45924000e-04)
( 6.80000000e+01, -6.53822100e-04)
( 6.90000000e+01, -1.67164900e-04)
( 7.00000000e+01,  5.70855200e-04)
( 7.10000000e+01,  1.16929600e-03)
( 7.20000000e+01,  1.20255000e-03)
( 7.30000000e+01,  4.70422300e-04)
( 7.40000000e+01, -7.47382500e-04)
( 7.50000000e+01, -1.74801100e-03)
( 7.60000000e+01, -1.80416800e-03)
( 7.70000000e+01, -6.89281500e-04)
( 7.80000000e+01,  1.14873500e-03)
( 7.90000000e+01,  2.58026700e-03)
( 8.00000000e+01,  2.45610800e-03)
( 8.10000000e+01,  5.91048300e-04)
( 8.20000000e+01, -2.03396700e-03)
( 8.30000000e+01, -3.69433300e-03)
( 8.40000000e+01, -2.96044800e-03)
( 8.50000000e+01,  7.07794700e-05)
( 8.60000000e+01,  3.55179700e-03)
( 8.70000000e+01,  5.02808100e-03)
( 8.80000000e+01,  3.05132800e-03)
( 8.90000000e+01, -1.48153500e-03)
( 9.00000000e+01, -5.66353300e-03)
( 9.10000000e+01, -6.38740100e-03)
( 9.20000000e+01, -2.55028500e-03)
( 9.30000000e+01,  3.72655400e-03)
( 9.40000000e+01,  8.29359700e-03)
( 9.50000000e+01,  7.55608900e-03)
( 9.60000000e+01,  1.30871600e-03)
( 9.70000000e+01, -6.74218700e-03)
( 9.80000000e+01, -1.10651500e-02)
( 9.90000000e+01, -8.11626700e-03)
( 1.00000000e+02,  8.21210200e-04)
( 1.01000000e+02,  1.02480000e-02)
( 1.02000000e+02,  1.34319100e-02)
( 1.03000000e+02,  7.71973000e-03)
( 1.04000000e+02, -3.74411400e-03)
( 1.05000000e+02, -1.36980400e-02)
( 1.06000000e+02, -1.48284400e-02)
( 1.07000000e+02, -6.18617300e-03)
( 1.08000000e+02,  7.17929100e-03)
( 1.09000000e+02,  1.65185700e-02)
( 1.10000000e+02,  1.49713200e-02)
( 1.11000000e+02,  3.66341700e-03)
( 1.12000000e+02, -1.07113100e-02)
( 1.13000000e+02, -1.83712800e-02)
( 1.14000000e+02, -1.39417300e-02)
( 1.15000000e+02, -4.87173500e-04)
( 1.16000000e+02,  1.39391400e-02)
( 1.17000000e+02,  1.91619900e-02)
( 1.18000000e+02,  1.19669600e-02)
( 1.19000000e+02, -3.00899900e-03)
( 1.20000000e+02, -1.66014400e-02)
( 1.21000000e+02, -1.90149800e-02)
( 1.22000000e+02, -9.31620200e-03)
( 1.23000000e+02,  6.56673600e-03)
( 1.24000000e+02,  1.85741900e-02)
( 1.25000000e+02,  1.80624100e-02)
( 1.26000000e+02,  6.20000500e-03)
( 1.27000000e+02, -9.98496000e-03)
( 1.28000000e+02, -1.97921200e-02)
( 1.29000000e+02, -1.64302000e-02)
( 1.30000000e+02, -2.79314900e-03)
( 1.31000000e+02,  1.30929200e-02)
( 1.32000000e+02,  2.02757000e-02)
( 1.33000000e+02,  1.42246500e-02)
( 1.34000000e+02, -7.69316600e-04)
( 1.35000000e+02, -1.57596300e-02)
( 1.36000000e+02, -2.00870300e-02)
( 1.37000000e+02, -1.15500600e-02)
( 1.38000000e+02,  4.36849600e-03)
( 1.39000000e+02,  1.78827200e-02)
( 1.40000000e+02,  1.92543700e-02)
( 1.41000000e+02,  8.50577500e-03)
( 1.42000000e+02, -7.86739700e-03)
( 1.43000000e+02, -1.93750700e-02)
( 1.44000000e+02, -1.78493300e-02)
( 1.45000000e+02, -5.18544800e-03)
( 1.46000000e+02,  1.11529300e-02)
( 1.47000000e+02,  2.02143400e-02)
( 1.48000000e+02,  1.59034600e-02)
( 1.49000000e+02,  1.67529500e-03)};

\addplot [thick, dashed, color=magenta, forget plot]
coordinates {
( 0.00000000e+00, -1.16123700e-03)
( 1.00000000e+00, -3.67244900e-04)
( 2.00000000e+00, -2.31693500e-04)
( 3.00000000e+00, -2.31929300e-04)
( 4.00000000e+00, -2.47403600e-04)
( 5.00000000e+00, -2.86135200e-04)
( 6.00000000e+00, -3.53734100e-04)
( 7.00000000e+00, -3.88290900e-04)
( 8.00000000e+00, -3.03948500e-04)
( 9.00000000e+00, -3.44847300e-04)
( 1.00000000e+01, -3.63202700e-04)
( 1.10000000e+01, -3.10459700e-04)
( 1.20000000e+01, -3.51583400e-04)
( 1.30000000e+01, -3.40714100e-04)
( 1.40000000e+01, -3.59500100e-04)
( 1.50000000e+01, -3.37321400e-04)
( 1.60000000e+01, -4.06804400e-04)
( 1.70000000e+01, -3.31670100e-04)
( 1.80000000e+01, -3.62506900e-04)
( 1.90000000e+01, -3.62143700e-04)
( 2.00000000e+01, -3.59053500e-04)
( 2.10000000e+01, -3.89372200e-04)
( 2.20000000e+01, -3.70098300e-04)
( 2.30000000e+01, -3.66162300e-04)
( 2.40000000e+01, -4.17383000e-04)
( 2.50000000e+01, -4.58367000e-04)
( 2.60000000e+01, -4.57734600e-04)
( 2.70000000e+01, -4.41554200e-04)
( 2.80000000e+01, -4.63862400e-04)
( 2.90000000e+01, -4.52761000e-04)
( 3.00000000e+01, -4.04468600e-04)
( 3.10000000e+01, -4.71817000e-04)
( 3.20000000e+01, -4.66569900e-04)
( 3.30000000e+01, -4.38486800e-04)
( 3.40000000e+01, -4.55031100e-04)
( 3.50000000e+01, -3.78606600e-04)
( 3.60000000e+01, -4.21429600e-04)
( 3.70000000e+01, -4.10809100e-04)
( 3.80000000e+01, -4.58278400e-04)
( 3.90000000e+01, -4.46512900e-04)
( 4.00000000e+01, -4.87304400e-04)
( 4.10000000e+01, -5.40873400e-04)
( 4.20000000e+01, -5.30927900e-04)
( 4.30000000e+01, -5.20991000e-04)
( 4.40000000e+01, -5.41649400e-04)
( 4.50000000e+01, -5.15653900e-04)
( 4.60000000e+01, -4.98993100e-04)
( 4.70000000e+01, -5.15270500e-04)
( 4.80000000e+01, -4.47715500e-04)
( 4.90000000e+01, -4.48195400e-04)
( 5.00000000e+01, -4.36504100e-04)
( 5.10000000e+01, -4.31510800e-04)
( 5.20000000e+01, -4.93638600e-04)
( 5.30000000e+01, -5.36498900e-04)
( 5.40000000e+01, -5.76434600e-04)
( 5.50000000e+01, -6.41264700e-04)
( 5.60000000e+01, -5.72624700e-04)
( 5.70000000e+01, -5.70856800e-04)
( 5.80000000e+01, -5.37105500e-04)
( 5.90000000e+01, -3.94368000e-04)
( 6.00000000e+01, -2.80695200e-04)
( 6.10000000e+01, -2.05728600e-04)
( 6.20000000e+01, -1.69690600e-04)
( 6.30000000e+01, -2.81332900e-04)
( 6.40000000e+01, -5.12955900e-04)
( 6.50000000e+01, -7.90683300e-04)
( 6.60000000e+01, -1.19301100e-03)
( 6.70000000e+01, -1.46688800e-03)
( 6.80000000e+01, -1.36386700e-03)
( 6.90000000e+01, -8.78468300e-04)
( 7.00000000e+01, -1.22725000e-04)
( 7.10000000e+01,  4.93655000e-04)
( 7.20000000e+01,  6.24534500e-04)
( 7.30000000e+01,  1.40608900e-04)
( 7.40000000e+01, -9.98735300e-04)
( 7.50000000e+01, -2.16098000e-03)
( 7.60000000e+01, -2.37126100e-03)
( 7.70000000e+01, -1.44940100e-03)
( 7.80000000e+01,  3.35999700e-04)
( 7.90000000e+01,  1.87716900e-03)
( 8.00000000e+01,  2.25643800e-03)
( 8.10000000e+01,  7.92914600e-04)
( 8.20000000e+01, -1.84308600e-03)
( 8.30000000e+01, -3.91988500e-03)
( 8.40000000e+01, -3.78267900e-03)
( 8.50000000e+01, -1.07413300e-03)
( 8.60000000e+01,  2.70085200e-03)
( 8.70000000e+01,  4.90178800e-03)
( 8.80000000e+01,  3.82531300e-03)
( 8.90000000e+01, -4.18882500e-04)
( 9.00000000e+01, -4.91321500e-03)
( 9.10000000e+01, -6.83317200e-03)
( 9.20000000e+01, -4.03425100e-03)
( 9.30000000e+01,  2.08376300e-03)
( 9.40000000e+01,  7.57116400e-03)
( 9.50000000e+01,  8.61972300e-03)
( 9.60000000e+01,  4.10050100e-03)
( 9.70000000e+01, -3.75756400e-03)
( 9.80000000e+01, -8.97199900e-03)
( 9.90000000e+01, -7.79335800e-03)
( 1.00000000e+02, -6.44451400e-04)
( 1.01000000e+02,  8.86392700e-03)
( 1.02000000e+02,  1.46731800e-02)
( 1.03000000e+02,  1.19889400e-02)
( 1.04000000e+02,  2.36002300e-03)
( 1.05000000e+02, -8.27331100e-03)
( 1.06000000e+02, -1.28015700e-02)
( 1.07000000e+02, -8.03529700e-03)
( 1.08000000e+02,  4.21372200e-03)
( 1.09000000e+02,  1.59420500e-02)
( 1.10000000e+02,  1.80915300e-02)
( 1.11000000e+02,  9.14393600e-03)
( 1.12000000e+02, -5.43204300e-03)
( 1.13000000e+02, -1.65387800e-02)
( 1.14000000e+02, -1.58380300e-02)
( 1.15000000e+02, -4.75031600e-03)
( 1.16000000e+02,  1.00355200e-02)
( 1.17000000e+02,  1.99277100e-02)
( 1.18000000e+02,  1.66010500e-02)
( 1.19000000e+02,  3.76943500e-03)
( 1.20000000e+02, -1.17951300e-02)
( 1.21000000e+02, -1.86917300e-02)
( 1.22000000e+02, -1.32121500e-02)
( 1.23000000e+02,  9.56494500e-04)
( 1.24000000e+02,  1.59954400e-02)
( 1.25000000e+02,  2.10973200e-02)
( 1.26000000e+02,  1.33033700e-02)
( 1.27000000e+02, -2.61241800e-03)
( 1.28000000e+02, -1.67119600e-02)
( 1.29000000e+02, -1.86768500e-02)
( 1.30000000e+02, -8.47075000e-03)
( 1.31000000e+02,  7.58936300e-03)
( 1.32000000e+02,  2.01298600e-02)
( 1.33000000e+02,  1.92001800e-02)
( 1.34000000e+02,  7.17637600e-03)
( 1.35000000e+02, -9.32308600e-03)
( 1.36000000e+02, -1.90045400e-02)
( 1.37000000e+02, -1.57094600e-02)
( 1.38000000e+02, -2.43763500e-03)
( 1.39000000e+02,  1.38261800e-02)
( 1.40000000e+02,  2.13955100e-02)
( 1.41000000e+02,  1.53906500e-02)
( 1.42000000e+02,  4.11053600e-04)
( 1.43000000e+02, -1.50329100e-02)
( 1.44000000e+02, -1.94580400e-02)
( 1.45000000e+02, -1.13810500e-02)
( 1.46000000e+02,  4.16383200e-03)
( 1.47000000e+02,  1.85130500e-02)
( 1.48000000e+02,  2.04447300e-02)
( 1.49000000e+02,  1.01766400e-02)};

\nextgroupplot[width=1\textwidth, ytick={-0.03,0,0.03}, xlabel=$t$, ymax=0.035, title=Sparse reconstruction at $Re\text{$=$}382$, ymin=-0.035, ylabel=$C_{\mathrm{l}}$, height=0.33\textwidth]
\addplot [thick, color=black, forget plot]
coordinates {
( 0.00000000e+00, -1.28459700e-06)
( 1.00000000e+00, -3.68311600e-06)
( 2.00000000e+00, -6.09040600e-06)
( 3.00000000e+00, -3.62965500e-06)
( 4.00000000e+00,  9.26907000e-07)
( 5.00000000e+00,  5.28036300e-06)
( 6.00000000e+00,  8.47348700e-06)
( 7.00000000e+00,  1.01324300e-05)
( 8.00000000e+00,  1.04300300e-05)
( 9.00000000e+00,  9.21856300e-06)
( 1.00000000e+01,  7.49724900e-06)
( 1.10000000e+01,  5.96685300e-06)
( 1.20000000e+01,  4.67432400e-06)
( 1.30000000e+01,  3.93805200e-06)
( 1.40000000e+01,  3.81602900e-06)
( 1.50000000e+01,  4.21772500e-06)
( 1.60000000e+01,  4.90512900e-06)
( 1.70000000e+01,  5.86872300e-06)
( 1.80000000e+01,  6.79034100e-06)
( 1.90000000e+01,  8.13592400e-06)
( 2.00000000e+01,  9.37645900e-06)
( 2.10000000e+01,  1.05948600e-05)
( 2.20000000e+01,  1.14238400e-05)
( 2.30000000e+01,  1.15657100e-05)
( 2.40000000e+01,  1.09730700e-05)
( 2.50000000e+01,  8.97704700e-06)
( 2.60000000e+01,  5.80890600e-06)
( 2.70000000e+01,  1.72969200e-06)
( 2.80000000e+01, -2.78931900e-06)
( 2.90000000e+01, -6.42334700e-06)
( 3.00000000e+01, -7.57906300e-06)
( 3.10000000e+01, -5.70356700e-06)
( 3.20000000e+01, -9.05290500e-08)
( 3.30000000e+01,  8.38185400e-06)
( 3.40000000e+01,  1.83496700e-05)
( 3.50000000e+01,  2.79572600e-05)
( 3.60000000e+01,  3.52467800e-05)
( 3.70000000e+01,  3.85593900e-05)
( 3.80000000e+01,  3.69218000e-05)
( 3.90000000e+01,  2.96487700e-05)
( 4.00000000e+01,  1.57131400e-05)
( 4.10000000e+01, -2.56364600e-06)
( 4.20000000e+01, -2.33795900e-05)
( 4.30000000e+01, -4.29834700e-05)
( 4.40000000e+01, -5.79763100e-05)
( 4.50000000e+01, -6.36689800e-05)
( 4.60000000e+01, -5.71668000e-05)
( 4.70000000e+01, -3.77710800e-05)
( 4.80000000e+01, -5.22478200e-06)
( 4.90000000e+01,  3.60862800e-05)
( 5.00000000e+01,  7.91960100e-05)
( 5.10000000e+01,  1.18282100e-04)
( 5.20000000e+01,  1.49055500e-04)
( 5.30000000e+01,  1.57846000e-04)
( 5.40000000e+01,  1.27672300e-04)
( 5.50000000e+01,  4.43124600e-05)
( 5.60000000e+01, -8.43539600e-05)
( 5.70000000e+01, -2.27201300e-04)
( 5.80000000e+01, -3.30720100e-04)
( 5.90000000e+01, -3.38410200e-04)
( 6.00000000e+01, -1.72736000e-04)
( 6.10000000e+01,  1.86699900e-04)
( 6.20000000e+01,  6.29471500e-04)
( 6.30000000e+01,  8.34329600e-04)
( 6.40000000e+01,  4.99688800e-04)
( 6.50000000e+01, -3.37693400e-04)
( 6.60000000e+01, -1.04889900e-03)
( 6.70000000e+01, -1.03532400e-03)
( 6.80000000e+01, -1.51699100e-04)
( 6.90000000e+01,  1.02169200e-03)
( 7.00000000e+01,  1.49226600e-03)
( 7.10000000e+01,  7.63036100e-04)
( 7.20000000e+01, -7.37171700e-04)
( 7.30000000e+01, -1.87718500e-03)
( 7.40000000e+01, -1.61279700e-03)
( 7.50000000e+01,  1.20241800e-04)
( 7.60000000e+01,  2.08529700e-03)
( 7.70000000e+01,  2.59890500e-03)
( 7.80000000e+01,  7.76295200e-04)
( 7.90000000e+01, -2.16979000e-03)
( 8.00000000e+01, -3.81371000e-03)
( 8.10000000e+01, -2.43287300e-03)
( 8.20000000e+01,  1.51733900e-03)
( 8.30000000e+01,  4.89655300e-03)
( 8.40000000e+01,  4.41612800e-03)
( 8.50000000e+01, -3.36018700e-04)
( 8.60000000e+01, -5.95046400e-03)
( 8.70000000e+01, -7.53352400e-03)
( 8.80000000e+01, -2.76554100e-03)
( 8.90000000e+01,  5.66804800e-03)
( 9.00000000e+01,  1.08800500e-02)
( 9.10000000e+01,  7.54095900e-03)
( 9.20000000e+01, -3.07887300e-03)
( 9.30000000e+01, -1.34178400e-02)
( 9.40000000e+01, -1.40901300e-02)
( 9.50000000e+01, -3.37326300e-03)
( 9.60000000e+01,  1.22588800e-02)
( 9.70000000e+01,  1.97918800e-02)
( 9.80000000e+01,  1.22963500e-02)
( 9.90000000e+01, -5.67385500e-03)
( 1.00000000e+02, -2.15120500e-02)
( 1.01000000e+02, -2.08664900e-02)
( 1.02000000e+02, -5.11956200e-03)
( 1.03000000e+02,  1.64714000e-02)
( 1.04000000e+02,  2.57016500e-02)
( 1.05000000e+02,  1.59701900e-02)
( 1.06000000e+02, -5.61098000e-03)
( 1.07000000e+02, -2.45736600e-02)
( 1.08000000e+02, -2.40280200e-02)
( 1.09000000e+02, -7.00725600e-03)
( 1.10000000e+02,  1.65756200e-02)
( 1.11000000e+02,  2.73524600e-02)
( 1.12000000e+02,  1.78869900e-02)
( 1.13000000e+02, -4.17316500e-03)
( 1.14000000e+02, -2.45932500e-02)
( 1.15000000e+02, -2.52813300e-02)
( 1.16000000e+02, -8.73447200e-03)
( 1.17000000e+02,  1.52953600e-02)
( 1.18000000e+02,  2.76507900e-02)
( 1.19000000e+02,  1.93193500e-02)
( 1.20000000e+02, -2.33012500e-03)
( 1.21000000e+02, -2.37733700e-02)
( 1.22000000e+02, -2.60719600e-02)
( 1.23000000e+02, -1.04778600e-02)
( 1.24000000e+02,  1.35834700e-02)
( 1.25000000e+02,  2.75624600e-02)
( 1.26000000e+02,  2.06128600e-02)
( 1.27000000e+02, -3.50034800e-04)
( 1.28000000e+02, -2.26542700e-02)
( 1.29000000e+02, -2.66800400e-02)
( 1.30000000e+02, -1.22074700e-02)
( 1.31000000e+02,  1.16971400e-02)
( 1.32000000e+02,  2.72773900e-02)
( 1.33000000e+02,  2.18195000e-02)
( 1.34000000e+02,  1.65399500e-03)
( 1.35000000e+02, -2.13440800e-02)
( 1.36000000e+02, -2.71614900e-02)
( 1.37000000e+02, -1.38981500e-02)
( 1.38000000e+02,  9.71900800e-03)
( 1.39000000e+02,  2.68129200e-02)
( 1.40000000e+02,  2.29303800e-02)
( 1.41000000e+02,  3.64144200e-03)
( 1.42000000e+02, -1.98740700e-02)
( 1.43000000e+02, -2.75105500e-02)
( 1.44000000e+02, -1.55216200e-02)
( 1.45000000e+02,  7.69632300e-03)
( 1.46000000e+02,  2.61888800e-02)
( 1.47000000e+02,  2.39547500e-02)
( 1.48000000e+02,  5.60350000e-03)
( 1.49000000e+02, -1.82785900e-02)};

\addplot [thick, dashed, color=magenta, forget plot]
coordinates {
( 0.00000000e+00, -7.27957600e-04)
( 1.00000000e+00, -5.81752300e-04)
( 2.00000000e+00, -4.35326200e-04)
( 3.00000000e+00, -2.34241300e-04)
( 4.00000000e+00, -3.68533100e-04)
( 5.00000000e+00, -3.26711700e-04)
( 6.00000000e+00, -2.90373600e-04)
( 7.00000000e+00, -2.40249700e-04)
( 8.00000000e+00, -2.43944800e-04)
( 9.00000000e+00, -1.83560500e-04)
( 1.00000000e+01, -1.59193900e-04)
( 1.10000000e+01, -1.67571300e-04)
( 1.20000000e+01, -1.13431200e-04)
( 1.30000000e+01, -8.98604300e-05)
( 1.40000000e+01, -2.73144700e-04)
( 1.50000000e+01, -2.29779800e-04)
( 1.60000000e+01, -2.00273000e-04)
( 1.70000000e+01, -2.28436100e-04)
( 1.80000000e+01, -1.59292000e-04)
( 1.90000000e+01, -1.49502800e-04)
( 2.00000000e+01, -1.16418800e-04)
( 2.10000000e+01, -8.72330900e-05)
( 2.20000000e+01, -5.37212700e-05)
( 2.30000000e+01, -9.43559600e-05)
( 2.40000000e+01, -1.43462800e-04)
( 2.50000000e+01, -1.17997000e-04)
( 2.60000000e+01, -1.29194700e-04)
( 2.70000000e+01, -1.31906700e-04)
( 2.80000000e+01, -1.78205000e-04)
( 2.90000000e+01, -1.47156600e-04)
( 3.00000000e+01, -1.66693400e-04)
( 3.10000000e+01, -1.69609200e-04)
( 3.20000000e+01, -2.37806300e-04)
( 3.30000000e+01, -1.98499500e-04)
( 3.40000000e+01, -2.71128600e-04)
( 3.50000000e+01, -2.53882000e-04)
( 3.60000000e+01, -2.35747200e-04)
( 3.70000000e+01, -2.28380900e-04)
( 3.80000000e+01, -2.33622000e-04)
( 3.90000000e+01, -1.93680400e-04)
( 4.00000000e+01, -2.51237600e-04)
( 4.10000000e+01, -2.91457400e-04)
( 4.20000000e+01, -3.28285800e-04)
( 4.30000000e+01, -3.84830900e-04)
( 4.40000000e+01, -4.28468000e-04)
( 4.50000000e+01, -3.71159200e-04)
( 4.60000000e+01, -3.40094400e-04)
( 4.70000000e+01, -3.58759600e-04)
( 4.80000000e+01, -3.74378100e-04)
( 4.90000000e+01, -3.59612100e-04)
( 5.00000000e+01, -2.14462400e-04)
( 5.10000000e+01, -1.70186400e-04)
( 5.20000000e+01, -9.16185400e-05)
( 5.30000000e+01, -6.42637100e-05)
( 5.40000000e+01, -1.17906700e-04)
( 5.50000000e+01, -1.98833100e-04)
( 5.60000000e+01, -3.82446600e-04)
( 5.70000000e+01, -5.98303600e-04)
( 5.80000000e+01, -7.29719100e-04)
( 5.90000000e+01, -7.29720900e-04)
( 6.00000000e+01, -6.18565000e-04)
( 6.10000000e+01, -3.82067300e-04)
( 6.20000000e+01,  6.62650400e-05)
( 6.30000000e+01,  3.80000100e-04)
( 6.40000000e+01,  5.21741600e-04)
( 6.50000000e+01,  8.72947800e-05)
( 6.60000000e+01, -7.81767200e-04)
( 6.70000000e+01, -1.48005800e-03)
( 6.80000000e+01, -1.18851400e-03)
( 6.90000000e+01, -1.27987100e-04)
( 7.00000000e+01,  9.97243800e-04)
( 7.10000000e+01,  1.13943200e-03)
( 7.20000000e+01, -1.34268000e-04)
( 7.30000000e+01, -1.71390800e-03)
( 7.40000000e+01, -2.13726900e-03)
( 7.50000000e+01, -7.52241800e-04)
( 7.60000000e+01,  1.39242800e-03)
( 7.70000000e+01,  2.39438400e-03)
( 7.80000000e+01,  1.13390400e-03)
( 7.90000000e+01, -1.69319100e-03)
( 8.00000000e+01, -3.81865900e-03)
( 8.10000000e+01, -3.09215800e-03)
( 8.20000000e+01,  1.06031200e-03)
( 8.30000000e+01,  4.89420400e-03)
( 8.40000000e+01,  4.69342700e-03)
( 8.50000000e+01, -1.86922600e-04)
( 8.60000000e+01, -6.02248700e-03)
( 8.70000000e+01, -7.59478200e-03)
( 8.80000000e+01, -2.30876200e-03)
( 8.90000000e+01,  6.68637500e-03)
( 9.00000000e+01,  1.18145000e-02)
( 9.10000000e+01,  7.34477100e-03)
( 9.20000000e+01, -4.23961200e-03)
( 9.30000000e+01, -1.35157200e-02)
( 9.40000000e+01, -1.22570000e-02)
( 9.50000000e+01, -1.75759500e-04)
( 9.60000000e+01,  1.59889700e-02)
( 9.70000000e+01,  2.37198600e-02)
( 9.80000000e+01,  1.45167100e-02)
( 9.90000000e+01, -3.44675900e-03)
( 1.00000000e+02, -1.65897700e-02)
( 1.01000000e+02, -1.32548800e-02)
( 1.02000000e+02,  3.19397500e-03)
( 1.03000000e+02,  2.38120200e-02)
( 1.04000000e+02,  3.16160800e-02)
( 1.05000000e+02,  1.76266500e-02)
( 1.06000000e+02, -6.66827300e-03)
( 1.07000000e+02, -2.27786000e-02)
( 1.08000000e+02, -2.03258800e-02)
( 1.09000000e+02, -1.76092200e-03)
( 1.10000000e+02,  2.24480100e-02)
( 1.11000000e+02,  3.00948200e-02)
( 1.12000000e+02,  1.71529900e-02)
( 1.13000000e+02, -7.41398700e-03)
( 1.14000000e+02, -2.40823300e-02)
( 1.15000000e+02, -2.08191200e-02)
( 1.16000000e+02, -2.47923800e-03)
( 1.17000000e+02,  2.21939300e-02)
( 1.18000000e+02,  3.18771400e-02)
( 1.19000000e+02,  2.00730200e-02)
( 1.20000000e+02, -5.22585800e-03)
( 1.21000000e+02, -2.34735400e-02)
( 1.22000000e+02, -2.22899600e-02)
( 1.23000000e+02, -4.38035100e-03)
( 1.24000000e+02,  2.02248300e-02)
( 1.25000000e+02,  3.12937900e-02)
( 1.26000000e+02,  2.03567800e-02)
( 1.27000000e+02, -3.57284000e-03)
( 1.28000000e+02, -2.26898100e-02)
( 1.29000000e+02, -2.27545100e-02)
( 1.30000000e+02, -6.03351500e-03)
( 1.31000000e+02,  1.90128600e-02)
( 1.32000000e+02,  3.26610800e-02)
( 1.33000000e+02,  2.31692500e-02)
( 1.34000000e+02, -1.34504900e-03)
( 1.35000000e+02, -2.21787500e-02)
( 1.36000000e+02, -2.35335200e-02)
( 1.37000000e+02, -7.34621900e-03)
( 1.38000000e+02,  1.73731700e-02)
( 1.39000000e+02,  3.20351900e-02)
( 1.40000000e+02,  2.34971200e-02)
( 1.41000000e+02,  8.67229600e-05)
( 1.42000000e+02, -2.18216700e-02)
( 1.43000000e+02, -2.41503200e-02)
( 1.44000000e+02, -8.98083600e-03)
( 1.45000000e+02,  1.54929300e-02)
( 1.46000000e+02,  3.14359900e-02)
( 1.47000000e+02,  2.45024100e-02)
( 1.48000000e+02,  2.00062400e-03)
( 1.49000000e+02, -2.04916900e-02)};

\end{groupplot}\end{tikzpicture}

%% file: figures/sparse_nodes.tikz
\begin{tikzpicture}
\begin{axis}[
axis equal image,
width=1\textwidth,
xtick={-4, 0, 30},
ytick={-5, 0, 5},
xlabel=$x$,
ymax=5,
xmax=30,
ylabel=$y$,
xmin=-4,
ymin=-5]
\addplot [mark options={solid, thick}, mark=*, mark size=1, red, only marks]
coordinates {
(-3.83000000e+00, -4.87500000e+00)
(-3.49000000e+00, -3.62500000e+00)
(-3.15000000e+00, -2.37500000e+00)
(-2.81000000e+00, -1.12500000e+00)
(-2.47000000e+00,  1.25000000e-01)
(-2.13000000e+00,  1.37500000e+00)
(-1.79000000e+00,  2.62500000e+00)
(-1.45000000e+00,  3.87500000e+00)
(-7.70000000e-01, -4.87500000e+00)
(-4.30000000e-01, -3.62500000e+00)
(-9.00000000e-02, -2.37500000e+00)
( 8.53049000e-02, -4.95793000e-01)
( 5.90000000e-01,  2.37500000e+00)
( 9.30000000e-01,  3.62500000e+00)
( 1.27000000e+00,  4.87500000e+00)
( 1.95000000e+00, -3.87500000e+00)
( 2.29000000e+00, -2.62500000e+00)
( 2.54500000e+00, -1.43750000e+00)
( 2.84250000e+00, -2.18750000e-01)
( 3.22500000e+00,  1.06250000e+00)
( 3.65000000e+00,  2.37500000e+00)
( 3.99000000e+00,  3.62500000e+00)
( 4.33000000e+00,  4.87500000e+00)
( 5.01000000e+00, -3.87500000e+00)
( 5.35000000e+00, -2.62500000e+00)
( 5.69000000e+00, -1.37500000e+00)
( 6.03000000e+00, -1.25000000e-01)
( 6.37000000e+00,  1.12500000e+00)
( 6.71000000e+00,  2.37500000e+00)
( 7.05000000e+00,  3.62500000e+00)
( 7.39000000e+00,  4.87500000e+00)
( 8.07000000e+00, -3.87500000e+00)
( 8.41000000e+00, -2.62500000e+00)
( 8.75000000e+00, -1.37500000e+00)
( 9.09000000e+00, -1.25000000e-01)
( 9.43000000e+00,  1.12500000e+00)
( 9.77000000e+00,  2.37500000e+00)
( 1.01100000e+01,  3.62500000e+00)
( 1.04500000e+01,  4.87500000e+00)
( 1.11300000e+01, -3.87500000e+00)
( 1.14700000e+01, -2.62500000e+00)
( 1.18100000e+01, -1.37500000e+00)
( 1.21500000e+01, -1.25000000e-01)
( 1.24900000e+01,  1.12500000e+00)
( 1.28300000e+01,  2.37500000e+00)
( 1.31700000e+01,  3.62500000e+00)
( 1.35100000e+01,  4.87500000e+00)
( 1.41900000e+01, -3.87500000e+00)
( 1.45300000e+01, -2.62500000e+00)
( 1.48700000e+01, -1.37500000e+00)
( 1.52100000e+01, -1.25000000e-01)
( 1.55500000e+01,  1.12500000e+00)
( 1.58900000e+01,  2.37500000e+00)
( 1.62300000e+01,  3.62500000e+00)
( 1.65700000e+01,  4.87500000e+00)
( 1.72500000e+01, -3.87500000e+00)
( 1.75900000e+01, -2.62500000e+00)
( 1.79300000e+01, -1.37500000e+00)
( 1.82700000e+01, -1.25000000e-01)
( 1.86100000e+01,  1.12500000e+00)
( 1.89500000e+01,  2.37500000e+00)
( 1.92900000e+01,  3.62500000e+00)
( 1.96300000e+01,  4.87500000e+00)
( 2.03100000e+01, -3.87500000e+00)
( 2.06500000e+01, -2.62500000e+00)
( 2.09900000e+01, -1.37500000e+00)
( 2.13300000e+01, -1.25000000e-01)
( 2.16700000e+01,  1.12500000e+00)
( 2.20100000e+01,  2.37500000e+00)
( 2.23500000e+01,  3.62500000e+00)
( 2.26900000e+01,  4.87500000e+00)
( 2.33700000e+01, -3.87500000e+00)
( 2.37100000e+01, -2.62500000e+00)
( 2.40500000e+01, -1.37500000e+00)
( 2.43900000e+01, -1.25000000e-01)
( 2.47300000e+01,  1.12500000e+00)
( 2.50700000e+01,  2.37500000e+00)
( 2.54100000e+01,  3.62500000e+00)
( 2.57500000e+01,  4.87500000e+00)
( 2.64300000e+01, -3.87500000e+00)
( 2.67700000e+01, -2.62500000e+00)
( 2.71100000e+01, -1.37500000e+00)
( 2.74500000e+01, -1.25000000e-01)
( 2.77900000e+01,  1.12500000e+00)
( 2.81300000e+01,  2.37500000e+00)
( 2.84700000e+01,  3.62500000e+00)
( 2.88100000e+01,  4.87500000e+00)
( 2.94900000e+01, -3.87500000e+00)
( 2.98300000e+01, -2.62500000e+00)
(-2.48622000e-01, -5.06060000e-01)
(-1.00486000e-01,  6.16299000e-01)
( 4.07919000e-01,  3.81417000e-01)
( 3.10308000e-04, -5.03098000e-01)
(-5.17617000e-01,  1.20469000e-01)
(-5.07218000e-01,  1.28287000e-01)
( 5.08073000e-01,  1.28315000e-01)
(-1.84874000e-01, -5.05618000e-01)
( 5.13502000e-01, -2.41144000e-01)
( 3.45189000e-01,  4.29116000e-01)
(-2.11319000e-01,  4.93916000e-01)
( 2.01122000e-01, -4.70115000e-01)
(-5.02328000e-01,  2.10065000e-01)
( 2.10863000e-01,  4.61687000e-01)
(-4.33220000e-01, -2.60177000e-01)
( 4.11827000e-01,  2.98662000e-01)
(-3.39085000e-01, -3.86077000e-01)
(-1.02813000e-02, -5.09999000e-01)
( 5.07348000e-01,  1.51422000e-01)
(-4.54041000e-01,  2.28971000e-01)
( 4.75812000e-01, -1.98125000e-01)
(-4.85136000e-01, -1.43150000e-01)
( 5.21275000e-01,  4.30233000e-02)
(-4.85795000e-01,  1.74821000e-01)
( 3.08318000e-01, -4.02380000e-01)
( 4.74835000e-01,  1.74090000e-01)
(-9.18581000e-01,  1.09354000e-01)
(-2.17500000e-01, -1.03125000e+00)
( 4.30562000e-01, -6.32691000e-01)
( 1.14250000e+00, -5.31247000e-01)
( 1.82250000e+00, -7.81250000e-01)
( 2.16250000e+00,  9.68750000e-01)
( 2.84250000e+00,  7.18750000e-01)
(-2.27336000e-01, -5.05869000e-01)
( 7.48895000e-01, -5.46420000e-02)
(-5.65816000e-01, -1.32246000e-01)
( 5.55398000e-01,  7.78356000e-03)
(-7.91744000e-02, -5.68826000e-01)
( 5.14531000e-01, -3.66348000e-01)
( 4.65846000e-01,  2.52810000e-01)
(-6.08575000e-01,  7.00933000e-02)
( 1.32801000e-01, -5.53225000e-01)
( 6.63543000e-01,  2.41834000e-01)
(-5.02668000e-01,  2.72437000e-01)
(-4.81472000e-01, -2.56888000e-01)
(-3.67656000e-02, -5.53006000e-01)
( 5.34283000e-01,  1.78767000e-01)
( 7.38731000e-01, -6.09357000e-01)
(-5.23774000e-01,  2.25734000e-01)
( 2.60243000e-01, -5.69349000e-01)
(-3.34239000e-01, -6.16853000e-01)
( 1.75095000e-01,  5.21783000e-01)
(-4.55308000e-01,  2.45416000e-01)
( 3.45167000e-01, -4.13514000e-01)
( 5.13154000e-01,  2.09896000e-01)
( 3.11463000e-04, -5.17778000e-01)
(-5.07003000e-01, -1.43752000e-01)
(-4.22662000e-01, -3.30416000e-01)
(-1.02799000e-02,  5.17756000e-01)
( 5.30033000e-01, -2.73200000e-02)
( 1.38046000e-01, -5.02407000e-01)
(-4.76125000e-01,  2.37428000e-01)
(-3.70156000e-01, -3.77488000e-01)
(-1.53499000e+00, -3.12499000e-01)
(-9.18622000e-01, -2.03093000e-01)
(-2.38749000e-01, -8.90617000e-01)
( 5.05000000e-01,  1.43750000e+00)
( 1.14250000e+00, -6.56249000e-01)
( 1.48250000e+00,  9.37496000e-02)
( 1.82250000e+00,  8.43750000e-01)
( 2.54500000e+00, -1.31250000e+00)
( 2.84250000e+00, -6.56250000e-01)
( 3.22500000e+00,  4.37500000e-01)
( 3.56232000e-01, -7.34328000e-01)
( 4.51502000e-01,  4.91729000e-01)
(-2.27918000e-01,  6.16528000e-01)
( 2.17929000e-01, -6.16550000e-01)
(-7.58854000e-01,  1.48327000e-01)
(-4.39544000e-01, -3.97459000e-01)
( 2.60381000e-01, -6.00903000e-01)
( 4.23757000e-01,  3.23023000e-01)
(-7.48750000e-01, -9.84375000e-01)
(-8.33573000e-01,  2.65559000e-01)
(-5.46677000e-01, -4.91996000e-01)
(-1.64358000e-01, -7.42094000e-01)
( 3.98749000e-01,  8.90621000e-01)
( 7.38750000e-01,  8.90624000e-01)
( 9.08612000e-01,  7.81081000e-02)
( 1.39750000e+00, -5.93749000e-01)
( 1.39750000e+00,  6.56250000e-01)
( 1.73750000e+00, -9.37500000e-02)
( 2.07750000e+00, -8.43750000e-01)
( 2.07750000e+00,  4.06250000e-01)
( 2.41750000e+00, -3.43750000e-01)
( 2.41750000e+00,  9.06250000e-01)
( 2.75750000e+00,  1.56250000e-01)
( 2.92750000e+00,  2.18750000e-01)
(-6.72754000e-01,  2.33919000e-02)
(-7.93454000e-02, -6.95024000e-01)
( 4.08795000e-01, -4.60233000e-01)
( 7.38622000e-01, -4.21790000e-01)
( 7.06328000e-01,  2.41963000e-01)
(-6.95106000e-01, -2.57591000e-01)
(-2.70564000e-01, -6.79494000e-01)
(-3.66232000e-01,  7.34328000e-01)
( 9.04602000e-02,  5.84750000e-01)
(-5.67057000e-01, -3.19630000e-01)
(-1.42775000e-01, -5.68830000e-01)
( 5.35492000e-01, -3.19418000e-01)
( 9.93689000e-01,  2.34356000e-01)
(-7.15867000e-01, -5.46174000e-02)
(-6.94337000e-01,  3.90002000e-02)
( 7.06165000e-01, -1.32638000e-01)
( 7.48904000e-01, -8.58653000e-02)
( 6.41853000e-01,  1.32485000e-01)
(-5.03640000e-01, -4.13435000e-01)
( 5.36447000e-01, -4.44925000e-01)
( 5.26213000e-01,  6.40567000e-01)
( 3.24128000e-01, -5.85386000e-01)
( 4.93584000e-01,  4.13356000e-01)
(-3.12829000e-01, -5.69611000e-01)
(-1.19500000e+00, -1.56250000e+00)
(-8.97500000e-01,  1.09375000e+00)
(-1.75000000e-01,  1.68750000e+00)
( 8.45000000e-01, -1.56250000e+00)
( 6.93642000e-02,  7.26428000e-01)
(-9.61250000e-01, -8.59375000e-01)
(-9.61230000e-01,  4.53113000e-01)
(-8.76228000e-01, -5.15610000e-01)
(-3.23750000e-01, -9.84373000e-01)
( 1.62500000e-02, -9.84373000e-01)
( 3.56250000e-01, -9.84373000e-01)
( 6.96250000e-01, -9.84375000e-01)
( 8.23749000e-01, -7.65623000e-01)
( 4.80783000e-02,  6.32086000e-01)
( 6.20758000e-01, -2.57290000e-01)
(-4.40402000e-01, -5.23117000e-01)
(-8.33750000e-01, -9.84375000e-01)
(-5.46719000e-01,  5.07655000e-01)
( 1.86250000e-01, -9.84373000e-01)
( 5.26243000e-01,  7.65611000e-01)
( 8.23636000e-01,  3.59315000e-01)
( 5.62468000e-03,  7.10740000e-01)
( 6.85057000e-01,  2.73172000e-01)
(-2.91833000e-01,  6.95189000e-01)
(-9.61244000e-01,  5.78121000e-01)
(-7.48725000e-01, -5.78103000e-01)
(-7.91246000e-01,  7.03121000e-01)
(-3.23747000e-01, -8.59364000e-01)
(-4.08749000e-01,  9.21872000e-01)
(-1.43080000e-01, -6.95060000e-01)
(-1.53749000e-01,  8.90617000e-01)
( 1.75544000e-01, -6.63698000e-01)
( 6.53749000e-01, -8.90624000e-01)
( 7.38749000e-01,  8.28124000e-01)
( 9.08750000e-01, -7.96874000e-01)
( 7.70261000e-01, -1.63960000e-01)
( 9.08748000e-01,  7.03123000e-01)
(-1.56075000e-02, -6.47788000e-01)
(-2.28058000e-01,  6.79453000e-01)
( 2.81834000e-01, -6.95187000e-01)
( 1.75517000e-01,  6.47989000e-01)
( 6.21125000e-01, -3.19876000e-01)
( 6.21613000e-01,  4.29482000e-01)
(-7.16573000e-01,  3.20150000e-01)
(-4.40527000e-01, -5.85763000e-01)
( 4.73063000e-01, -6.17082000e-01)};\label{line:sparse_nodes}

\end{axis}
\end{tikzpicture}

%% file: main.bbl
\begin{thebibliography}{10}

\bibitem{brunton2019machine}
Steven~L Brunton, Bernd~R Noack, and Petros Koumoutsakos.
\newblock Machine learning for fluid mechanics.
\newblock {\em Annual Review of Fluid Mechanics}, 52, 2019.

\bibitem{climatenas}
National~Research Council.
\newblock {\em A National Strategy for Advancing Climate Modeling}.
\newblock The National Academies Press, 2012.

\bibitem{dhariwal2021diffusion}
Prafulla Dhariwal and Alexander Nichol.
\newblock Diffusion models beat gans on image synthesis.
\newblock {\em Advances in neural information processing systems}, 34:8780--8794, 2021.

\bibitem{du2024confild}
Pan Du, Meet~Hemant Parikh, Xiantao Fan, Xin-Yang Liu, and Jian-Xun Wang.
\newblock Confild: Conditional neural field latent diffusion model generating spatiotemporal turbulence.
\newblock {\em arXiv preprint arXiv:2403.05940}, 2024.

\bibitem{gao2024bayesian}
Han Gao, Xu~Han, Xiantao Fan, Luning Sun, Li-Ping Liu, Lian Duan, and Jian-Xun Wang.
\newblock Bayesian conditional diffusion models for versatile spatiotemporal turbulence generation.
\newblock {\em Computer Methods in Applied Mechanics and Engineering}, 427:117023, 2024.

\bibitem{geneva2022transformers}
Nicholas Geneva and Nicholas Zabaras.
\newblock Transformers for modeling physical systems.
\newblock {\em Neural Networks}, 146:272--289, 2022.

\bibitem{han2022predicting}
XU~HAN, Han Gao, Tobias Pfaff, Jian-Xun Wang, and Liping Liu.
\newblock Predicting physics in mesh-reduced space with temporal attention.
\newblock In {\em International Conference on Learning Representations}, 2022.

\bibitem{holzapfel2002nonlinear}
Gerhard~A Holzapfel.
\newblock Nonlinear solid mechanics: a continuum approach for engineering science, 2002.

\bibitem{jacobsen2023cocogen}
Christian Jacobsen, Yilin Zhuang, and Karthik Duraisamy.
\newblock Cocogen: Physically-consistent and conditioned score-based generative models for forward and inverse problems.
\newblock {\em arXiv preprint arXiv:2312.10527}, 2023.

\bibitem{jadhav2023stressd}
Yayati Jadhav, Joseph Berthel, Chunshan Hu, Rahul Panat, Jack Beuth, and Amir~Barati Farimani.
\newblock Stressd: 2d stress estimation using denoising diffusion model.
\newblock {\em Computer Methods in Applied Mechanics and Engineering}, 416:116343, 2023.

\bibitem{kaltenbach2020incorporating}
Sebastian Kaltenbach and Phaedon-Stelios Koutsourelakis.
\newblock Incorporating physical constraints in a deep probabilistic machine learning framework for coarse-graining dynamical systems.
\newblock {\em Journal of Computational Physics}, 419:109673, 2020.

\bibitem{kaltenbach2021physics}
Sebastian Kaltenbach and Phaedon-Stelios Koutsourelakis.
\newblock Physics-aware, probabilistic model order reduction with guaranteed stability.
\newblock {\em ICLR}, 2021.

\bibitem{karnakov2022optimizing}
Petr Karnakov, Sergey Litvinov, and Petros Koumoutsakos.
\newblock Optimizing a discrete loss (odil) to solve forward and inverse problems for partial differential equations using machine learning tools.
\newblock {\em arXiv preprint arXiv:2205.04611}, 2022.

\bibitem{karnakov2024solving}
Petr Karnakov, Sergey Litvinov, and Petros Koumoutsakos.
\newblock Solving inverse problems in physics by optimizing a discrete loss: Fast and accurate learning without neural networks.
\newblock {\em PNAS nexus}, page pgae005, 2024.

\bibitem{karniadakis2021physics}
George~Em Karniadakis, Ioannis~G Kevrekidis, Lu~Lu, Paris Perdikaris, Sifan Wang, and Liu Yang.
\newblock Physics-informed machine learning.
\newblock {\em Nature Reviews Physics}, 3(6):422--440, 2021.

\bibitem{keeling2005networks}
Matt~J Keeling and Ken~TD Eames.
\newblock Networks and epidemic models.
\newblock {\em Journal of the royal society interface}, 2(4):295--307, 2005.

\bibitem{kingma2013auto}
Diederik~P Kingma and Max Welling.
\newblock Auto-encoding variational bayes.
\newblock {\em arXiv preprint arXiv:1312.6114}, 2013.

\bibitem{koutsourelakis2009accurate}
Phaedon-Stelios Koutsourelakis.
\newblock Accurate uncertainty quantification using inaccurate computational models.
\newblock {\em SIAM Journal on Scientific Computing}, 31(5):3274--3300, 2009.

\bibitem{li2023multi}
Tianyi Li, Alessandra~S Lanotte, Michele Buzzicotti, Fabio Bonaccorso, and Luca Biferale.
\newblock Multi-scale reconstruction of turbulent rotating flows with generative diffusion models.
\newblock {\em Atmosphere}, 15(1):60, 2023.

\bibitem{li2020fourier}
Zongyi Li, Nikola Kovachki, Kamyar Azizzadenesheli, Burigede Liu, Kaushik Bhattacharya, Andrew Stuart, and Anima Anandkumar.
\newblock Fourier neural operator for parametric partial differential equations.
\newblock {\em arXiv preprint arXiv:2010.08895}, 2020.

\bibitem{lienen2023generative}
Marten Lienen, Jan Hansen-Palmus, David L{\"u}dke, and Stephan G{\"u}nnemann.
\newblock Generative diffusion for 3d turbulent flows.
\newblock {\em arXiv preprint arXiv:2306.01776}, 2023.

\bibitem{liu2024uncertainty}
Qiang Liu and Nils Thuerey.
\newblock Uncertainty-aware surrogate models for airfoil flow simulations with denoising diffusion probabilistic models.
\newblock {\em AIAA Journal}, pages 1--22, 2024.

\bibitem{lu2021learning}
Lu~Lu, Pengzhan Jin, Guofei Pang, Zhongqiang Zhang, and George~Em Karniadakis.
\newblock Learning nonlinear operators via deeponet based on the universal approximation theorem of operators.
\newblock {\em Nature machine intelligence}, 3(3):218--229, 2021.

\bibitem{menier2023interpretable}
Emmanuel Menier, Sebastian Kaltenbach, Mouadh Yagoubi, Marc Schoenauer, and Petros Koumoutsakos.
\newblock Interpretable learning of effective dynamics for multiscale systems.
\newblock {\em arXiv preprint arXiv:2309.05812}, 2023.

\bibitem{moser2023numerical}
Robert~D Moser.
\newblock Numerical challenges in turbulence simulation.
\newblock In {\em Numerical Methods in Turbulence Simulation}, pages 1--43. Elsevier, 2023.

\bibitem{Palmer2015}
Tim Palmer.
\newblock Modelling: Build imprecise supercomputers.
\newblock {\em Nature}, 526(7571):32--33, 2015.

\bibitem{price2023gencast}
Ilan Price, Alvaro Sanchez-Gonzalez, Ferran Alet, Timo Ewalds, Andrew El-Kadi, Jacklynn Stott, Shakir Mohamed, Peter Battaglia, Remi Lam, and Matthew Willson.
\newblock Gencast: Diffusion-based ensemble forecasting for medium-range weather.
\newblock {\em arXiv preprint arXiv:2312.15796}, 2023.

\bibitem{raissi2019physics}
Maziar Raissi, Paris Perdikaris, and George~E Karniadakis.
\newblock Physics-informed neural networks: A deep learning framework for solving forward and inverse problems involving nonlinear partial differential equations.
\newblock {\em Journal of Computational physics}, 378:686--707, 2019.

\bibitem{saharia2022photorealistic}
Chitwan Saharia, William Chan, Saurabh Saxena, Lala Li, Jay Whang, Emily~L Denton, Kamyar Ghasemipour, Raphael Gontijo~Lopes, Burcu Karagol~Ayan, Tim Salimans, et~al.
\newblock Photorealistic text-to-image diffusion models with deep language understanding.
\newblock {\em Advances in Neural Information Processing Systems}, 35:36479--36494, 2022.

\bibitem{shu2023physics}
Dule Shu, Zijie Li, and Amir~Barati Farimani.
\newblock A physics-informed diffusion model for high-fidelity flow field reconstruction.
\newblock {\em Journal of Computational Physics}, 478:111972, 2023.

\bibitem{song2020score}
Yang Song, Jascha Sohl-Dickstein, Diederik~P Kingma, Abhishek Kumar, Stefano Ermon, and Ben Poole.
\newblock Score-based generative modeling through stochastic differential equations.
\newblock {\em arXiv preprint arXiv:2011.13456}, 2020.

\bibitem{stabile2018finite}
Giovanni Stabile and Gianluigi Rozza.
\newblock Finite volume pod-galerkin stabilised reduced order methods for the parametrised incompressible navier--stokes equations.
\newblock {\em Computers \& Fluids}, 173:273--284, 2018.

\bibitem{vlachas2022multiscale}
Pantelis~R Vlachas, Georgios Arampatzis, Caroline Uhler, and Petros Koumoutsakos.
\newblock Multiscale simulations of complex systems by learning their effective dynamics.
\newblock {\em Nature Machine Intelligence}, 4(4):359--366, 2022.

\bibitem{vlachas2018}
Pantelis~R Vlachas, Wonmin Byeon, Zhong~Y Wan, Themistoklis~P Sapsis, and Petros Koumoutsakos.
\newblock Data-driven forecasting of high-dimensional chaotic systems with long short-term memory networks.
\newblock {\em Proc. R. Soc. A}, 474(2213):20170844, 2018.

\bibitem{wan2024debias}
Zhong~Yi Wan, Ricardo Baptista, Anudhyan Boral, Yi-Fan Chen, John Anderson, Fei Sha, and Leonardo Zepeda-N{\'u}{\~n}ez.
\newblock Debias coarsely, sample conditionally: Statistical downscaling through optimal transport and probabilistic diffusion models.
\newblock {\em Advances in Neural Information Processing Systems}, 36, 2024.

\bibitem{wang2021learning}
Sifan Wang, Hanwen Wang, and Paris Perdikaris.
\newblock Learning the solution operator of parametric partial differential equations with physics-informed deeponets.
\newblock {\em Science advances}, 7(40):eabi8605, 2021.

\bibitem{weller1998tensorial}
Henry~G Weller, Gavin Tabor, Hrvoje Jasak, and Christer Fureby.
\newblock A tensorial approach to computational continuum mechanics using object-oriented techniques.
\newblock {\em Computers in physics}, 12(6):620--631, 1998.

\bibitem{wilcox1988multiscale}
David~C Wilcox.
\newblock Multiscale model for turbulent flows.
\newblock {\em AIAA journal}, 26(11):1311--1320, 1988.

\end{thebibliography}
